\documentclass[acmtog, nonacm]{acmart}

\usepackage{url}
\usepackage{amsmath,graphicx}
\usepackage{multicol}
\usepackage{multirow}
\usepackage{svg}
\usepackage{overpic}
\usepackage{wrapfig}
\usepackage{capt-of}
\usepackage{tikz}
\usetikzlibrary{arrows.meta}
\usetikzlibrary{calc}
\usepackage{xcolor}
\usepackage[normalem]{ulem}
\usepackage{xspace}
\usepackage{enumitem}

\usepackage[switch]{lineno}

\renewcommand{\eqref}[1]{Eq. (\ref{#1})}
\newcommand{\tabref}[1]{Tab. \ref{#1}}
\newcommand{\figref}[1]{Fig. \ref{#1}}
\newcommand{\secref}[1]{Sec. \ref{#1}}
\newcommand{\oursw}[0]{Ours-wGlare\xspace}
\newcommand{\ourswo}[0]{Ours\xspace}
\definecolor{psnrgreen}{RGB}{17,127,5}
\usepackage{booktabs}
\usepackage{pifont}
\newcommand{\cmark}{\textcolor{green!60!black}{\ding{51}}}
\newcommand{\xmark}{\textcolor{red!70!black}{\ding{55}}}
\newcommand{\pmark}{\textcolor{orange!85!black}{\ding{51}}}  %
\AtBeginDocument{%
  }

\begin{document}

\title{Lens Flare Removal and Reconstruction}

\author{Tarun Yenamandra}
\authornote{Work done during an internship at Meta Reality Labs Research.}
\affiliation{%
  \institution{Meta Reality Labs Research}
  \city{Redmond}
  \country{USA}
}

\author{Jonathon Luiten}
\affiliation{%
  \institution{Meta Reality Labs}
  \city{New York}
  \country{USA}}

\author{Daniel Cremers}
\affiliation{%
  \institution{TU Munich, MCML}
  \city{Munich}
  \country{Germany}}

\author{Nathan Matsuda}
\affiliation{%
  \institution{Meta Reality Labs Research}
  \city{Redmond}
  \country{USA}
}

\renewcommand{\shortauthors}{Yenamandra et al.}

\begin{abstract}
The presence of lens flares in images can significantly reduce the quality of downstream application results for tasks such as 3D scene reconstruction. This is because lens flares are a property of the camera imaging system, and not a part of the underlying scene being modeled. There are previous methods that tackle the removal of small flares focused around a light source. However, existing methods struggle with large flares, such as those that fill the entire image. In this work, we compile a novel dataset for large-flare removal, combining publicly available real-world data with a procedural generation pipeline. We fine-tune a diffusion-based model on our dataset to remove complex, large lens flares. 
On the other hand, lens flares remain effective artistic tools, widely used in the media. While there are ways to simulate 2D flares, representing and reconstructing lens flares consistently across multiple views has not yet been explored. To achieve this, we introduce a flare representation model that leverages the symmetry of lens flares about the camera's principal point.  We propose a computational pipeline to jointly optimize this flare model and a Gaussian splatting model (3DGS). This enables the decomposition of a 3D scene into lens flares and the scene itself, using our flare-removal model. Because the reconstructed flare is explicit and re-renderable, it can be edited and transferred to novel images and new 3D scenes. We evaluate removal on an established benchmark and a new one for large reflective flares, quantify the flare/scene decomposition directly, and show that the pipeline is robust to errors in automatic light-source localization.
\end{abstract}

\ccsdesc[300]{Computing methodologies~Computational photography}
\ccsdesc[500]{Computing methodologies~Appearance and texture representations}
\ccsdesc[300]{Computing methodologies~Shape representations}

\begin{teaserfigure}
  \begin{center}
    \centering
    \begin{overpic}[width=\linewidth]{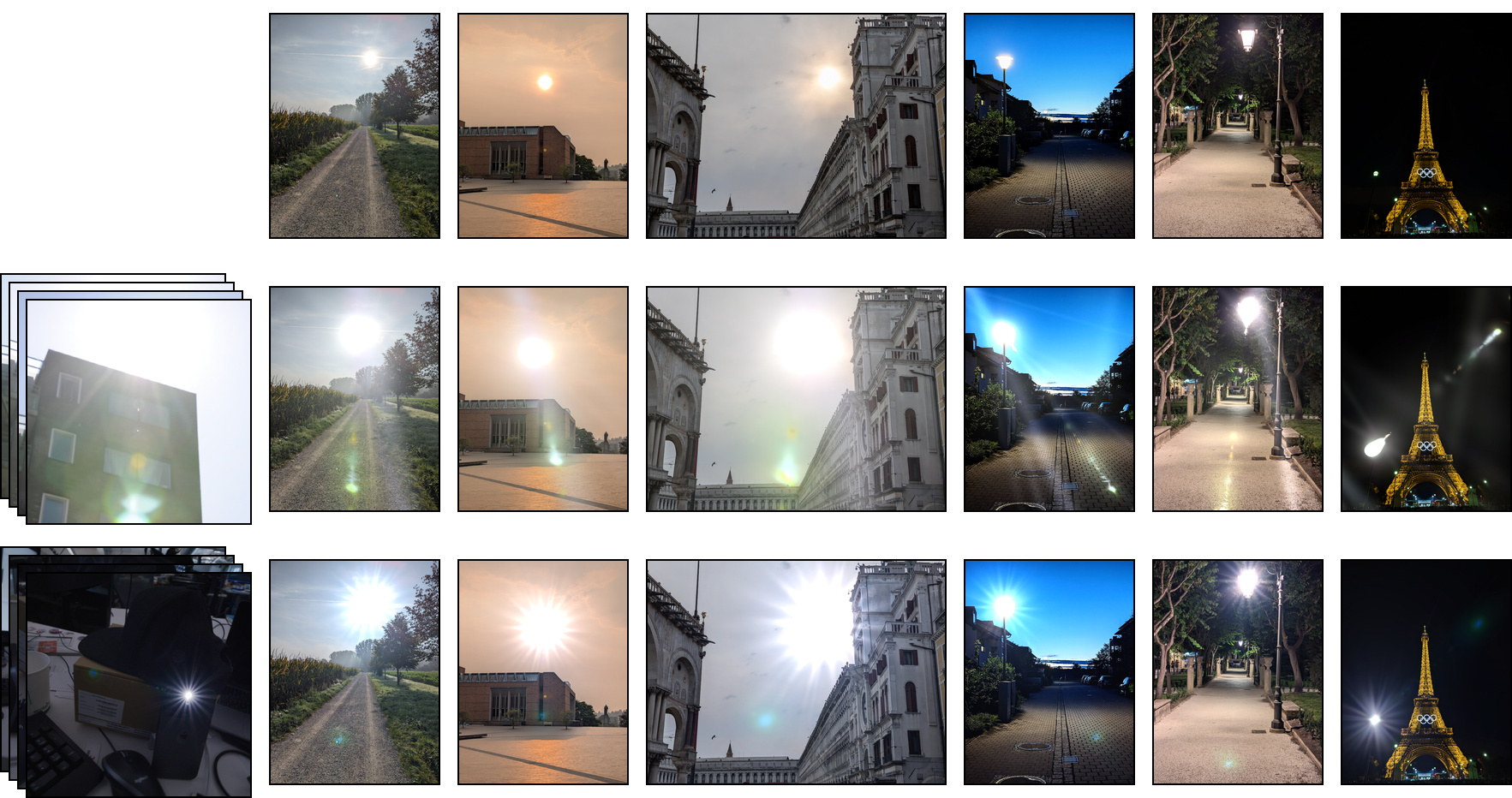}
        \put(53,53){\small\bfseries Target Images}
        \put(-0.5,-1.5){\small\bfseries Source Flares from Scenes}
        \put(46,-1.5){\small\bfseries Lens Flares Transferred to Images}
    \end{overpic}
    \captionof{figure}{%
      \textbf{Lens-flare transfer to images.}
        Given multi-view images (left column) of scenes with lens flares, our model decomposes each scene into a flare-free scene and our 3DGS-based lens-flare representation. We use the reconstructed flares to transfer them onto a diverse set of target images (top row). Each following row composites the source flare reconstructed on the
  left onto every target image shown above it.}
    \label{fig:teaser}
  \end{center}
\end{teaserfigure}

\maketitle

\section{Introduction}
\label{sec:intro}
\begin{figure}
    \centering
    \begin{overpic}[width=0.8\linewidth,]{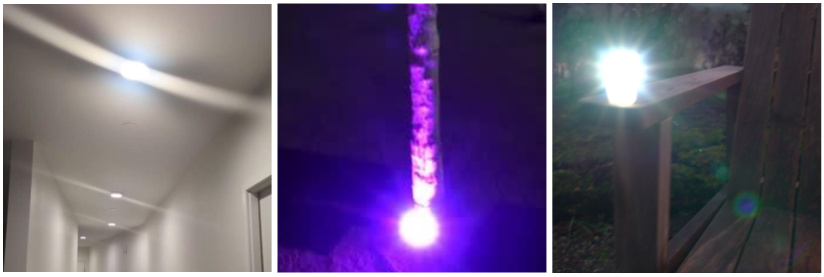}

    \put(9,34){\small Streaks}
    \put(45,34.5){\small Glare}
    \put(24,-4.5){\small Scattering}
    \put(73,-4.5){\small Reflective}
    \end{overpic}
    \captionof{figure}{Categories of lens flares for our discussion.}
    \Description{Three photographs of lens flares. The left two, grouped under the label ``Scattering'', show a hallway ceiling light with radial streaks (labeled ``Streaks'') and a bright purple-white light with a hazy vertical beam (labeled ``Glare''). The right photo, labeled ``Reflective'', shows a bright light on a wooden deck railing at night with a distinct circular reflective ghost artifact nearby.}
    \label{fig:lensflaretypes}
\end{figure}
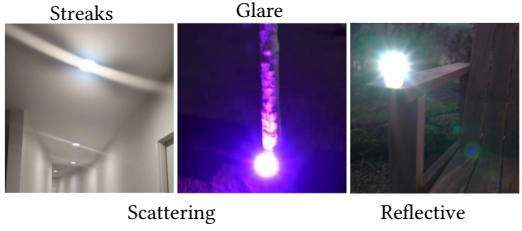

Lens flares are artifacts caused by unintended light paths passing through the lens when a camera observes a sufficiently bright light source in the scene. In this work, we will refer to two subjective categories of lens flare appearance, see~\figref{fig:lensflaretypes}. \textit{Scattering flares} occur due to dust, lens defects, and scratches on lenses, causing lens glares, shimmer, and streaks in images, respectively. \textit{Reflective flares} occur due to lens ensembles where lenses reflect a small percentage of the incident light.

\subsection{Lens flare removal} In many downstream applications, it is important to remove lens flares. Existing flare removal methods ~\cite{dai2023flare7kpp,wu2021train,zhou2025improving} use real and synthetic lens flare datasets to add synthetic flares to images from a dataset without flares and learn to predict the original flare-free images. Such datasets are mainly composed of scattering flares. Although one dataset contains a few reflective flares, they cover a relatively small area and are not full-frame reflective flares. To augment these datasets, we compile a novel dataset with real-world large reflective flares and augment it with procedurally generated flares. As baselines, we train previously proposed models on our dataset for flare removal. Furthermore, we propose a novel approach that demonstrates that fine-tuning a diffusion model with LoRA for flare removal on these datasets yields better results, which also lead to better multi-view 3D reconstructions (\secref{sec:lensflaresremoval3d}). We evaluate all of the models on the scattering-flare-removal benchmark from Flare7K++ ~\cite{dai2023flare7kpp} and propose a new benchmark for large reflective-flare removal. Further, we compile a dataset of $5$ flare corrupted scenes to evaluate the performance of the flare-removal models across multiple views. We demonstrate that, on each benchmark, one of our two models outperforms all prior methods.

\begin{table*}[t]
    \centering
    \begin{tabular}{lcccc}
        \toprule
        & \textbf{Stock flares} & \textbf{Procedural} & \textbf{Physically-based} & \textbf{Ours} \\
        \midrule
        No lens/optical model needed                             & \cmark & \cmark & \xmark & \cmark \\
        No precise 3D light geometry needed$^{d}$                       & \cmark & \cmark & \xmark & \cmark \\
        Viewpoint-correct optical appearance$^{a}$                  & \xmark & \xmark & \pmark & \cmark \\
        Reproduces the target scene's own captured flare signature & \xmark & \xmark & \xmark & \cmark \\
        3D-consistent transfer of a captured flare to new scenes$^{c}$                  & \xmark & \xmark & \xmark & \cmark \\
        Real-time rendering$^{b}$                                    & \cmark & \cmark & \pmark & \cmark \\
        \bottomrule
    \end{tabular}
    \caption{Comparison against a stock (real, but unrelated) captured flare, procedural (2D/game-engine) simulators, and physically-based (ray-traced) simulators, the latter two used as forward renderers configured by hand or from a lens prescription rather than fitted to the capture. $^{a}$Ghost/streak shapes correctly reflowing with viewpoint change; physically-based methods achieve this only when the lens prescription matches the real lens, which generic models miss (unit-specific imperfections, dust, coatings). For our method this holds when re-rendering the reconstructed scene; in transfer, appearance is queried at the source viewpoint by design (\secref{sec:transfer}). $^{b}$Real-time procedural and GPU physically-based demos exist, but high-fidelity physically-based rendering (diffraction, multi-bounce) is typically offline. $^{c}$A stock flare is a real capture, but of an unrelated lens/light, and is pasted in 2D without adapting to the target camera's geometry. $^{d}$Our method needs only an approximate 3D light position, estimated automatically, and is robust to large errors in it (\secref{sec:light_robustness}).}
    \label{tab:flare_modelling_comparison}
\end{table*}
\subsection{Lens flare representation and reconstruction} 
\begin{figure}[ht]
\centering
\vspace{2mm}
\begin{overpic}[width=0.3\linewidth,]   {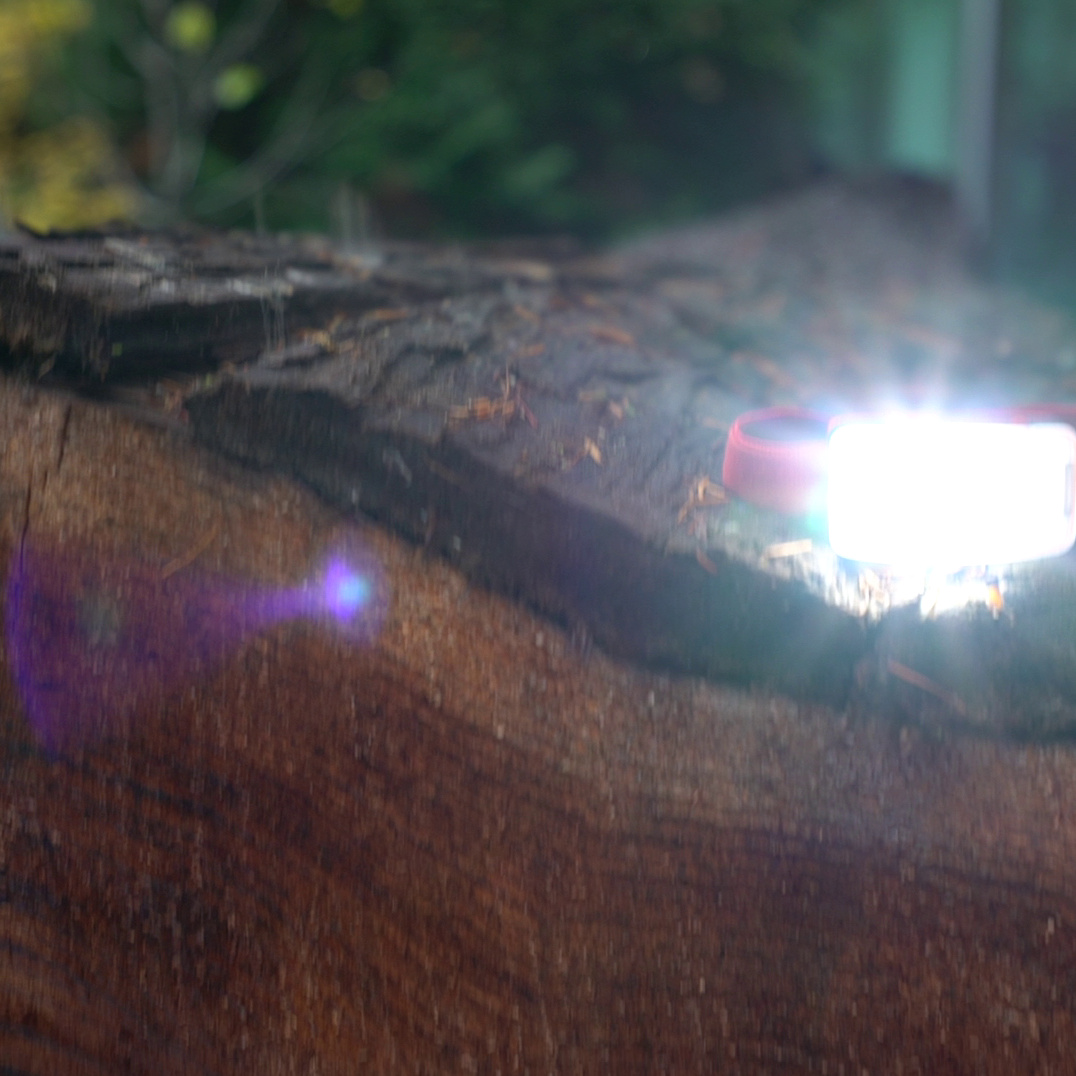}
\linethickness{1pt}
\put(0,-3){\line(1,0){203}}
\put(25,-13){\small Same light, different lenses}
\end{overpic}
\begin{overpic}[width=0.3\linewidth,]   {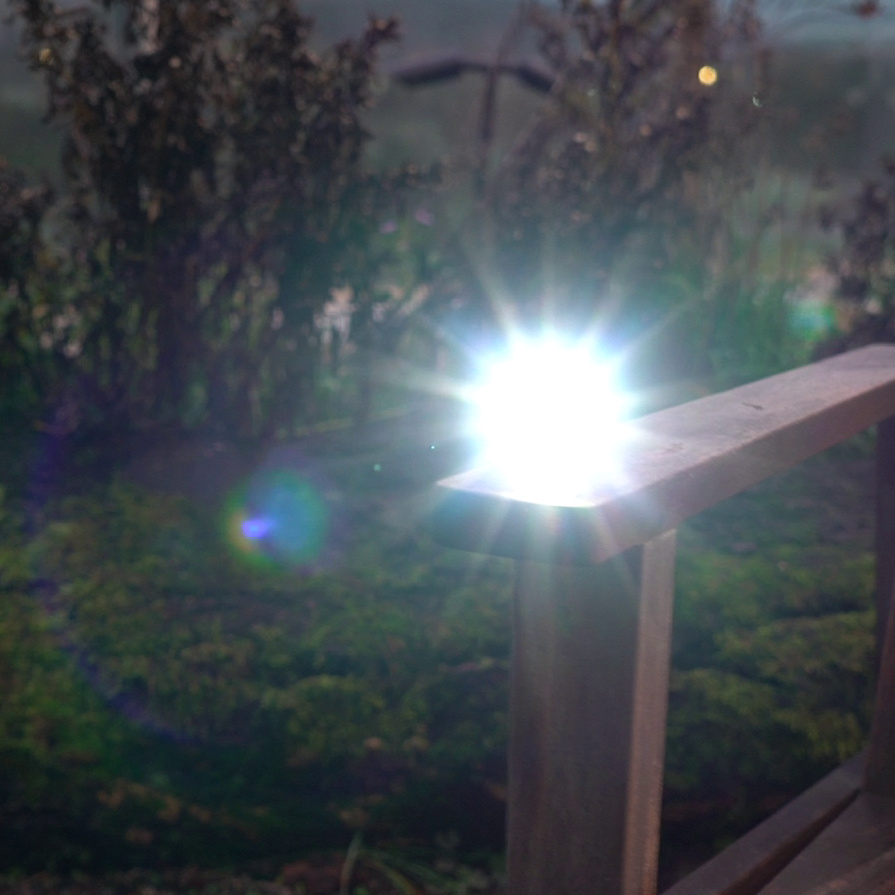}
\linethickness{1pt}
\put(0,103){\line(1,0){203}}
\put(25,107){\small Different lights, same lens}
\end{overpic}
\begin{overpic}[width=0.3\linewidth,]   {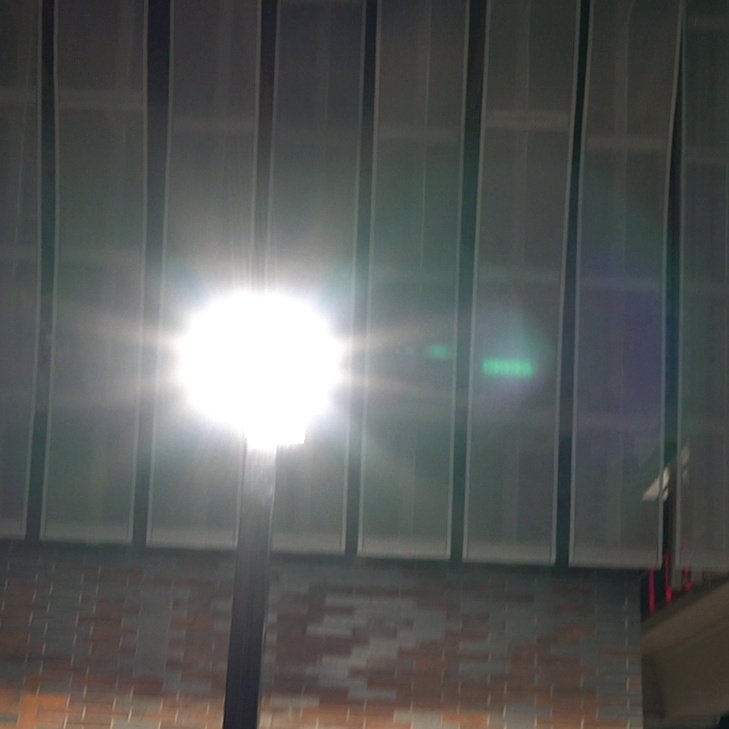}
\end{overpic}
\captionof{figure}{Varying lens flares due to lenses and light sources from our dataset.}
\Description{Three photographs illustrating flare variation. The first two, under the label ``Same light, different lenses'', show a tree-trunk scene and a chair scene each with a different lens flare pattern despite similar lighting. The third photo, under the label ``Different lights, same lens'', shows a distinct starburst-shaped light source flare.}
\label{fig:lensflareshapes}
  \end{figure}

Lens flares are an important artistic tool in film, games, and virtual production, where they add cinematic realism and convey the presence of bright light sources. For downstream reconstruction tasks, although they are undesirable, understanding their effects and synthesizing them would help us better control the 3D reconstruction as lens flares are unavoidable in many situations. Existing flare simulators -- stock compositing, procedural, and physically-based (\tabref{tab:flare_modelling_comparison}) -- can synthesize plausible flares, but none reconstruct the \emph{actual} pattern of a given capture: physically-based methods only achieve viewpoint-correct appearance when the lens prescription is precisely known, diffractive effects, among the most distinctive aspects of lens flares, are computationally intractable to simulate at macroscopic lens scales, and modeling every possible real-world light source is not scalable. Our goal is instead to reconstruct the actual flare patterns present in captured footage, preserving lens- and capture-specific characteristics. For instance, in~\figref{fig:lensflareshapes}, we see that the same lens leads to different types of lens flares with different light sources. Further, realistically editing lens flares into images requires extensive manual artistic effort. A view-consistent 3D flare representation is therefore valuable beyond removal: it enables reconstruction, and transfer of captured flares onto novel target images (see~\figref{fig:teaser}) and onto novel target scenes (see~\figref{fig:scenetransfer}), none of which a per-image 2D removal model can provide.

\begin{figure}[ht]
\centering
\begin{overpic}[width=0.9\linewidth, ]
{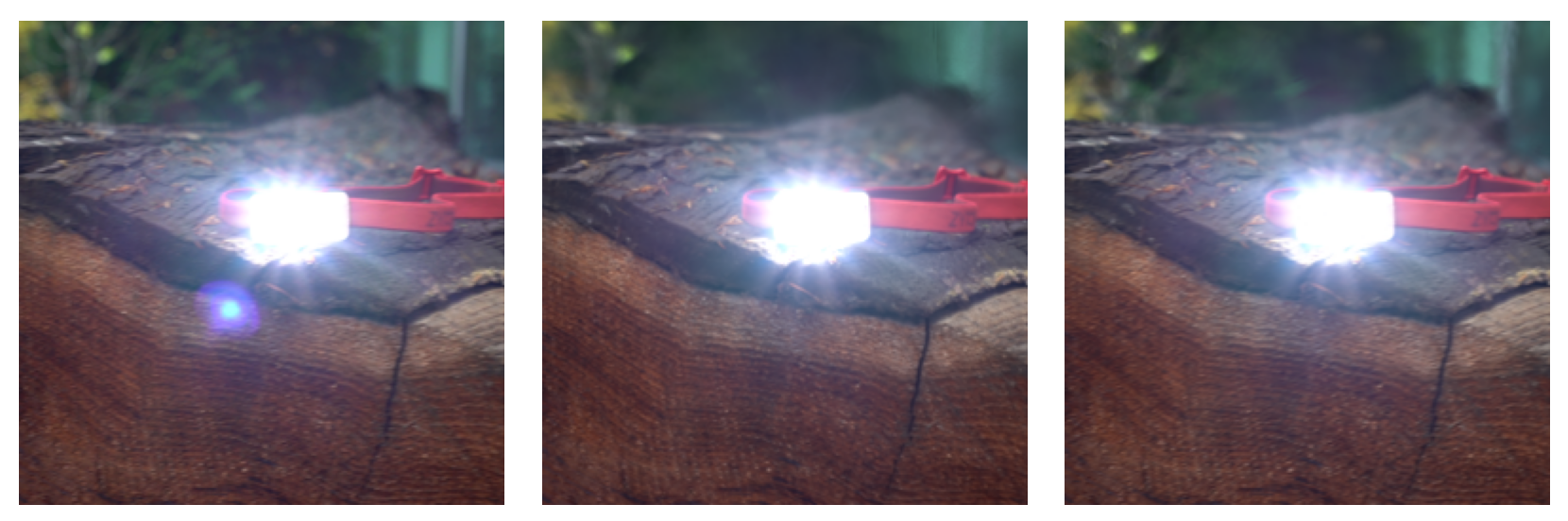}
\put(12,-3.5){Input}
\put(14,13){\color{green}\circle{8}}
\put(44,-3.5){3DGS}
\put(46,13){\color{red}\circle{8}}
\put(73,-3.5){Def-3DGS}
\put(82,13){\color{red}\circle{8}}
\end{overpic}
\captionof{figure}{Reconstructions of a 3D scene with lens flares using 3DGS~\cite{kerbl3Dgaussians}, and Deformable 3DGS~\cite{yang2023deformable3dgs}. As lens flares are not 3D-consistent, the methods struggle to reconstruct them.}
\Description{Three near-identical close-up photographs of a bright headlamp strapped to a tree trunk, labeled ``Input'', ``3DGS'', and ``Def-3DGS''. The Input photo shows a small blue circular flare artifact beside the light, marked with a green circle. In the 3DGS and Def-3DGS reconstructions, this same location is marked with a red circle, and the blue flare artifact is missing or much fainter, showing that both 3D reconstruction methods fail to reproduce it.}
\label{fig:gsreconflares}
  \end{figure}
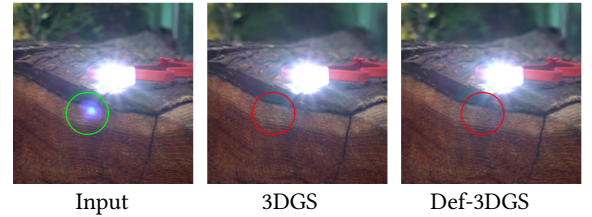

Our goal is to enable reconstruction and real-time rendering of flares from any viewpoint. Reconstructing lens flares is inherently different from reconstructing a 3D scene, for instance, using Gaussian Splatting (3DGS). In 3DGS, we assume that we can capture a static 3D scene from different viewpoints. However, this does not hold for lens flares, which are not part of the 3D scene and change with the relative motion between the light sources and the camera, i.e., they are 3D inconsistent. Even deformable 3DGS methods assume that the underlying motion is 3D consistent and fail to reconstruct reflective lens flares. Hence, when we reconstruct a scene with images containing lens flares using 3DGS, the models either bake them into the 3D scene or ignore them altogether, as shown in~\figref{fig:gsreconflares}.

Therefore, we need a novel 3D representation that can reproduce the changing lens flare across different views. To achieve this, we model lens flares using Gaussians that are \emph{anchored to the camera} on a near-plane in front of the scene, rather than as free-floating geometry in world space; this reflects the fact that flares are produced by the imaging system itself and move with the camera. Further, we design the flare representation model per light source based on the fact that lens flares approximately lie along the line joining the light source and the principal point in 2D \cite{koreban2009geometry,dai2023nighttime}. We constrain, by design, that all Gaussians lie on the projected 2D line connecting each light source to the camera's principal point. We learn a deformation field to move and modulate lens flares based on the light and camera locations. We position the plane right after the near plane and rasterize it together with the scene in a single pass, so the flare adds little to rasterization cost and the full model still renders in real time. This makes captured flares directly usable in interactive settings such as games and VR/AR, where view-consistent camera effects are needed to blend virtual content convincingly with real footage. Our representation is consistent with 3DGS rasterizers and can be ported to any of its variants.

Moreover, we collect a dataset of $9$ diverse scenes with reflective flares from multiple views to demonstrate that our lens flare model can be trained alongside 3DGS to automatically decompose a set of images into flares and scene. To better control decomposition, we supervise the 3DGS model using our removal model, enabling clean 3DGS reconstruction and lens-flare reconstruction across multiple views.

Our contributions can be summarized as follows,
\begin{itemize}
\item a flare representation model for 3D scenes using Gaussian splatting,

\item a flare removal model obtained by fine-tuning a pretrained diffusion model with real-world scattering flares and our novel dataset of large, real-world reflective flares

\item a pipeline for decomposed reconstruction of a set of multi-view images into scene Gaussians, and flare Gaussians

\item applications enabled by our flare representation, including editing and transfer of a captured flare onto novel images and 3D scenes

\end{itemize}

\section{Related Work}
\label{sec:related_work}
In this section, we discuss works most relevant to our paper. We discuss existing methods for removing flares, followed by methods for 3D reconstruction.

\begin{figure*}[ht]
    \centering
    \includegraphics[width=\linewidth]{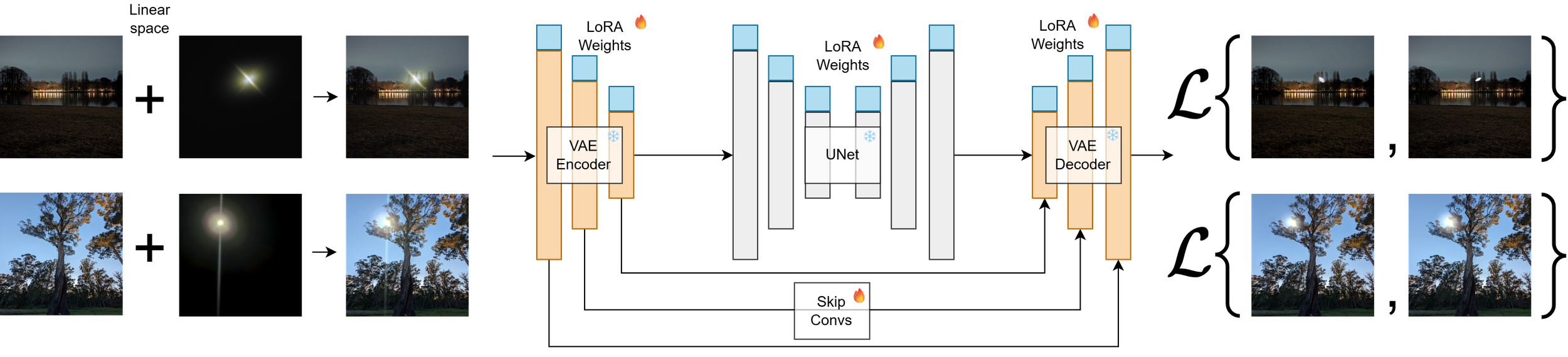}
    \caption{Our lens flare removal model (middle). We fine-tune the diffusion model from Difix3D+~\cite{wu2025difix3d} with LoRA~\cite{hu2022lora}, enabling one-step flare removal. We train two models, \ourswo and \oursw, without lens glare and with lens glare, respectively, as shown above (right, target supervision). We note that while the model with light glare might be subjectively preferred by an end user, the one without light glare is more realistic. To train our models, we add a flare image from our flare datasets to an image from the training dataset, yielding a corrupted input, and train them using the loss from ~\eqref{eq:diffloss}.}
    \label{fig:diffpipeline}
    \Description{A diagram of the lens flare removal pipeline. On the left, two example inputs are formed by adding a flare image (a starburst light) to a clean image (a lake at dusk, or a tree against the sky) in linear space, producing a flare-corrupted image. These inputs pass through a VAE encoder, a UNet, and a VAE decoder, each augmented with orange LoRA weight blocks and skip connections, to produce flare-removed outputs. On the right, the outputs are compared via a loss function against two target images: one with the light source's glare removed, and one with it retained.}
\end{figure*}
\subsection{Flare Removal}

FR ~\cite{wu2021train} is the pioneering work to train a UNet~\cite{ronneberger2015unet} to remove lens flares. Synthetic flares are added to a dataset of images, and the model is supervised with the flare-free images. Light sources are added to the generated flare-free images through threshold-based light detection. Later, many methods updated the model to UFormer~\cite{wang2022uformer}, which remains the best performer across many datasets. Flare7K~\cite{dai2022flare7k} introduced a synthetic dataset composed of scattering and reflective flares, which helped improve the flare removal models greatly. Building on this method, Flare7K++~\cite{dai2023flare7kpp} introduced a real scattering flares dataset, which further improved flare removal models. Along with the dataset, it also proposed a real-world benchmark for flare removal. However, this benchmark only tests for scattering flares and ignores lens glares, discussed later in the experiments section. BracketFlare~\cite{dai2023nighttime} is trained to remove a single small reflective flare that is symmetric around the principal point of the camera using synthetic data generation based on images captured using different exposures. On the other hand, DiffFlare~\cite{zhou2024difflare} learns to guide a pretrained multi-step latent diffusion model for scattering flare removal. ACL-FR~\cite{zhou2023improving,zhou2025improving} employs adversarial curve learning to model the image signal processing pipeline, thereby achieving improved flare removal for various camera types and enhanced preservation of lighting characteristics. FlareX~\cite{lishenqu2025lishen} proposes a novel synthetic dataset for flare removal, including some large flares, which might benefit our model. LightsOut~\cite{Tsai2025lightsout} proposes to outpaint an image to generate light sources for removing off-screen lens flares. Deng et al.~\cite{deng2024blind} estimate a flare-level map to guide blind flare removal, and Deflare-Net~\cite{ghodesawar2023deflarenet} couples flare detection with removal; neither releases code or models, so we do not compare against them quantitatively.
Contrary to existing methods, which focus on more exactly modeling the physics of flare corruption, we leverage a pretrained single-step diffusion model that enables fast flare removal by fine-tuning with LoRA~\cite{hu2022lora} on relatively small real data. Furthermore, we introduce a dataset of real full-image reflective flares and propose a reflective flare-removal benchmark that is not limited to nighttime flares.

\subsubsection{Flare removal in 3D Reconstruction.} GN-FR~\cite{matta2024gnfr} is designed to remove small scattering flares from 3D scenes using a predicted flare mask. The idea is to completely ignore the regions with flares during the supervision of a generalizable NeRF 3D reconstruction model. \textit{Note that their data and code are not publicly released}. The idea is built similarly to robust novel view synthesis methods ~\cite{sabour2025spotlessgs, zhu2023occlusion}, where distractors are detected and ignored. However, when the input contains large scattering or reflective flares, most of the information from these images is lost. We demonstrate the strength of our \textit{flare removal model} in that it consistently removes flares across multi-view images without losing scene information. We also propose a novel benchmark for evaluating 3D reconstruction and novel-view synthesis under the presence of challenging scattering flares.

\subsection{Lens Flare Representation} 
\subsubsection{Physically-based Lens Flare Representations.} Most existing methods~\cite{hullin2011phys, maurer2024capturing, bodonyi2023tiled, bodonyi2025polynomial, lee2013practicalrt} model characteristics of specific lens ensembles for producing lens flares and focus on enabling real-time rendering of the modeled lens flares. Closer to our work, Koreban and Schechner~\cite{koreban2009geometry} analyze the geometric structure of aperture-ghost flares and show that, under lens rotational symmetry, ghost positions lie on the line connecting the light source projection and the optical center in the image plane. We exploit this same geometric property as a constraint inside a differentiable 3DGS pipeline.
\subsubsection{3D Lens Flare Reconstruction} Flare representation is a hard task, as lens flares, in general, are not 3D consistent. Therefore, existing 3D Gaussian splatting~\cite{kerbl3Dgaussians} and even deformable 3DGS~\cite{luiten2023dynamic3dgs,yang2023deformable3dgs,wu20244dgs} reconstruction methods fail to represent lens flares, see~\figref{fig:gsreconflares}. However, it can be seen that scattering lens flares closer to the light source (in images) do not change significantly across views and are reconstructed near the light source in a 3D scene, even though they are not 3D-consistent. Yet, the models ignore reflective flares completely in the reconstructions and are heavily affected by scattering flares, as shown in~\figref{fig:3dgs_removed}. Our proposed reconstruction pipeline, based on our flare representation model, decomposes and represents different kinds of lens flares.

\section{Method}
\label{sec:method}
In this work, we aim to develop a lens flare removal model and use it to learn a 3D representation of lens flares. We fine-tune a latent diffusion model as our flare removal model. We elaborate on it in the first part of this section.

To represent lens flares in 3D, we adopt a dynamic Gaussian splatting model, treating them as 3D Gaussians moving on a 2D plane. While rendering the scene, we place this 2D plane in front of the scene 3DGS and render the scene and the flares together.

\subsection{Lens Flare Removal}
\label{sec:flareremoval}

Lens flares are primarily caused by unintended light paths passing through a lens when a camera observes a sufficiently bright light source. They are undesirable in many cases. Given the intractability of modeling the effects of different light sources on various lens types, we need a learning-based solution for lens flare removal. Most existing lens-flare removal models exclusively focus on scattering flares, small reflective flares, and shimmer.

\subsubsection{Image formation model}
\label{sec:forward_model}
Following prior flare-removal work~\cite{wu2021train, dai2023flare7kpp}, we adopt the standard additive linear-radiance composition,
\begin{equation}
\label{eq:forward_model}
    I = \left( I_{\text{scene}}^{\gamma} + I_{\text{flare}}^{\gamma} \right)^{1/\gamma}\,,
\end{equation}
where $I_{\text{scene}}$ is the underlying flare-free radiance, $I_{\text{flare}}$ is the additive flare contribution, and $\gamma$ approximates an inverse camera response function (CRF). \eqref{eq:forward_model} is exact under a monotone, non-saturating CRF; known deviations -- highlight clipping, ISP non-linearities, and veiling glare -- are inherited by all existing flare-removal baselines. Capturing per-pixel-paired real flare and flare-free images is impractical at scale, since it would require two perfectly co-located cameras with and without flare contributions. Hence, we (like prior work) train on synthetic composites obtained by blending real flare layers onto clean images using \eqref{eq:forward_model}, inheriting the additivity assumption. The additive model is used only for this synthesis; it plays no role in the 3D flare representation.

\subsubsection{Diffusion-based removal model}
We introduce a unified flare removal model that works with different types of lens flares - scattering, and reflective (both full image and small). We fine-tune a latent diffusion model for this task. Specifically, we adopt the diffusion model in Difix3D+~\cite{wu2025difix3d}. The model is trained on a dataset of billions of images, making it well-suited for generating flare-free images. Furthermore, the fine-tuning paradigm in Difix3D+, based on single-step image-to-image translation models, is ideally suited to our task.

The model consists of three parts: a variational autoencoder (VAE) with an encoder ($\mathcal{E}(I;\theta_e)$) and decoder ($\mathcal{D}(z;\theta_d)$), a text conditioning ($txt$), and a latent diffusion UNet model ($e(z;\theta,txt)$) that denoises a noise-corrupted version of the latent code from VAE's encoder, based on text conditioning, see ~\figref{fig:diffpipeline}. Unlike Difix3D+, however, we do not use a reference-view conditioning. We refer the reader to the Difix3D+ paper for more details.

We fine-tune the models using LoRA adapters~\cite{hu2022lora} with a rank of $16$. We fine-tune the VAE encoder and decoder, as well as the UNet diffusion model. We deviate from Difix3D's training and even fine-tune the encoder, as the flare-corrupted images are most likely out of distribution for the VAE (see ablations).

Let $I_i$ be the input image corrupted with a flare. The encoder $\mathcal{E}(\theta_{le})$ encodes the image into a latent $z = \mathcal{E}(I_i;\theta_{le},\theta_e)$. The latent is corrupted with noise (corresponding to the time step $t=199$) and denoised based on a text conditioning to obtain the edited latent $z_e =e(\alpha_t z + \sigma_t \epsilon;\theta_{l},\theta,txt)$, for some noise $\epsilon$. Finally, the edited latent code is decoded as the output image $I_o = \mathcal{D}(z_e;\theta_{ld},\theta_d)$. Here, $\theta_{lx}$ are the LoRA weights and $\theta_x$ are the pretrained weights. We fine-tune the models using the following loss function,
\begin{equation}
\label{eq:diffloss}
    \mathcal{L}_{\text{diff}}(\theta_{le},\theta_l,\theta_{ld}) = \| I_g - I_o \|_2^2
    + \text{LPIPS}(I_g,I_o) \,,
\end{equation}

where $I_g$ is a given flare free image.
\subsubsection{Training data}
Obtaining real-world paired data for flare removal is challenging, as it requires two identical cameras at the exact same locations: one capturing an image with flares and one without. Therefore, similar to baselines~\cite{dai2023flare7kpp, wu2021train, zhou2025improving}, we obtain paired data by adding flare images from our lens flare datasets (both synthetic and real) to an image dataset, Flickr24K~\cite{zhang2018single} in the linear space. Let $I_{image}$ be an image from the image dataset, we corrupt it with flare using $I_i = (I_{image}^\gamma + I_f^\gamma)^{1/\gamma}$ (see~\eqref{eq:forward_model}, and \figref{fig:diffpipeline}), where $I_f$ is the flare image and $\gamma \in [1.8,2.2]$ is a scalar. Further, we add the expected light source from the dataset to the supervision image, $I_g = (I_{image}^\gamma + I_{light}^\gamma)^{1/\gamma}$. The light source is only added for supervising scattering flare removal from the Flare7K++ dataset, and not for our VFX benchmark. We found that this improves our diffusion model's performance (see ablations `wo light' vs ours).

\subsection{Lens Flare Representation and Reconstruction}
\label{sec:flarerep}
Representing lens flares using physically based models across different lens types is challenging, as different lens ensembles produce distinct flares under varying light sources, necessitating accurate physical models of those sources in real-world scenes. Further, accurately modeling the effect of the lenses requires not only precise prescriptions for all lens elements but also modeling of coherent propagation effects at the wavelength scale to precisely reproduce lens flares, with corresponding geometric detail for all components in the lens, particularly parts like iris blades that produce diffraction.
Therefore, we adopt Gaussian splatting to represent and reconstruct lens flares in a given scene.

\begin{figure}[t]
    \includegraphics[width=\linewidth]{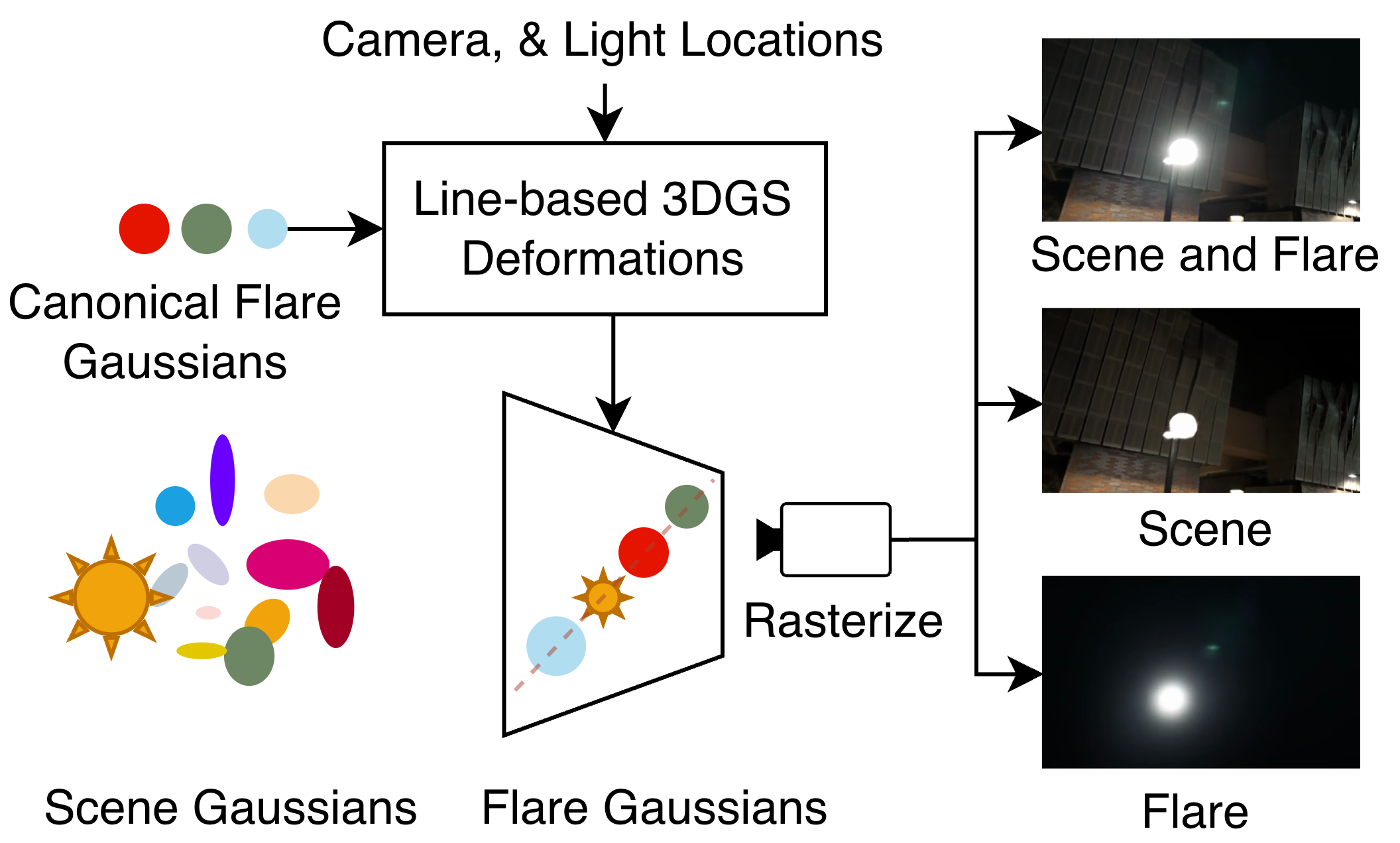}
    \caption{Our lens flare representation model (left, top) along with a 3DGS model(left, bottom). We represent flares as Gaussians with 1D locations. Given input camera and light positions, we deform the Gaussians and place them along a line connecting the image's principal point to the light. We render the scene and flare Gaussians together (right, top) and supervise them using the provided scene images. We render the scene using Gaussians (right, middle) and supervise using images from our flare-removal model.}
    \label{fig:repPipeline}
    \Description{A diagram illustrating the flare representation pipeline. On the left, colored 2D Gaussian ellipses represent flare components and scene Gaussians, arranged along and around a diagonal line on a camera-frustum plane labeled ``Line-based 3DGS Deformations''. On the right, three photographs of a streetlamp scene at night show the rendered composite of scene and flare Gaussians (top), the scene rendered alone (middle), and a faint flare-only render (bottom).}

\end{figure}

To achieve this, we model lens flares as 3D Gaussians on a plane in front of the near plane, and move them based on the camera location. We render the scene together with the Gaussians. Furthermore, we model the Gaussians representing flares as lying on a line between the image's principal point and the light source's position in the image. We illustrate this in ~\figref{fig:repPipeline}. The flare model is jointly optimized along with a Gaussian splatting~\cite{kerbl3Dgaussians} model representing the scene.

\subsubsection{Gaussian splatting}  Our goal is to represent lens flares with minimum overhead over the scene representation - 3DGS. Therefore, we use 3D Gaussians as the final output of our lens flare representation, allowing us to rasterize the scene along with the lens flare effects. 
In 3DGS~\cite{kerbl3Dgaussians}, the scene is modeled as a collection of 3D Gaussians $\mathcal{G}$, endowed with a mean $\mu$, scale $s$, covariance $\Sigma$, opacity $\sigma$, and color as spherical harmonics with $b$-bases $sh$; $\mathcal{G} = \{ \mu, s, \Sigma, \sigma, sh \}$. The Gaussians are initialized with a point cloud from Colmap~\cite{schoenberger2016mvs,schoenberger2016sfm}. The Gaussians are rasterized to input camera locations,  and the parameters are optimized using a loss function that compares the input images with the rasterized images. We refer the readers to 3DGS ~\cite{kerbl3Dgaussians} for a more elaborate explanation.

\subsubsection{Flare Representation}
\label{sec:flare_representation}

\begin{figure}[t]
    \centering
    \includegraphics[width=\linewidth]{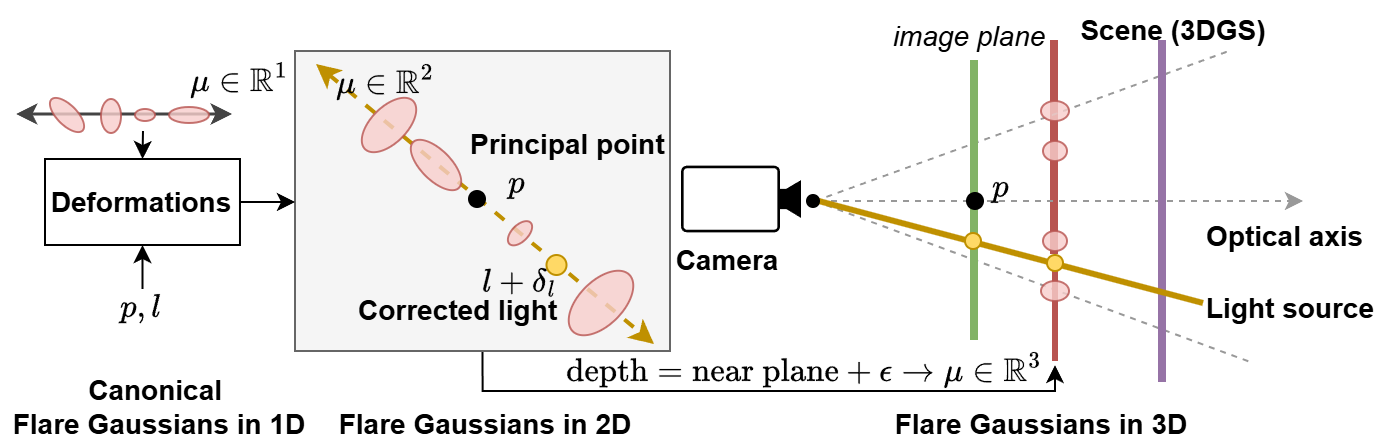}
    \caption{Geometric construction of a single flare. Each flare component is represented by a set of canonical 1D Gaussians with scalar mean $\mu \in \mathbb{R}$, encoding the radial offset along the image-plane line through the principal point $p$ and the (corrected) light projection $l+\delta_l$. At render time, the scalar $\mu$ is mapped to a 2D image point via $\mu'\big((l+\delta_l)-p\big)+p$, then unprojected one near-plane offset in front of the scene to obtain a standard 3DGS primitive in the camera frame.}
    \label{fig:flare_geometry}
    \Description{A diagram showing three stages of the flare geometry construction, left to right. First, ``Canonical Flare Gaussians in 1D'': a row of pink ellipses of varying size and orientation arranged along a single axis, with scalar mean mu, fed into a ``Deformations'' box conditioned on principal point p and light position l. Second, ``Flare Gaussians in 2D'': the deformed ellipses arranged along a diagonal dashed line inside a square, connecting a black dot labeled ``Principal point p'' to a yellow dot labeled ``Corrected light l plus delta-l''. Third, ``Flare Gaussians in 3D'': a camera icon projects the same line through an image plane onto a near-plane offset in front of a scene panel, placing the flare Gaussians as standard 3DGS primitives along the optical axis between the camera and the light source.}
\end{figure}

Our representation places flares on the image-plane line through the principal point $p$ and the projected light location $l$~\cite{koreban2009geometry,dai2023nighttime}, and parameterizes each flare component by its radial position along that line. Concretely, each flare component is a set of \emph{canonical 1D Gaussians} whose scalar mean $\mu\in\mathbb{R}$ encodes radial offset. A small MLP, conditioned on the camera and light positions, deforms these canonical Gaussians; the deformed scalar mean $\mu'$ is then (i) lifted to a 2D image point via $\mu'\big((l+\delta_l)-p\big)+p$, (ii) unprojected to a depth equal to the near plane $+\epsilon$ in the camera frame, and (iii) emitted as a standard 3DGS primitive (\figref{fig:flare_geometry}). Because the entire construction lives in the camera frame and produces standard 3DGS primitives, the scene rasterizer is reused unchanged, and the flare component is rigidly tied to the camera, independent of the scene geometry behind it. The remainder of this section details the parameterization, deformation MLP, and light-position correction.

Our lens flare model is inspired by the Deformable 3DGS~\cite{yang2023deformable3dgs} (Def-3DGS). Our model features a set of canonical Gaussians representing lens flares, along with a deformation model that deforms them based on the light location and camera position. Unlike in Def-3DGS, however, the Gaussians in our lens flare model are one-dimensional ($\mu \in \mathbb{R}^1$, with a slight abuse of notation), as the location of the lens flares is generally restricted to the line connecting the principal point to the 2D position of the light source in image space~\cite{koreban2009geometry,dai2023nighttime}. While Def-3DGS models condition on time, we condition our deformation model on the camera and light locations to modulate the canonical flare Gaussians, since lens flares depend on the relative positions of the camera and light.

Therefore, each of our flare Gaussians has the parameters $\mathcal{G} = \{ \mu, s, \Sigma, \sigma, sh \}$, with $\mu \in \mathbb{R}^1$. We employ a hash-grid MLP from Instant-NGP~\cite{mueller2022instant} per parameter to deform the Gaussians based on the camera view and the 2D light location ($l$),
\begin{equation}
\label{eq:deformation_mlp}
\delta_x = f_x(\mu,c,l;\theta_{hx})\,,
\end{equation}

where $c$ is the camera location, $x$ are the different Gaussian parameters, and $\theta_{hx}$ are the hash-grid MLP parameters. For a given camera location, we get the deformed flare Gaussians,
\begin{equation}
\label{eq:gaussian_deformation}
x' = x + \delta_x, \qquad \forall\, x \in \{\mu, s, \Sigma, \sigma, sh\}\,.
\end{equation}
Further, to allow for some asymmetries and off-axis flares, we allow the Gaussians to deform from the line equation by adding a 2D deformation to the Gaussians on the 2D plane. We learn these `finer deformations' as
\begin{equation}
    \label{eq:finer_deformations}
    \delta_f = f_f(\mu,c,l; \theta_{hf}) \,,
\end{equation}
consistent with the notation from above. These deformations help represent asymmetries in lens flares; see `Ours wo $\delta_f$' vs `Ours' in the ablations (~\secref{sec:ablations}).

Additionally, to account for inconsistencies in 2D light locations at the pixel level, we learn a 2D deformation of the light location based on the camera location as,
\begin{equation}
    \label{eq:light_pos_correction}
    \delta_l = f_c(c,l; \theta_{hl}) \,.
\end{equation} We transform the 1D $\mu'$ to a 2D location using $\mu' ((l+\delta_l)-p) + p$, where $p$ is the principal point. We transform the Gaussians to the current camera's coordinate system, unprojecting the 2D locations to a depth equal to the near plane depth $+\epsilon$. Finally, we concatenate the scene and flare Gaussians and rasterize them jointly with the 3DGS rasterizer to obtain the composited image.

\begin{figure}[ht]
    \centering
    \input{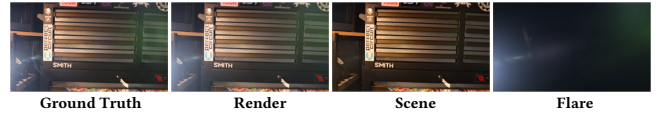}
    \vspace{-4mm}
    \caption{\emph{Workshop} test view: the generating light source lies outside the image, yet our reconstructed flare combined with the scene reproduces the ground truth. See~\secref{sec:flarereconstruction} for further experimental details and results.}
    \label{fig:offscreen}
    \Description{A strip of four images of a workshop tool cabinet labeled ``Ground Truth'', ``Render'', ``Scene'', and ``Flare''. Ground Truth and Render show near-identical photographs of the cabinet with a faint blue-green flare glow entering from the left edge. The Scene image shows the cabinet without any flare, appearing cleaner and slightly darker on the left. The Flare image is a mostly dark frame showing only the isolated bluish-green flare glow, with the light source itself outside the visible frame.}
\end{figure}

Moreover, for a scene with multiple light sources, we instantiate one flare Gaussian model per light source (see the \emph{workshop} row of~\figref{fig:flare_modeling_comparison}, which has three light sources, two shown in the image). Furthermore, our model only requires the 3D position of each light source: we project it into every training view to obtain $l$, regardless of whether the source itself is actually visible in that view. As long as a light source has positive depth in the camera frame, we enable its flare model. This means off-screen light sources (whose projected $l$ falls outside the image extent) and occluded light sources (e.g., reflective glare where the source is hidden behind scene geometry) are handled identically to on-screen, unoccluded ones -- visibility of the source is not required by our model, only its image-plane projection (see~\figref{fig:offscreen}).

\subsubsection{Light Positions}  \label{sec:light_position}
In an ideal lens-flare model, every Gaussian in the scene contributes to the lens flares and is attenuated proportionally to its intensity. However, even with reliable Gaussian intensities, instantiating the ideal flare model for every Gaussian in the scene -- rather than for a small, sparse set of discrete emitters -- is computationally wasteful. Further, an ideal camera sensor that captures the full dynamic range would make it
trivial to identify light sources directly from the Gaussians themselves, e.g., by
thresholding on intensity. But captured images have limited dynamic range, so the intensities recovered by a 3DGS reconstruction are an unreliable proxy for true radiance. Light sources are precisely the regions most prone to over-exposure and clipping.

Therefore, we approximate the flare contribution of the highest-intensity sources by estimating a discrete set of source centers (i.e., light sources). This means our model takes as input the number of light sources in the scene and their locations, which -- not their intensities -- drive the flare reconstruction (\secref{sec:flare_representation}).

There are many ways to obtain this information, such as manual annotation,
heuristic/algorithmic thresholding, as in earlier flare-removal
works~\cite{zhou2023improving,zhou2025improving}, open-vocabulary detection, or our own representation model applied iteratively. We adopt an off-the-shelf
open-vocabulary detection pipeline for its simplicity. Our ablations
(\secref{sec:ablations}, `wo-G-SAM2') show that the model degrades gracefully rather than failing outright when light detections are missing entirely, and it remains robust when detections are imperfect but present (\secref{sec:light_robustness}). A good light-position initialization sharpens flare placement and orientation, but is not strictly required for a plausible reconstruction.

We obtain locations and the number of light sources using Grounded-SAM2~\cite{ren2024grounded}
and TRASE~\cite{li2026trase}. Grounded-SAM2 has two stages: GroundingDINO~\cite{liu2023grounding} produces
bounding boxes, which are then input to SAM2~\cite{ravi2024sam2segmentimages} for masks. We first obtain all bounding-box detections for the text prompt, ``light source'', from GroundingDINO,
and pass them to SAM2 to obtain a mask per box. We then apply two refinement steps
to suppress spurious detections. First, since light sources are typically
over-exposed, we discard any detection whose mask is not over-exposed, i.e. whose
mean brightness (the HSV value channel) falls below $190$ out of $255$; this removes
detections on non-emissive objects. Second, when multiple bounding boxes overlap on
the same source (high intersection relative to the smaller box, or one box contained
in another), we keep the box with the \emph{smallest} area, as it most tightly
localizes the emitter and avoids region-level boxes that merge a source with its
surroundings. We measure the reliability of Grounded-SAM2 in the supplemental Sec.~A.3.

Note that, even after refinement, these stages give noisy results --
missing some light sources or producing spurious ones. Therefore, we aggregate this
2D information in 3D using TRASE~\cite{li2026trase} to obtain Gaussians corresponding
to the light sources. We calculate the centroids of the Gaussians representing each
light source in the scene by clustering them. We utilize these 3D light locations
and project them onto the images to obtain the 2D light positions for training our
model.

\subsubsection{Reconstruction} We closely follow 3DGS training to train our scene models. However, since our goal is decomposable reconstruction, we render a scene-only image $I_s$ with just the scene Gaussians for scene supervision, and separately obtain the composited image $I$ by concatenating the scene and flare Gaussians and rasterizing them jointly. We minimize the following loss during each iteration.

\begin{equation}
\label{eq:flarereploss}
\begin{split}
    \mathcal{L} = & \lambda \| I_s-I_o \|_1 + (1-\lambda )\big(1-\text{SSIM}(I_s,I_o)\big) \\
                 + & \lambda \| I-I_{input} \|_1 + (1-\lambda )\big(1-\text{SSIM}(I,I_{input})\big)\,,
\end{split}
\end{equation}

where $I_o$ is the flare-removed image using our flare removal model (see ~\secref{sec:flareremoval}), and $I_{input}$ is the given image. We set $\lambda = 0.9$.

\section{Datasets}
\label{sec:datasets}
\subsection{Flare removal} We train our flare removal model with the Flare7k++~\cite{dai2023flare7kpp} and the Flickr24K~\cite{zhang2018single} datasets. However, the flare dataset is limited to small reflective flares. Therefore, we compile a large reflective flares dataset from publicly available real visual effects videos of large flares~\cite{ActionVFX_LensFlares_2026}. We obtain $1866$ training images from $90$ clips spanning three effect categories (halo, icon and shimmer), and hold out $18$ further clips entirely for testing, so that no test flare shares a source clip with any training flare. We compile the VFX benchmark by compositing flares from these held-out clips onto $128$ images from our captures.

To quantify the differences between Flare7K++'s and our VFX flares, we measure, for each isolated flare layer, the fraction of the frame that contains $90\%$ of the flare's energy, a threshold- and resolution-independent measure of spatial extent. Flare7K++'s reflective flares have a median coverage of $0.39\%$ (interquartile range (IQR) $0.13$--$1.81\%$, $N=2000$), whereas the flares in our VFX dataset have a median of $78.0\%$ (IQR $69.1$--$82.5\%$, $N=1866$); the IQRs do not overlap. Large full-frame reflective flares are therefore absent from existing data rather than merely under-represented. The two sets also differ in provenance. Flare7K++'s reflective flares are synthetic, generated from $10$ base patterns; its real world subset, captured on the rear cameras of three smartphones, contains scattering flares only~\cite{dai2023flare7kpp}. Ours are real captures of large reflective flares. Lens diversity is therefore not a meaningful axis of comparison here. Synthetic flares have no associated lens, and the visual-effects footage does not publish one, so we characterize the two sets by flare extent instead.

\begin{figure}[ht]
    \centering
    \begin{overpic}[width=0.8\linewidth]{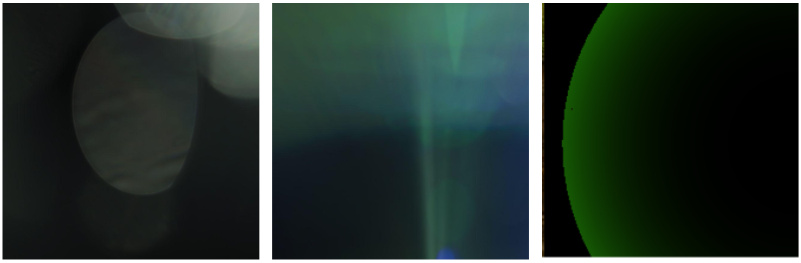}
    \put(13,-4){\small Full image real flares}
    \put(73,-4) {\small Ring flares}
    \end{overpic}
    \captionof{figure}{Examples of real large lens flares from our dataset compiled using videos from~\cite{ActionVFX_LensFlares_2026} (left, middle), and procedural ring-shaped reflective flares (right).}
    \label{fig:flareexamples}
    \Description{Three dark photographs of large reflective lens flares. The first two, under ``Full image real flares'', show a soft glowing oval shape filling most of a dark frame and a vertical greenish-blue streak with a bright spot at its base. The third, under ``Ring flares'', shows a procedurally generated green ring-shaped flare with opacity fading from the ring toward the center of the image.}
\end{figure}

Moreover, we identified that some lenses produce large, almost ring-shaped reflective lens flares that cover the entire image and are not captured by any of these datasets. Therefore, we procedurally create large circular flares with radially diminishing opacity towards the center. We achieve this by randomly sampling a center for the circle in the image and then a radius (a multiple of the minimum of the image's height and width). We draw this circle, choosing a random color and opacity ($<0.7$). Finally, we multiply the image by a radially increasing opacity mask, which is $0$ at the center. Therefore, obtaining a ring-like flare. See~\figref{fig:flareexamples} for examples.

\subsubsection{Scenes with Flares} Existing 3D scene datasets are carefully captured to not have any light sources. Therefore, we capture two datasets to demonstrate the effectiveness of our flare removal model across multiple views and that of our flare representation model.

\subsubsection{Flare Removal from 3D Scenes} We capture a set of $5$ diverse indoor and outdoor scenes with different types of light sources and lens flares in the scenes. The scenes consist of $50$--$110$ images, of which we reserve $10\%$ for testing. For a fair evaluation, we deliberately select test views that are flare-free, captured from positions where the light source falls outside the camera frustum or is sufficiently occluded; these views are held out of training and serve as unbiased scene-only ground truth for evaluating flare removal across multiple views, independent of any removal model's output. See the supplement for examples of our test and training images.

\subsection{Flare Reconstruction} \label{sec:flare_rec_dataset} We capture $9$ diverse scenes spanning a mix of indoor and outdoor settings, multiple lens types (varying in focal length and ensemble), and a variety of real-world light sources -- an LED headlamp, workshop lighting, a smartphone flashlight, a street lamp, and a purple-hued decorative light -- so as to elicit a broad range of reflective flare patterns. We capture $\sim 1$ min sequences, use $300$--$400$ evenly spaced video frames for training, and $40$ images from the remaining frames for testing. Held-out views far from the training trajectory are impractical here: we captured each sequence only for as long as the flare remained visible, so camera motion was bounded by the point at which the light left the frame and the flare vanished, rather than by any preference for a narrow sweep. These held-out frames are therefore interleaved with the training frames, so they test interpolation between nearby viewpoints within that range; extrapolation to new viewpoints and scenes is shown through transfer (\secref{sec:transfer}) and in the supplementary webpage. Unlike ordinary scene content, flare appearance is governed by the light's position relative to the optical axis rather than by scene-point parallax, and this mapping is highly nonlinear: small camera or light motions can produce disproportionately large changes in ghost position, halo extent, and streak orientation (\secref{sec:light_robustness}). Interpolating this deformation between nearby frames is therefore non-trivial. Unlike the $5$-scene removal dataset above, a per-pixel ground-truth \emph{flare-free} image and \emph{flare layer} are intractable to obtain in real captures, since the scene and flare radiance are entangled at the sensor; separating them optically would require specialized capture hardware that is impractical for handheld video. We therefore evaluate decomposition against a model-derived reference, which we define and justify in \secref{sec:flarereconstruction}.

\section{Experiments}
In this section, we first present a comparison with existing works on the Flare7K++ and our VFX benchmarks. Later, we present ablation experiments that inform our design choices. Finally, we evaluate our flare reconstruction model and show ablation experiments for the design choices of the representation model and the reconstruction pipeline. We show additional qualitative results and \textit{provide implementation details in the supplement}.
\label{sec:experiments}

\begin{table*}
\centering
\caption{Quantitative comparisons (left) and ablations (right) on flare removal. On each benchmark, one of our two models (\oursw and \ourswo, with and without glare) outperforms all baselines on every metric. LPIPS uses the VGG backbone. Since Flare7k++ contains residual light glare, \oursw performs best. Our models train in $20$ hours, whereas baselines take $>4$ days. Ablations: Performance on the Flare7k++ benchmark degrades without light sources in target images or without VAE encoder fine-tuning (since flare-corrupted images are OOD). Performance on the VFX benchmark degrades without VFX training data, which, in exchange, slightly improves Flare7k++ PSNR and LPIPS. Best results among the compared methods are shown in \textcolor{red}{\textbf{red bold}}, second-best in \textcolor{blue}{\uline{blue underline}}; ablations are not ranked.
}

\label{tab:flareremoval}
\resizebox{\linewidth}{!}{
\begin{tabular}{l|l||c|c|c|c|c|c|c||c|c|c}
\multicolumn{9}{c||}{} &	\multicolumn{3}{c}{Ablations on \oursw} \\
\cline{10-12}
\textbf{Benchmark} & \textbf{Metric} & \textbf{LightsOut} & \textbf{FR} & \textbf{Flare7K++} & \textbf{ACL-FR} & \textbf{FlareX} & \textbf{\oursw} & \textbf{\ourswo} & \textbf{wo} & \textbf{wo} & \textbf{wo} \\
& & ~\cite{Tsai2025lightsout} & ~\cite{wu2021train} & ~\cite{dai2023flare7kpp} & ~\cite{zhou2025improving} & ~\cite{lishenqu2025lishen} & & & \textbf{light} & \textbf{enc FT} & \textbf{VFX}  \\
\hline

\multirow{3}{*}{Flare7k++} & PSNR $\uparrow$ & 16.04 & 24.49 & \textcolor{blue}{\uline{26.42}} & 24.48 & 25.26 & \textcolor{red}{\textbf{26.68}} & 25.76 & 25.84 & 25.64 & 26.96 \\
 & SSIM $\uparrow$ & 0.6954 & 0.8797 & \textcolor{blue}{\uline{0.8926}} & 0.8689 & 0.8870 & \textcolor{red}{\textbf{0.8993}} & 0.8925 & 0.8831 & 0.8696 & 0.8850 \\
 & LPIPS $\downarrow$ & 0.2686 & 0.0980 & 0.0908 & 0.0925 & 0.0934 & \textcolor{red}{\textbf{0.0777}} & \textcolor{blue}{\uline{0.0848}} & 0.0850 & 0.0856 & 0.0767 \\
\hline
\multirow{3}{*}{vfx\_dataset} & PSNR $\uparrow$ & 13.04 & 21.04 & 21.37 & 20.20 & 24.17 & \textcolor{blue}{\uline{25.93}} & \textcolor{red}{\textbf{27.06}} & 25.03 & 23.75 & 24.77 \\
 & SSIM $\uparrow$ & 0.3360 & 0.9268 & 0.9192 & 0.8862 & \textcolor{blue}{\uline{0.9377}} & 0.9366 & \textcolor{red}{\textbf{0.9409}} & 0.9329 & 0.9157 & 0.9295 \\
 & LPIPS $\downarrow$ & 0.4889 & 0.0968 & 0.0852 & 0.0920 & 0.0704 & \textcolor{blue}{\uline{0.0671}} & \textcolor{red}{\textbf{0.0659}} & 0.0715 & 0.0799 & 0.0877 \\
\hline
\end{tabular}

}
\end{table*}

\subsection{Removing Lens Flares}

In this section, we evaluate the performance of our lens flare removal model. First, we compare our model against established baselines on the Flare7K++~\cite{dai2023flare7kpp} benchmark for scattering flare removal. Then, we compare our model against baselines on our proposed VFX benchmark. One of the most significant benefits of our method is that it requires only $20$ hours to fine-tune our model (on a single GPU), whereas our baselines take more than $4$ days to train.

\subsubsection{Scattering Flares} 
\begin{figure*}
    \centering
    \begin{overpic}[width=\linewidth]{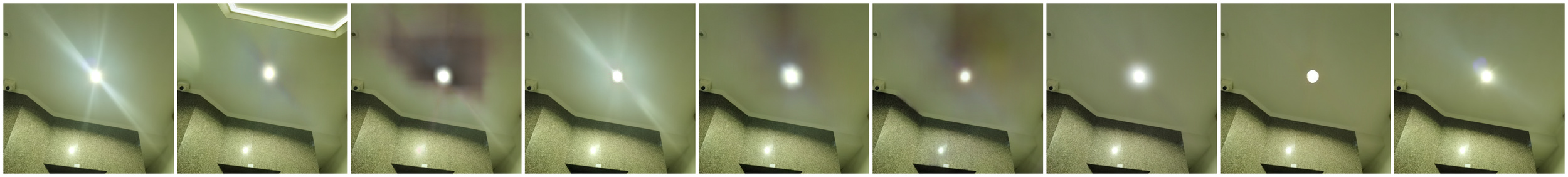}

    \put(5.65, 11.5){\makebox[0pt]{\scriptsize \textbf{Input}}}
    \put(16.74,11.5){\makebox[0pt]{\scriptsize \textbf{LightsOut}}}
    \put(27.82,11.5){\makebox[0pt]{\scriptsize \textbf{FR}}}
    \put(38.91,11.5){\makebox[0pt]{\scriptsize \textbf{ACL-FR}}}
    \put(50.00,11.5){\makebox[0pt]{\scriptsize \textbf{Flare7K++}}}
    \put(61.09,11.5){\makebox[0pt]{\scriptsize \textbf{FlareX}}}
    \put(72.18,11.5){\makebox[0pt]{\scriptsize \textbf{\oursw}}}
    \put(83.26,11.5){\makebox[0pt]{\scriptsize \textbf{\ourswo}}}
    \put(94.35,11.5){\makebox[0pt]{\scriptsize \textbf{Ground Truth}}}
    \put(16.74,1.0){\makebox[0pt][c]{\scriptsize \colorbox{black}{\textcolor{white}{PSNR: 19.57}}}}
    \put(27.82,1.0){\makebox[0pt][c]{\scriptsize \colorbox{black}{\textcolor{white}{PSNR: 21.13}}}}
    \put(38.91,1.0){\makebox[0pt][c]{\scriptsize \colorbox{black}{\textcolor{white}{PSNR: 22.62}}}}
    \put(50.00,1.0){\makebox[0pt][c]{\scriptsize \colorbox{black}{\textcolor{white}{PSNR: 25.34}}}}
    \put(61.09,1.0){\makebox[0pt][c]{\scriptsize \colorbox{black}{\textcolor{white}{PSNR: 21.96}}}}
    \put(72.18,1.0){\makebox[0pt][c]{\scriptsize \colorbox{psnrgreen}{\textcolor{white}{PSNR: 30.08}}}}
    \put(83.26,1.0){\makebox[0pt][c]{\scriptsize \colorbox{black}{\textcolor{white}{PSNR: 27.24}}}}
\end{overpic}

\begin{overpic}[width=\linewidth]{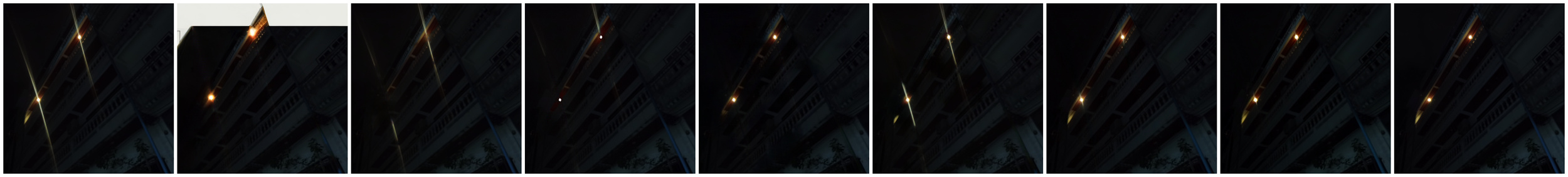}
    \put(16.74,1.0){\makebox[0pt][c]{\scriptsize \colorbox{black}{\textcolor{white}{PSNR: 9.98}}}}
    \put(27.82,1.0){\makebox[0pt][c]{\scriptsize \colorbox{black}{\textcolor{white}{PSNR: 29.40}}}}
    \put(38.91,1.0){\makebox[0pt][c]{\scriptsize \colorbox{black}{\textcolor{white}{PSNR: 31.58}}}}
    \put(50.00,1.0){\makebox[0pt][c]{\scriptsize \colorbox{black}{\textcolor{white}{PSNR: 33.54}}}}
    \put(61.09,1.0){\makebox[0pt][c]{\scriptsize \colorbox{black}{\textcolor{white}{PSNR: 29.77}}}}
    \put(72.18,1.0){\makebox[0pt][c]{\scriptsize \colorbox{black}{\textcolor{white}{PSNR: 38.12}}}}
    \put(83.26,1.0){\makebox[0pt][c]{\scriptsize \colorbox{psnrgreen}{\textcolor{white}{PSNR: 38.29}}}}
\end{overpic}

\begin{overpic}[width=\linewidth]{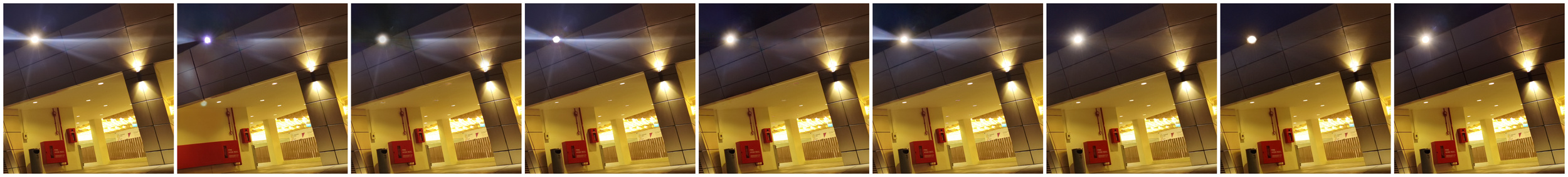}
\put(-2,0) {\rotatebox{90}{Flare7K++}}
    \put(16.74,1.0){\makebox[0pt][c]{\scriptsize \colorbox{black}{\textcolor{white}{PSNR: 20.38}}}}
    \put(27.82,1.0){\makebox[0pt][c]{\scriptsize \colorbox{black}{\textcolor{white}{PSNR: 22.87}}}}
    \put(38.91,1.0){\makebox[0pt][c]{\scriptsize \colorbox{black}{\textcolor{white}{PSNR: 21.26}}}}
    \put(50.00,1.0){\makebox[0pt][c]{\scriptsize \colorbox{black}{\textcolor{white}{PSNR: 26.37}}}}
    \put(61.09,1.0){\makebox[0pt][c]{\scriptsize \colorbox{black}{\textcolor{white}{PSNR: 22.65}}}}
    \put(72.18,1.0){\makebox[0pt][c]{\scriptsize \colorbox{psnrgreen}{\textcolor{white}{PSNR: 29.18}}}}
    \put(83.26,1.0){\makebox[0pt][c]{\scriptsize \colorbox{black}{\textcolor{white}{PSNR: 25.69}}}}
\end{overpic}

\begin{overpic}[width=\linewidth]{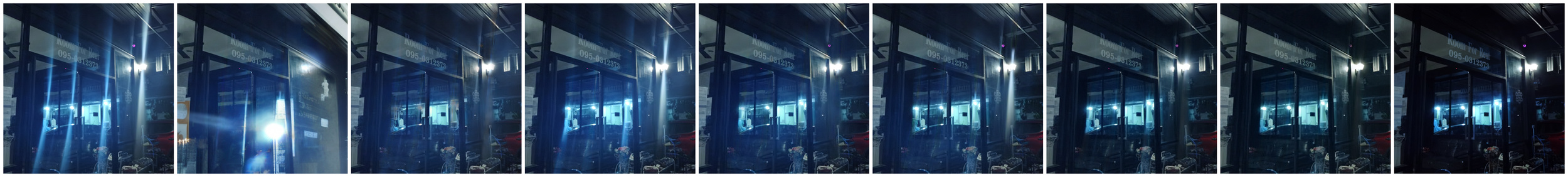}
    \put(16.74,1.0){\makebox[0pt][c]{\scriptsize \colorbox{black}{\textcolor{white}{PSNR: 10.81}}}}
    \put(27.82,1.0){\makebox[0pt][c]{\scriptsize \colorbox{black}{\textcolor{white}{PSNR: 16.89}}}}
    \put(38.91,1.0){\makebox[0pt][c]{\scriptsize \colorbox{black}{\textcolor{white}{PSNR: 14.84}}}}
    \put(50.00,1.0){\makebox[0pt][c]{\scriptsize \colorbox{black}{\textcolor{white}{PSNR: 18.15}}}}
    \put(61.09,1.0){\makebox[0pt][c]{\scriptsize \colorbox{black}{\textcolor{white}{PSNR: 17.60}}}}
    \put(72.18,1.0){\makebox[0pt][c]{\scriptsize \colorbox{black}{\textcolor{white}{PSNR: 22.72}}}}
    \put(83.26,1.0){\makebox[0pt][c]{\scriptsize \colorbox{psnrgreen}{\textcolor{white}{PSNR: 23.65}}}}
\end{overpic}

\begin{overpic}[width=\linewidth]{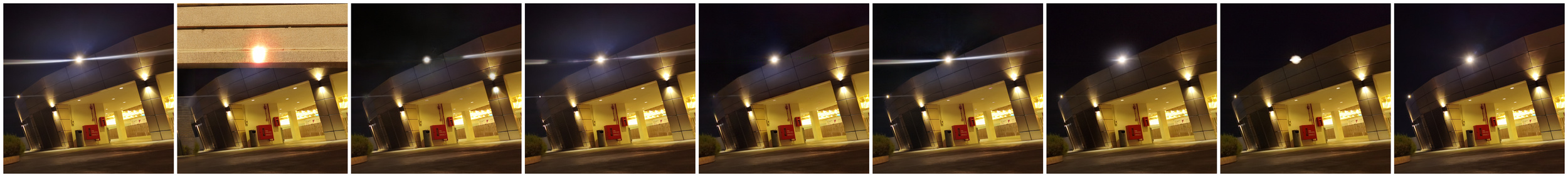}
    \put(16.74,1.0){\makebox[0pt][c]{\scriptsize \colorbox{black}{\textcolor{white}{PSNR: 10.26}}}}
    \put(27.82,1.0){\makebox[0pt][c]{\scriptsize \colorbox{black}{\textcolor{white}{PSNR: 24.96}}}}
    \put(38.91,1.0){\makebox[0pt][c]{\scriptsize \colorbox{black}{\textcolor{white}{PSNR: 23.80}}}}
    \put(50.00,1.0){\makebox[0pt][c]{\scriptsize \colorbox{black}{\textcolor{white}{PSNR: 29.42}}}}
    \put(61.09,1.0){\makebox[0pt][c]{\scriptsize \colorbox{black}{\textcolor{white}{PSNR: 23.53}}}}
    \put(72.18,1.0){\makebox[0pt][c]{\scriptsize \colorbox{psnrgreen}{\textcolor{white}{PSNR: 30.58}}}}
    \put(83.26,1.0){\makebox[0pt][c]{\scriptsize \colorbox{black}{\textcolor{white}{PSNR: 27.45}}}}
\end{overpic}
    \begin{overpic}[width=\linewidth]{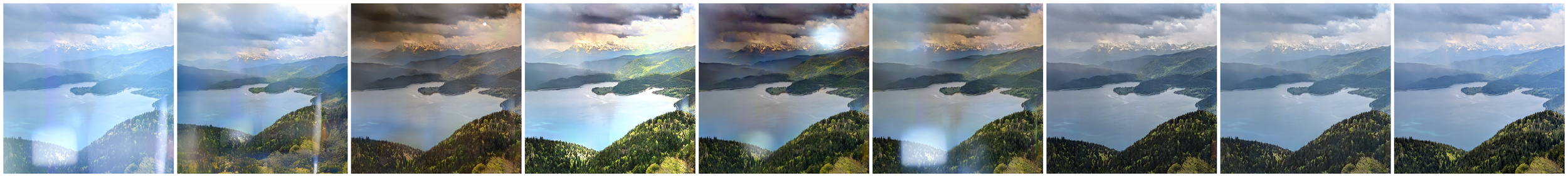}
    \put(5.65,98.12){\makebox[0pt]{\scriptsize \textbf{Input}}}
    \put(16.74,98.12){\makebox[0pt]{\scriptsize \textbf{LightsOut}}}
    \put(16.74,1.0){\makebox[0pt][c]{\scriptsize \colorbox{black}{\textcolor{white}{PSNR: 15.01}}}}
    \put(27.82,98.12){\makebox[0pt]{\scriptsize \textbf{FR}}}
    \put(27.82,1.0){\makebox[0pt][c]{\scriptsize \colorbox{black}{\textcolor{white}{PSNR: 12.75}}}}
    \put(38.91,98.12){\makebox[0pt]{\scriptsize \textbf{ACL-FR}}}
    \put(38.91,1.0){\makebox[0pt][c]{\scriptsize \colorbox{black}{\textcolor{white}{PSNR: 16.73}}}}
    \put(50.00,98.12){\makebox[0pt]{\scriptsize \textbf{flare7kpp}}}
    \put(50.00,1.0){\makebox[0pt][c]{\scriptsize \colorbox{black}{\textcolor{white}{PSNR: 12.26}}}}
    \put(61.09,98.12){\makebox[0pt]{\scriptsize \textbf{FlareX}}}
    \put(61.09,1.0){\makebox[0pt][c]{\scriptsize \colorbox{black}{\textcolor{white}{PSNR: 13.55}}}}
    \put(72.18,98.12){\makebox[0pt]{\scriptsize \textbf{oursw}}}
    \put(72.18,1.0){\makebox[0pt][c]{\scriptsize \colorbox{black}{\textcolor{white}{PSNR: 18.19}}}}
    \put(83.26,98.12){\makebox[0pt]{\scriptsize \textbf{ours}}}
    \put(83.26,1.0){\makebox[0pt][c]{\scriptsize \colorbox{psnrgreen}{\textcolor{white}{PSNR: 24.55}}}}
    \put(94.35,98.12){\makebox[0pt]{\scriptsize \textbf{Ground Truth}}}
\end{overpic}

\begin{overpic}[width=\linewidth]{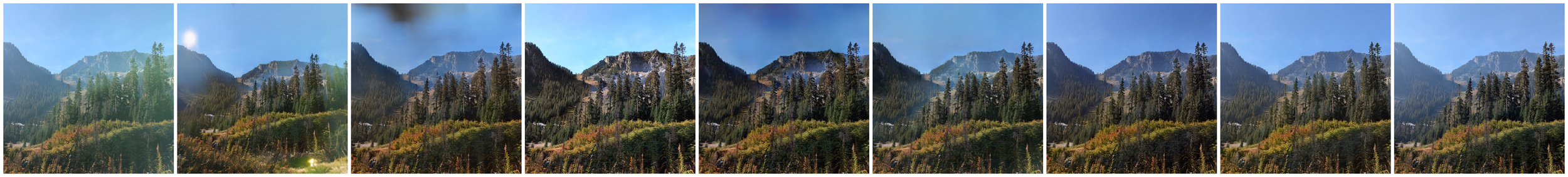}
    \put(16.74,1.0){\makebox[0pt][c]{\scriptsize \colorbox{black}{\textcolor{white}{PSNR: 15.38}}}}
    \put(27.82,1.0){\makebox[0pt][c]{\scriptsize \colorbox{black}{\textcolor{white}{PSNR: 18.77}}}}
    \put(38.91,1.0){\makebox[0pt][c]{\scriptsize \colorbox{black}{\textcolor{white}{PSNR: 18.58}}}}
    \put(50.00,1.0){\makebox[0pt][c]{\scriptsize \colorbox{black}{\textcolor{white}{PSNR: 15.70}}}}
    \put(61.09,1.0){\makebox[0pt][c]{\scriptsize \colorbox{black}{\textcolor{white}{PSNR: 20.11}}}}
    \put(72.18,1.0){\makebox[0pt][c]{\scriptsize \colorbox{black}{\textcolor{white}{PSNR: 20.41}}}}
    \put(83.26,1.0){\makebox[0pt][c]{\scriptsize \colorbox{psnrgreen}{\textcolor{white}{PSNR: 25.15}}}}
\end{overpic}

\begin{overpic}[width=\linewidth]{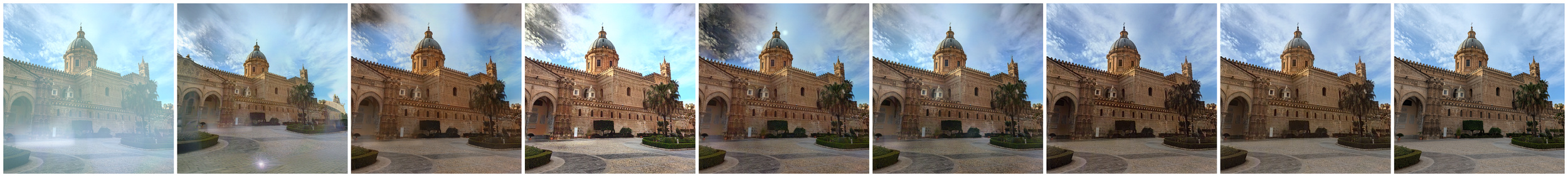}
    \put(-2,0) {\rotatebox{90}{VFX benchmark}}
    \put(16.74,1.0){\makebox[0pt][c]{\scriptsize \colorbox{black}{\textcolor{white}{PSNR: 13.93}}}}
    \put(27.82,1.0){\makebox[0pt][c]{\scriptsize \colorbox{black}{\textcolor{white}{PSNR: 18.78}}}}
    \put(38.91,1.0){\makebox[0pt][c]{\scriptsize \colorbox{black}{\textcolor{white}{PSNR: 17.83}}}}
    \put(50.00,1.0){\makebox[0pt][c]{\scriptsize \colorbox{black}{\textcolor{white}{PSNR: 16.27}}}}
    \put(61.09,1.0){\makebox[0pt][c]{\scriptsize \colorbox{black}{\textcolor{white}{PSNR: 22.08}}}}
    \put(72.18,1.0){\makebox[0pt][c]{\scriptsize \colorbox{black}{\textcolor{white}{PSNR: 22.70}}}}
    \put(83.26,1.0){\makebox[0pt][c]{\scriptsize \colorbox{psnrgreen}{\textcolor{white}{PSNR: 27.11}}}}
\end{overpic}

\begin{overpic}[width=\linewidth]{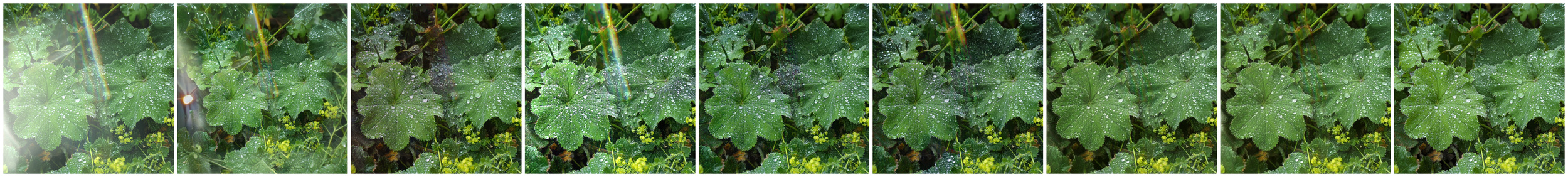}
    \put(16.74,1.0){\makebox[0pt][c]{\scriptsize \colorbox{black}{\textcolor{white}{PSNR: 13.16}}}}
    \put(27.82,1.0){\makebox[0pt][c]{\scriptsize \colorbox{black}{\textcolor{white}{PSNR: 24.30}}}}
    \put(38.91,1.0){\makebox[0pt][c]{\scriptsize \colorbox{black}{\textcolor{white}{PSNR: 19.97}}}}
    \put(50.00,1.0){\makebox[0pt][c]{\scriptsize \colorbox{black}{\textcolor{white}{PSNR: 25.05}}}}
    \put(61.09,1.0){\makebox[0pt][c]{\scriptsize \colorbox{black}{\textcolor{white}{PSNR: 21.82}}}}
    \put(72.18,1.0){\makebox[0pt][c]{\scriptsize \colorbox{black}{\textcolor{white}{PSNR: 25.28}}}}
    \put(83.26,1.0){\makebox[0pt][c]{\scriptsize \colorbox{psnrgreen}{\textcolor{white}{PSNR: 26.17}}}}
\end{overpic}

\begin{overpic}[width=\linewidth]{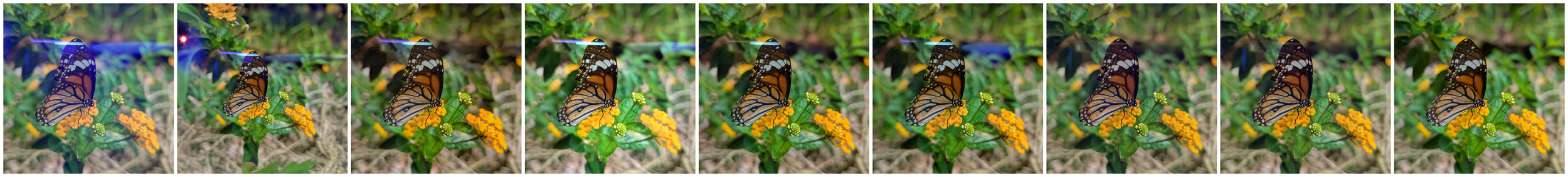}
    \put(16.74,1.0){\makebox[0pt][c]{\scriptsize \colorbox{black}{\textcolor{white}{PSNR: 12.98}}}}
    \put(27.82,1.0){\makebox[0pt][c]{\scriptsize \colorbox{black}{\textcolor{white}{PSNR: 22.78}}}}
    \put(38.91,1.0){\makebox[0pt][c]{\scriptsize \colorbox{black}{\textcolor{white}{PSNR: 19.30}}}}
    \put(50.00,1.0){\makebox[0pt][c]{\scriptsize \colorbox{black}{\textcolor{white}{PSNR: 26.93}}}}
    \put(61.09,1.0){\makebox[0pt][c]{\scriptsize \colorbox{black}{\textcolor{white}{PSNR: 22.87}}}}
    \put(72.18,1.0){\makebox[0pt][c]{\scriptsize \colorbox{black}{\textcolor{white}{PSNR: 26.20}}}}
    \put(83.26,1.0){\makebox[0pt][c]{\scriptsize \colorbox{psnrgreen}{\textcolor{white}{PSNR: 27.16}}}}
\end{overpic}
    \caption{ Qualitative comparison of different flare removal models, from left to right, given image, LightsOut~\cite{Tsai2025lightsout}, FR~\cite{wu2021train}, ACL-FR~\cite{zhou2025improving}, Flare7K++~\cite{dai2023flare7kpp}, FlareX~\cite{lishenqu2025lishen}, \oursw, \ourswo, and ground truth. In the top five rows, we show the results on the Flare7K++ benchmark. As shown, our models consistently remove flares without residual flares and more effectively recover the light source in the image, leading to improved quantitative performance. In the last five rows, we show the results from our VFX benchmark. As shown, existing models mistakenly bake large flare corruptions into the sky, whereas our model performs much better at removing them.}
    \label{fig:qualitative_flareremoval}
    \Description{A grid of ten image strips, each row showing the same scene processed by different flare-removal methods side by side: the flare-corrupted input, then LightsOut, FR, ACL-FR, Flare7K++, FlareX, our model with lens glare, our model without lens glare, and the ground truth, with a PSNR value overlaid on each result. The top five rows are scenes from the Flare7K++ benchmark with small scattering flares; baseline outputs retain visible residual glare and streaks, while our two models' outputs closely resemble the flare-free ground truth. The bottom five rows are scenes from the VFX benchmark with large reflective flares covering much of the sky; baseline methods leave large corrupted regions, while our models recover plausible sky and scene detail.}
\end{figure*}
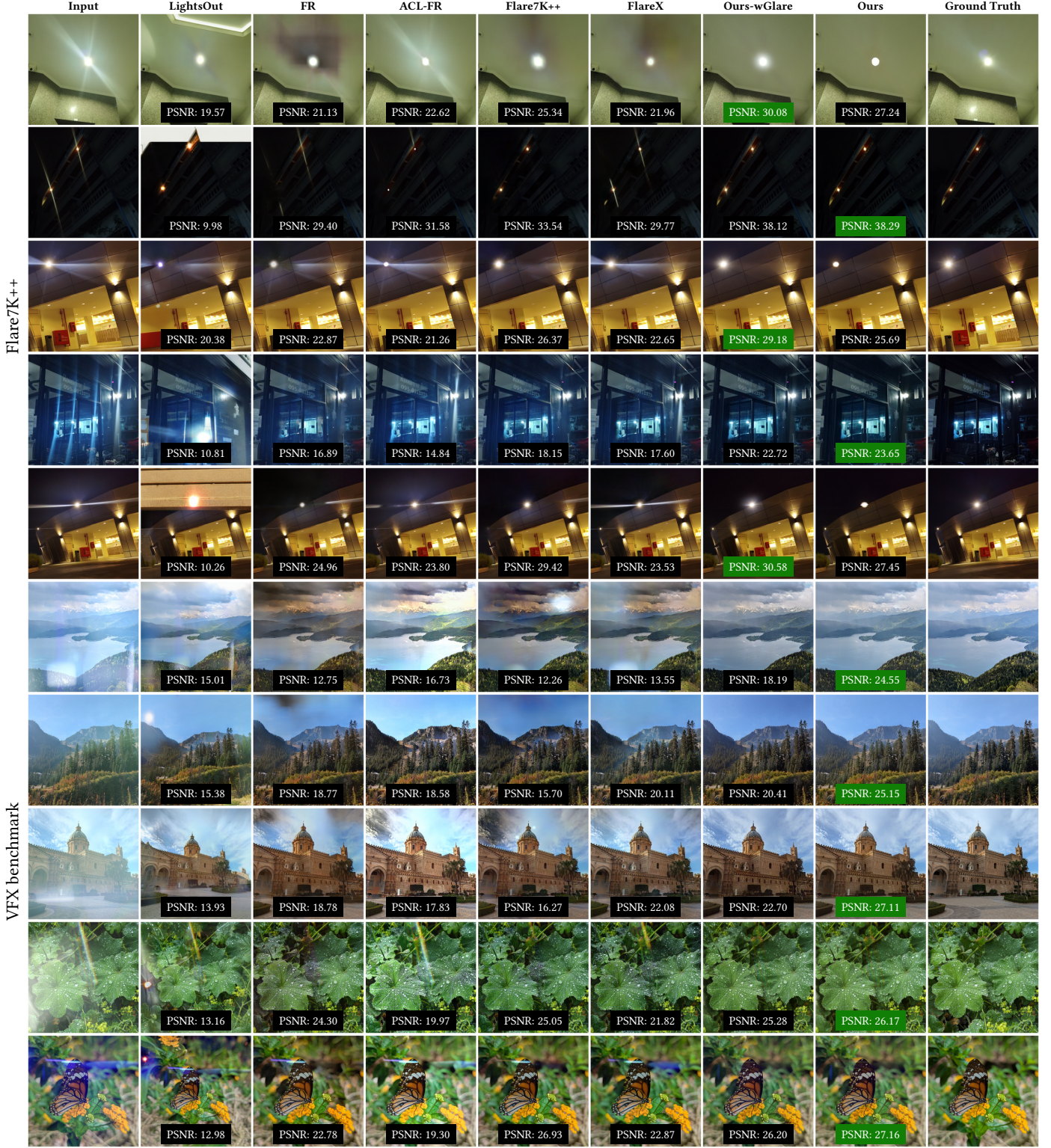

The Flare7K++ benchmark mainly evaluates scattering flare removal. We compare our model with the top-performing scattering-flare-removal models: Flare7K++~\cite{dai2023flare7kpp}, FR~\cite{wu2021train}, and ACL-FR~\cite{zhou2025improving} by retraining the models with our VFX dataset and procedural flares in addition to the Flare7K++ dataset. Further, we compare with the released versions of LightsOut~\cite{Tsai2025lightsout} and FlareX~\cite{lishenqu2025lishen}; note that LightsOut is designed to outpaint off-screen light sources ahead of an existing removal network, whereas we evaluate it as a standalone remover. Quantitative results for the comparison are shown in~\tabref{tab:flareremoval}. As shown, our model, which incorporates supervision of light sources and lens glare, outperforms all existing models on the benchmark, which also contains lens glare. %
As shown in~\figref{fig:qualitative_flareremoval}, our models better recover the scene behind the flares and generate more plausible images. Furthermore, note that our model, trained without lens glares, also performs very well qualitatively in removing flares. Our baselines, however, still contain a substantial residual flare.

\subsubsection{Reflective Flares}
Removing large reflective flares is particularly difficult, as most of the details that help a model recover the scene's exact features are obscured by them. We train our models on our proposed flare-removal dataset. We show the quantitative results in~\tabref{tab:flareremoval}. We evaluate the models' performance on $128$ images of our proposed VFX benchmark. \ourswo\ outperforms all baselines on every metric, and \oursw\ on PSNR and LPIPS.

\subsubsection{Ablations}
\begin{figure*}
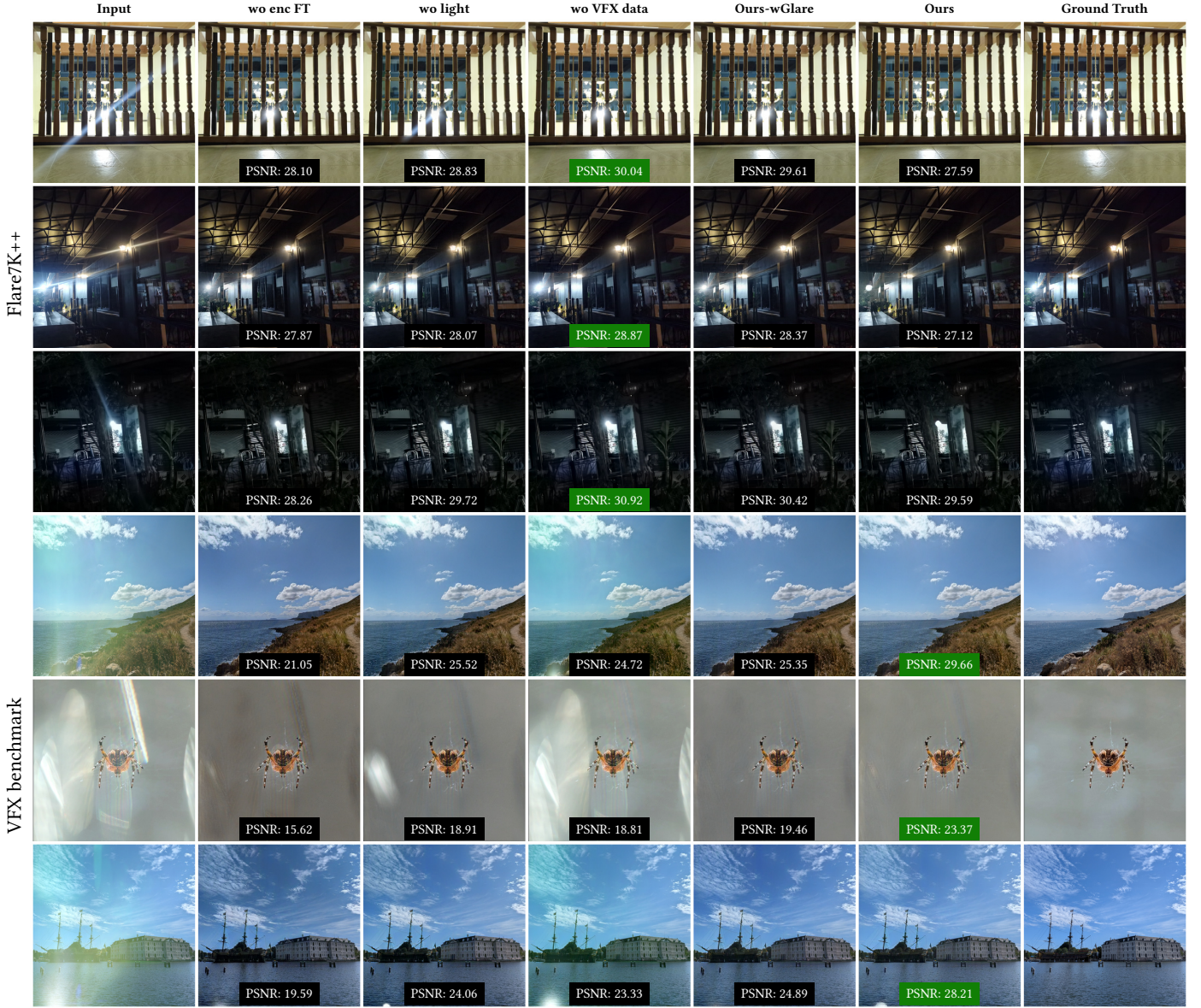

    \centering

    \vspace{5mm}
    \input{figures/fr_abs/flare7kpp_strips/latex_codes}
    \input{figures/fr_abs/vfx_strips/latex_codes}
    \caption{Qualitative comparisons of ablations of our flare removal model, \oursw, on Flare7K++ benchmark (top), and VFX benchmark (bottom). Left to right, we show the input, `wo enc FT', where we do not finetune the VAE encoder, `wo light', where a light source is added after flare removal,  `wo VFX data', where the model is not trained on the VFX dataset, and `\oursw' and `\ourswo'. As shown, without encoder finetuning, the model struggles to recover color characteristics. Post-processing the output to add a light source does not accurately reproduce it. Without the VFX dataset, the model struggles to remove large flares. Finally, \oursw and \ourswo models perform the best on both the benchmarks.}
    \label{fig:ablations_quals}
    \Description{Two rows of image strips comparing ablations of the flare removal model on a Flare7K++ scene (top) and a VFX scene (bottom). Each row shows, left to right: the flare-corrupted input, `wo enc FT' (VAE encoder not fine-tuned, showing color shifts), `wo light' (light source added back after removal, appearing less accurate), `wo VFX data' (large flares only partially removed), the full model with lens glare, and the full model without lens glare. The two rightmost results in each row most closely match a flare-free, correctly colored scene.}
\end{figure*}

We evaluate our design choices of our flare removal model qualitatively in~\figref{fig:ablations_quals}, and quantitatively in~\tabref{tab:flareremoval}. One key design choice is to incorporate light sources into the supervision images, eliminating the need for a post-processing step to add light back to the flare-removed images, as in our baselines. We demonstrate this in `wo light', where we train the model to remove light sources and then add them back as a post-processing step. While baseline models, especially Flare7K++~\cite{dai2023flare7kpp}, perform well when a light source is added back to the images, we observed that our diffusion-based model does not perform as well on either benchmark (\tabref{tab:flareremoval}, `wo light').
Another important design choice is to fine-tune the VAE encoder with LoRA. We demonstrate the importance of this in `wo enc FT', where we do not fine-tune the VAE encoder. Quantitatively, our models outperform the model without encoder finetuning. We believe this is due to the distribution of the flare-corrupted images not being close to the VAE's training distribution. Finally, training without our VFX data specializes the model to scattering flares: it gains $0.28$\,dB on Flare7K++ ($26.96$ vs.\ $26.68$\,dB) but loses $1.16$\,dB on VFX ($24.77$ vs.\ $25.93$\,dB). While it still performs well at removing large reflective flares, it misses important flare regions and leaves flare artifacts, as shown in~\figref{fig:ablations_quals}.

\begin{table*}
\centering
\small
\caption{ Quantitative comparison of performance of different flare removal models on 3D reconstruction. We evaluate the flare removal performance across multiple views of FR~\cite{wu2021train}, Flare7k++~\cite{dai2023flare7kpp}, ACL-FR~\cite{zhou2025improving}, FlareX~\cite{lishenqu2025lishen}, \ourswo and \oursw by training a 3DGS model on the different sets of flare-removed images. Additionally, we compare with BilateralGrid~\cite{wang2024bilateral} and vanilla 3DGS~\cite{kerbl3Dgaussians}. While the models are trained on flare-removed images, they are evaluated on actual captured images of the scenes without processing, which happen to not contain flares (see~\secref{sec:lensflaresremoval3d}). Both our models outperform all baselines on average; on plant, \oursw falls below four of the six baselines on every metric, while \ourswo remains best. Best results are shown in \textcolor{red}{\textbf{red bold}}, second-best in \textcolor{blue}{\uline{blue underline}}.
}
\label{tab:3dgs_removed}
\resizebox{\textwidth}{!}{

\setlength{\tabcolsep}{4pt}
\begin{tabular}{l|ccc|ccc|ccc|ccc|ccc|ccc|ccc|ccc}
\textbf{Scene} & \multicolumn{3}{c|}{\textbf{Vanilla 3DGS}} & \multicolumn{3}{c|}{\textbf{w BilateralGrid}} & \multicolumn{3}{c|}{\textbf{w Flare7k++}} & \multicolumn{3}{c|}{\textbf{w FR}} & \multicolumn{3}{c|}{\textbf{w ACL-FR}} & \multicolumn{3}{c|}{\textbf{w FlareX}} & \multicolumn{3}{c|}{\textbf{w \oursw}} & \multicolumn{3}{c}{\textbf{w \ourswo}} \\
  & PSNR$\uparrow$ & SSIM$\uparrow$ & LPIPS$\downarrow$ & PSNR$\uparrow$ & SSIM$\uparrow$ & LPIPS$\downarrow$ & PSNR$\uparrow$ & SSIM$\uparrow$ & LPIPS$\downarrow$ & PSNR$\uparrow$ & SSIM$\uparrow$ & LPIPS$\downarrow$ & PSNR$\uparrow$ & SSIM$\uparrow$ & LPIPS$\downarrow$ & PSNR$\uparrow$ & SSIM$\uparrow$ & LPIPS$\downarrow$ & PSNR$\uparrow$ & SSIM$\uparrow$ & LPIPS$\downarrow$ & PSNR$\uparrow$ & SSIM$\uparrow$ & LPIPS$\downarrow$  \\
\hline
amp & 12.25 & 0.518 & 0.599 & 14.40 & 0.637 & 0.413 & 17.24 & 0.714 & 0.435 & 18.50 & 0.689 & 0.450 & 16.75 & 0.669 & 0.455 & 17.03 & 0.681 & 0.440 & \textcolor{blue}{\uline{19.84}} & \textcolor{blue}{\uline{0.743}} & \textcolor{blue}{\uline{0.379}} & \textcolor{red}{\textbf{21.06}} & \textcolor{red}{\textbf{0.751}} & \textcolor{red}{\textbf{0.377}} \\
billboard & 18.93 & 0.645 & 0.327 & 19.59 & \textcolor{blue}{\uline{0.707}} & \textcolor{blue}{\uline{0.273}} & 19.16 & 0.656 & 0.345 & 19.84 & 0.686 & 0.319 & 18.61 & 0.668 & 0.336 & 19.43 & 0.680 & 0.301 & \textcolor{red}{\textbf{20.67}} & \textcolor{red}{\textbf{0.740}} & \textcolor{red}{\textbf{0.256}} & \textcolor{blue}{\uline{20.47}} & 0.705 & 0.286 \\
dog & 21.77 & 0.869 & 0.185 & 21.74 & 0.903 & 0.146 & 27.22 & 0.931 & 0.118 & 25.88 & 0.917 & 0.149 & 25.01 & 0.888 & 0.173 & 25.44 & 0.916 & 0.135 & \textcolor{blue}{\uline{28.42}} & \textcolor{red}{\textbf{0.945}} & \textcolor{blue}{\uline{0.090}} & \textcolor{red}{\textbf{28.48}} & \textcolor{blue}{\uline{0.943}} & \textcolor{red}{\textbf{0.089}} \\
guitar & 20.07 & 0.835 & 0.255 & 20.03 & 0.866 & 0.242 & 24.53 & 0.903 & 0.171 & 25.65 & 0.905 & 0.173 & 22.64 & 0.870 & 0.214 & 24.97 & 0.900 & 0.176 & \textcolor{red}{\textbf{26.21}} & \textcolor{red}{\textbf{0.911}} & \textcolor{red}{\textbf{0.151}} & \textcolor{blue}{\uline{25.72}} & \textcolor{blue}{\uline{0.910}} & \textcolor{blue}{\uline{0.162}} \\
plant & 20.36 & 0.836 & 0.226 & 19.05 & 0.820 & 0.266 & \textcolor{blue}{\uline{23.83}} & \textcolor{blue}{\uline{0.872}} & \textcolor{blue}{\uline{0.192}} & 23.40 & 0.848 & 0.218 & 22.53 & 0.848 & 0.215 & 21.32 & 0.815 & 0.258 & 21.06 & 0.831 & 0.252 & \textcolor{red}{\textbf{23.91}} & \textcolor{red}{\textbf{0.874}} & \textcolor{red}{\textbf{0.186}} \\
\hline
\textbf{Avg.} & 18.68 & 0.741 & 0.318 & 18.96 & 0.787 & 0.268 & 22.40 & 0.815 & 0.252 & 22.65 & 0.809 & 0.262 & 21.11 & 0.789 & 0.279 & 21.64 & 0.799 & 0.262 & \textcolor{blue}{\uline{23.24}} & \textcolor{blue}{\uline{0.834}} & \textcolor{blue}{\uline{0.226}} & \textcolor{red}{\textbf{23.93}} & \textcolor{red}{\textbf{0.837}} & \textcolor{red}{\textbf{0.220}} \\
\end{tabular}
}
\end{table*}

\begin{figure*}

    \vspace{5mm}
    \centering
    \begin{overpic}[width=\linewidth]{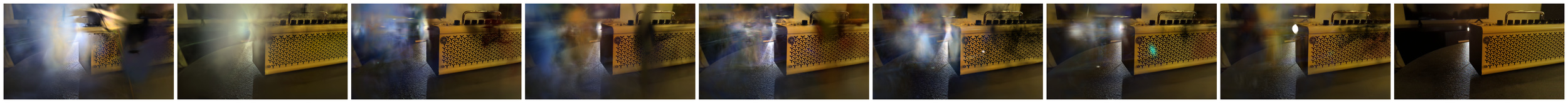}
    \put(-1.6,2) {\rotatebox{90}{amp}}
    \put(3,7){Vanilla}
    \put(12,7) {BilateralGrid}
    \put(24.5,7){Flare7k++}
    \put(38,7){FR}
    \put(47,7){ACL-FR}
    \put(58,7){FlareX}
    \put(68,7){\oursw}
    \put(81.5,7){\ourswo}
    \put(89.5,7){Ground Truth}
    \put(5.66,96.55){\makebox[0pt]{\tiny \textbf{Vanilla 3DGS}}}
    \put(5.66,0.9){\makebox[0pt][c]{\tiny \colorbox{black}{\textcolor{white}{PSNR: 9.69}}}}
    \put(16.74,96.55){\makebox[0pt]{\tiny \textbf{BilGrid}}}
    \put(16.74,0.9){\makebox[0pt][c]{\tiny \colorbox{black}{\textcolor{white}{PSNR: 13.29}}}}
    \put(27.83,96.55){\makebox[0pt]{\tiny \textbf{Flare7K}}}
    \put(27.83,0.9){\makebox[0pt][c]{\tiny \colorbox{black}{\textcolor{white}{PSNR: 14.53}}}}
    \put(38.91,96.55){\makebox[0pt]{\tiny \textbf{Google}}}
    \put(38.91,0.9){\makebox[0pt][c]{\tiny \colorbox{black}{\textcolor{white}{PSNR: 15.75}}}}
    \put(50.00,96.55){\makebox[0pt]{\tiny \textbf{Improving}}}
    \put(50.00,0.9){\makebox[0pt][c]{\tiny \colorbox{black}{\textcolor{white}{PSNR: 15.21}}}}
    \put(61.09,96.55){\makebox[0pt]{\tiny \textbf{FlareX}}}
    \put(61.09,0.9){\makebox[0pt][c]{\tiny \colorbox{black}{\textcolor{white}{PSNR: 13.61}}}}
    \put(72.17,96.55){\makebox[0pt]{\tiny \textbf{Ours (LB)}}}
    \put(72.17,0.9){\makebox[0pt][c]{\tiny \colorbox{black}{\textcolor{white}{PSNR: 17.44}}}}
    \put(83.26,96.55){\makebox[0pt]{\tiny \textbf{Ours (Sat)}}}
    \put(83.26,0.9){\makebox[0pt][c]{\tiny \colorbox{psnrgreen}{\textcolor{white}{PSNR: 18.62}}}}
    \put(94.34,96.55){\makebox[0pt]{\tiny \textbf{Ground Truth}}}
\end{overpic}

\begin{overpic}[width=\linewidth]{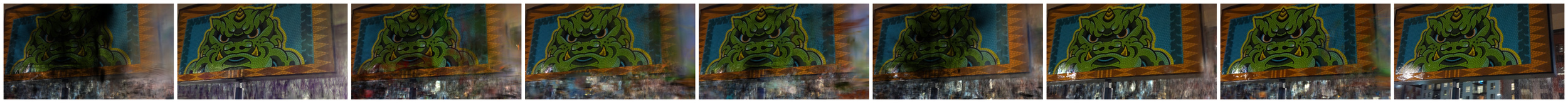}
    \put(-2,0) {\rotatebox{90}{billboard}}
    \put(5.66,0.9){\makebox[0pt][c]{\tiny \colorbox{black}{\textcolor{white}{PSNR: 15.52}}}}
    \put(16.74,0.9){\makebox[0pt][c]{\tiny \colorbox{black}{\textcolor{white}{PSNR: 17.71}}}}
    \put(27.83,0.9){\makebox[0pt][c]{\tiny \colorbox{black}{\textcolor{white}{PSNR: 17.04}}}}
    \put(38.91,0.9){\makebox[0pt][c]{\tiny \colorbox{black}{\textcolor{white}{PSNR: 18.31}}}}
    \put(50.00,0.9){\makebox[0pt][c]{\tiny \colorbox{black}{\textcolor{white}{PSNR: 17.47}}}}
    \put(61.09,0.9){\makebox[0pt][c]{\tiny \colorbox{black}{\textcolor{white}{PSNR: 17.21}}}}
    \put(72.17,0.9){\makebox[0pt][c]{\tiny \colorbox{psnrgreen}{\textcolor{white}{PSNR: 20.93}}}}
    \put(83.26,0.9){\makebox[0pt][c]{\tiny \colorbox{black}{\textcolor{white}{PSNR: 20.27}}}}
\end{overpic}

\begin{overpic}[width=\linewidth]{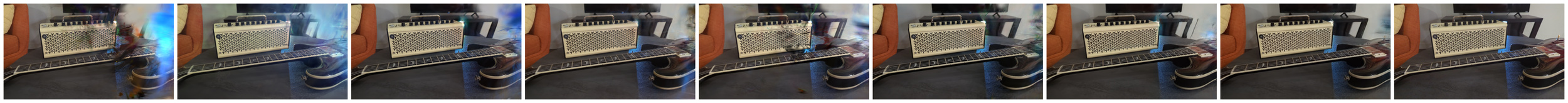}
    \put(-2,1) {\rotatebox{90}{guitar}}
    \put(5.66,0.9){\makebox[0pt][c]{\tiny \colorbox{black}{\textcolor{white}{PSNR: 15.92}}}}
    \put(16.74,0.9){\makebox[0pt][c]{\tiny \colorbox{black}{\textcolor{white}{PSNR: 18.93}}}}
    \put(27.83,0.9){\makebox[0pt][c]{\tiny \colorbox{black}{\textcolor{white}{PSNR: 21.73}}}}
    \put(38.91,0.9){\makebox[0pt][c]{\tiny \colorbox{black}{\textcolor{white}{PSNR: 24.21}}}}
    \put(50.00,0.9){\makebox[0pt][c]{\tiny \colorbox{black}{\textcolor{white}{PSNR: 21.18}}}}
    \put(61.09,0.9){\makebox[0pt][c]{\tiny \colorbox{black}{\textcolor{white}{PSNR: 24.27}}}}
    \put(72.17,0.9){\makebox[0pt][c]{\tiny \colorbox{psnrgreen}{\textcolor{white}{PSNR: 26.27}}}}
    \put(83.26,0.9){\makebox[0pt][c]{\tiny \colorbox{black}{\textcolor{white}{PSNR: 25.47}}}}
\end{overpic}

\begin{overpic}[width=\linewidth]{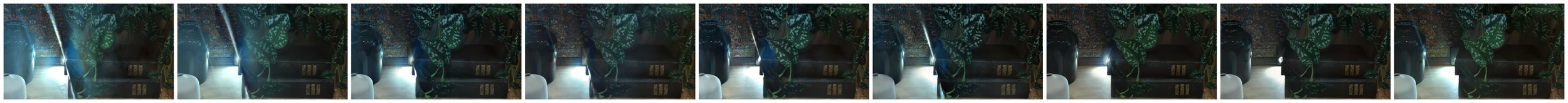}
    \put(-2,2) {\rotatebox{90}{dog}}
    \put(5.66,0.9){\makebox[0pt][c]{\tiny \colorbox{black}{\textcolor{white}{PSNR: 16.35}}}}
    \put(16.74,0.9){\makebox[0pt][c]{\tiny \colorbox{black}{\textcolor{white}{PSNR: 19.57}}}}
    \put(27.83,0.9){\makebox[0pt][c]{\tiny \colorbox{black}{\textcolor{white}{PSNR: 26.29}}}}
    \put(38.91,0.9){\makebox[0pt][c]{\tiny \colorbox{black}{\textcolor{white}{PSNR: 25.30}}}}
    \put(50.00,0.9){\makebox[0pt][c]{\tiny \colorbox{black}{\textcolor{white}{PSNR: 25.74}}}}
    \put(61.09,0.9){\makebox[0pt][c]{\tiny \colorbox{black}{\textcolor{white}{PSNR: 23.63}}}}
    \put(72.17,0.9){\makebox[0pt][c]{\tiny \colorbox{black}{\textcolor{white}{PSNR: 26.32}}}}
    \put(83.26,0.9){\makebox[0pt][c]{\tiny \colorbox{psnrgreen}{\textcolor{white}{PSNR: 28.09}}}}
\end{overpic}

\begin{overpic}[width=\linewidth]{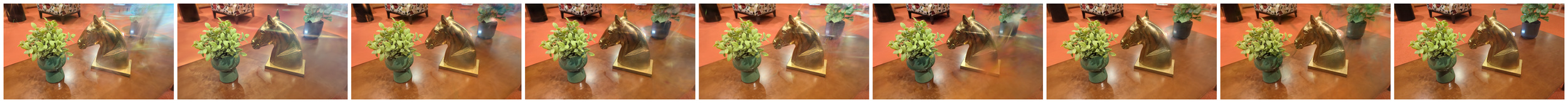}
    \put(-2,1) {\rotatebox{90}{plant}}
    \put(5.66,0.9){\makebox[0pt][c]{\tiny \colorbox{black}{\textcolor{white}{PSNR: 16.19}}}}
    \put(16.74,0.9){\makebox[0pt][c]{\tiny \colorbox{black}{\textcolor{white}{PSNR: 17.30}}}}
    \put(27.83,0.9){\makebox[0pt][c]{\tiny \colorbox{black}{\textcolor{white}{PSNR: 21.51}}}}
    \put(38.91,0.9){\makebox[0pt][c]{\tiny \colorbox{black}{\textcolor{white}{PSNR: 22.25}}}}
    \put(50.00,0.9){\makebox[0pt][c]{\tiny \colorbox{black}{\textcolor{white}{PSNR: 19.20}}}}
    \put(61.09,0.9){\makebox[0pt][c]{\tiny \colorbox{black}{\textcolor{white}{PSNR: 21.09}}}}
    \put(72.17,0.9){\makebox[0pt][c]{\tiny \colorbox{black}{\textcolor{white}{PSNR: 22.88}}}}
    \put(83.26,0.9){\makebox[0pt][c]{\tiny \colorbox{psnrgreen}{\textcolor{white}{PSNR: 23.29}}}}
\end{overpic}
    \caption{ Qualitative comparison of results from 3D reconstruction in the presence of lens flares. From left to right, we show the results from 3DGS on input images, trained with BilateralGrid~\cite{wang2024bilateral}, flare removed images from Flare7K++~\cite{dai2023flare7kpp}, from FR~\cite{wu2021train}, from ACL-FR~\cite{zhou2025improving}, from FlareX~\cite{lishenqu2025lishen}, from \oursw, and from \ourswo. As shown, our models achieve the best reconstruction, leading to fewer artifacts. This implies that our models are more consistent across different views than existing models.}
    \label{fig:3dgs_removed}
    \Description{A grid of 3D-scene renderings across five scenes (rows: plant, amp, billboard, guitar, dog), comparing, left to right, vanilla 3DGS trained on raw flare-corrupted input images, 3DGS with BilateralGrid, and 3DGS trained on images pre-processed by Flare7K++, FR, ACL-FR, FlareX, our model with lens glare, and our model without lens glare. The leftmost columns show visible flare-shaped artifacts baked into the 3D reconstruction, while the rightmost two columns (our models) show cleaner reconstructions with fewer flare-related artifacts.}
\end{figure*}

\begin{table*}[htbp]
\centering
\small

\caption{Quantitative comparison on decomposable flare reconstruction. Vanilla 3DGS reconstructs the composite image but produces no flare/scene decomposition, so it appears only in the Scene\,\&\,Flare sub-table. We group the remaining variants into two families. \textbf{Design alternatives} each replace a core design decision: `Ours 2D' drops the 1D constraint and lets the flare Gaussians move freely on the plane (similar to Deformable 3DGS~\cite{yang2023deformable3dgs}); `Ours sup in-flare' and `Ours unsup' replace our scene supervision with the input minus the reconstructed flare, and with no decomposition supervision at all, respectively. `Ours 2D' is trained against the same reference as `Ours'; it drops the 1D constraint, and with it the finer deformations $\delta_f$, which are redundant once the Gaussians move freely on the plane: it matches `Ours'\ on scene decomposition but loses $5.5$\,dB on flare, which we attribute to the 1D representation. The two supervision alternatives collapse on both ($11.5$--$14.9$\,dB on flare, $10.8$--$13.8$\,dB on scene) because their scene branch retains the flare; this margin is modestly inflated, since unlike the other variants they are not supervised with the reference (\secref{sec:flarereconstruction}). \textbf{Component ablations} remove one element of the same design and therefore land within $1.1$\,dB of `Ours': `Ours wo $\delta_l$' drops the light-position correction and performs on par with `Ours', since the finer deformations $\delta_f$ compensate for light-position error; only removing both breaks robustness (\secref{sec:light_robustness}); `Ours wo G-SAM2' replaces detection with a noisy scene centroid, showing the line equation is recovered automatically; `Ours wo $\delta_f$' removes the finer deformations and cannot represent asymmetric flares. `Ours unsup' attains the highest Scene\,\&\,Flare PSNR while decomposing poorly; we treat this as an operating-mode trade-off rather than a deficiency (\secref{sec:ablations}, `Operating modes'). Best and second-best are marked across \emph{all} variants within each sub-table, in \textcolor{red}{\textbf{red bold}} and \textcolor{blue}{\uline{blue underline}}.}

\label{tab:flare_modeling_comparison}

\resizebox{\textwidth}{!}{
\setlength{\tabcolsep}{2pt}

\begin{tabular}{l|ccc|ccc|ccc|ccc|ccc|ccc|ccc}
\multicolumn{22}{c}{\large \textbf{Flare - Decomposed}} \\ 
& \multicolumn{9}{c|}{ \textbf{Design alternatives}} & \multicolumn{9}{c|}{ \textbf{Component ablations}} & & & \\ \cmidrule(lr){2-10}\cmidrule(lr){11-19}
\multirow{2}{*}{Scene} & \multicolumn{3}{c|}{Ours 2D} & \multicolumn{3}{c|}{Ours sup in-flare}  & \multicolumn{3}{c|}{Ours unsup} & \multicolumn{3}{c|}{Ours wo $\delta_l$} & \multicolumn{3}{c|}{Ours wo G-SAM2} & \multicolumn{3}{c|}{Ours wo $\delta_f$} & \multicolumn{3}{c}{Ours} \\
 & PSNR$\uparrow$ & SSIM$\uparrow$ & LPIPS$\downarrow$ & PSNR$\uparrow$ & SSIM$\uparrow$ & LPIPS$\downarrow$ & PSNR$\uparrow$ & SSIM$\uparrow$ & LPIPS$\downarrow$ & PSNR$\uparrow$ & SSIM$\uparrow$ & LPIPS$\downarrow$ & PSNR$\uparrow$ & SSIM$\uparrow$ & LPIPS$\downarrow$ & PSNR$\uparrow$ & SSIM$\uparrow$ & LPIPS$\downarrow$ & PSNR$\uparrow$ & SSIM$\uparrow$ & LPIPS$\downarrow$ \\
\hline

hat & 31.14 & 0.850 & 0.299 & 28.01 & 0.659 & 0.358 & 30.29 & 0.720 & 0.301 & \textcolor{blue}{\uline{37.22}} & \textcolor{blue}{\uline{0.873}} & \textcolor{blue}{\uline{0.271}} & 35.75 & 0.862 & 0.286 & 36.93 & 0.867 & \textcolor{red}{\textbf{0.268}} & \textcolor{red}{\textbf{37.24}} & \textcolor{red}{\textbf{0.876}} & 0.273 \\
park & 24.93 & 0.508 & 0.355 & 18.79 & 0.666 & 0.307 & 20.90 & 0.704 & 0.285 & \textcolor{blue}{\uline{32.24}} & 0.809 & 0.269 & 32.07 & 0.785 & 0.276 & 32.14 & \textcolor{red}{\textbf{0.839}} & \textcolor{red}{\textbf{0.244}} & \textcolor{red}{\textbf{32.36}} & \textcolor{blue}{\uline{0.820}} & \textcolor{blue}{\uline{0.258}} \\
outtrunk & 27.96 & 0.855 & \textcolor{red}{\textbf{0.291}} & 16.87 & 0.446 & 0.516 & 19.78 & 0.511 & 0.427 & \textcolor{blue}{\uline{33.03}} & \textcolor{blue}{\uline{0.876}} & 0.294 & 30.52 & 0.867 & 0.303 & 32.59 & 0.871 & 0.294 & \textcolor{red}{\textbf{33.22}} & \textcolor{red}{\textbf{0.877}} & \textcolor{blue}{\uline{0.293}} \\
outtree & 28.87 & 0.879 & 0.266 & 19.82 & 0.526 & 0.488 & 20.37 & 0.473 & 0.403 & \textcolor{red}{\textbf{33.92}} & 0.906 & 0.263 & 33.75 & \textcolor{red}{\textbf{0.908}} & \textcolor{red}{\textbf{0.256}} & 33.11 & 0.903 & 0.260 & \textcolor{blue}{\uline{33.85}} & \textcolor{blue}{\uline{0.907}} & \textcolor{blue}{\uline{0.257}} \\
metro & 26.82 & 0.733 & 0.346 & 19.89 & 0.443 & 0.448 & 23.62 & 0.376 & 0.445 & \textcolor{red}{\textbf{35.21}} & \textcolor{red}{\textbf{0.882}} & \textcolor{blue}{\uline{0.266}} & 32.68 & 0.841 & 0.276 & \textcolor{blue}{\uline{35.02}} & \textcolor{blue}{\uline{0.880}} & \textcolor{red}{\textbf{0.263}} & 34.97 & 0.880 & 0.269 \\
intrunk & 26.26 & 0.839 & 0.269 & 17.11 & 0.441 & 0.446 & 22.64 & 0.633 & 0.323 & 32.95 & 0.893 & 0.257 & \textcolor{red}{\textbf{33.28}} & \textcolor{blue}{\uline{0.894}} & \textcolor{blue}{\uline{0.249}} & 33.19 & 0.890 & \textcolor{red}{\textbf{0.239}} & \textcolor{blue}{\uline{33.23}} & \textcolor{red}{\textbf{0.895}} & 0.257 \\
decos & 29.28 & 0.842 & 0.309 & 15.20 & 0.546 & 0.473 & 21.69 & 0.602 & 0.368 & 33.09 & 0.872 & 0.302 & \textcolor{blue}{\uline{33.17}} & \textcolor{red}{\textbf{0.872}} & 0.306 & 32.83 & 0.868 & \textcolor{red}{\textbf{0.286}} & \textcolor{red}{\textbf{33.18}} & \textcolor{blue}{\uline{0.872}} & \textcolor{blue}{\uline{0.301}} \\
chair & 24.81 & 0.832 & 0.314 & 16.31 & 0.542 & 0.495 & 20.47 & 0.635 & 0.387 & 29.55 & \textcolor{blue}{\uline{0.877}} & 0.299 & 28.66 & 0.859 & 0.315 & \textcolor{blue}{\uline{29.59}} & 0.875 & \textcolor{red}{\textbf{0.286}} & \textcolor{red}{\textbf{29.80}} & \textcolor{red}{\textbf{0.884}} & \textcolor{blue}{\uline{0.291}} \\
workshop & 29.88 & 0.842 & 0.251 & 13.13 & 0.560 & 0.482 & 16.34 & 0.224 & 0.524 & \textcolor{blue}{\uline{31.54}} & \textcolor{red}{\textbf{0.857}} & \textcolor{red}{\textbf{0.225}} & 29.86 & 0.844 & 0.239 & 31.52 & 0.853 & \textcolor{blue}{\uline{0.230}} & \textcolor{red}{\textbf{31.56}} & \textcolor{blue}{\uline{0.855}} & 0.235 \\
\hline
\textbf{Avg.} & 27.77 & 0.798 & 0.300 & 18.35 & 0.537 & 0.446 & 21.79 & 0.542 & 0.385 & \textcolor{blue}{\uline{33.19}} & \textcolor{blue}{\uline{0.872}} & 0.272 & 32.19 & 0.859 & 0.279 & 32.99 & \textcolor{blue}{\uline{0.872}} & \textcolor{red}{\textbf{0.263}} & \textcolor{red}{\textbf{33.27}} & \textcolor{red}{\textbf{0.874}} & \textcolor{blue}{\uline{0.270}} \\
\end{tabular}}

\vspace{2mm}

\resizebox{\textwidth}{!}{

\setlength{\tabcolsep}{2pt}
\begin{tabular}{l|ccc|ccc|ccc|ccc|ccc|ccc|ccc}
\multicolumn{22}{c}{\large \textbf{Scene - Decomposed}} \\  
& \multicolumn{9}{c|}{ \textbf{Design alternatives}} & \multicolumn{9}{c|}{ \textbf{Component ablations}} & & & \\ \cmidrule(lr){2-10}\cmidrule(lr){11-19}
\multirow{2}{*}{Scene} & \multicolumn{3}{c|}{Ours 2D} & \multicolumn{3}{c|}{Ours sup in-flare} & \multicolumn{3}{c|}{Ours unsup} & \multicolumn{3}{c|}{Ours wo $\delta_l$} & \multicolumn{3}{c|}{Ours wo G-SAM2} & \multicolumn{3}{c|}{Ours wo $\delta_f$} & \multicolumn{3}{c}{Ours} \\
 & PSNR$\uparrow$ & SSIM$\uparrow$ & LPIPS$\downarrow$ & PSNR$\uparrow$ & SSIM$\uparrow$ & LPIPS$\downarrow$ & PSNR$\uparrow$ & SSIM$\uparrow$ & LPIPS$\downarrow$ & PSNR$\uparrow$ & SSIM$\uparrow$ & LPIPS$\downarrow$ & PSNR$\uparrow$ & SSIM$\uparrow$ & LPIPS$\downarrow$ & PSNR$\uparrow$ & SSIM$\uparrow$ & LPIPS$\downarrow$ & PSNR$\uparrow$ & SSIM$\uparrow$ & LPIPS$\downarrow$ \\
\hline
hat & 35.12 & 0.960 & 0.128 & 26.63 & 0.782 & 0.170 & 28.95 & 0.889 & 0.155 & \textcolor{red}{\textbf{35.12}} & 0.960 & \textcolor{blue}{\uline{0.126}} & \textcolor{blue}{\uline{35.12}} & 0.960 & 0.127 & 35.08 & \textcolor{red}{\textbf{0.960}} & \textcolor{red}{\textbf{0.125}} & 35.08 & \textcolor{blue}{\uline{0.960}} & 0.126 \\
park & 30.78 & 0.903 & 0.338 & 18.63 & 0.747 & 0.426 & 20.67 & 0.832 & 0.374 & \textcolor{blue}{\uline{31.36}} & 0.914 & 0.330 & \textcolor{red}{\textbf{31.42}} & 0.914 & 0.329 & 31.35 & \textcolor{red}{\textbf{0.916}} & \textcolor{blue}{\uline{0.328}} & 31.28 & \textcolor{blue}{\uline{0.915}} & \textcolor{red}{\textbf{0.327}} \\
outtrunk & 29.17 & 0.862 & 0.222 & 16.69 & 0.746 & 0.322 & 19.08 & 0.788 & 0.310 & \textcolor{blue}{\uline{29.48}} & \textcolor{blue}{\uline{0.867}} & \textcolor{blue}{\uline{0.216}} & \textcolor{red}{\textbf{29.75}} & \textcolor{red}{\textbf{0.873}} & \textcolor{red}{\textbf{0.205}} & 29.09 & 0.861 & 0.224 & 29.16 & 0.862 & 0.223 \\
outtree & \textcolor{red}{\textbf{32.92}} & 0.928 & 0.134 & 19.39 & 0.804 & 0.207 & 18.91 & 0.837 & 0.208 & 32.68 & \textcolor{red}{\textbf{0.930}} & \textcolor{red}{\textbf{0.133}} & 32.70 & \textcolor{blue}{\uline{0.929}} & 0.133 & 32.63 & 0.927 & 0.133 & \textcolor{blue}{\uline{32.75}} & 0.928 & \textcolor{blue}{\uline{0.133}} \\
metro & 31.34 & 0.924 & \textcolor{blue}{\uline{0.171}} & 19.47 & 0.757 & 0.254 & 22.54 & 0.783 & 0.239 & 31.30 & 0.925 & 0.172 & \textcolor{red}{\textbf{31.52}} & \textcolor{red}{\textbf{0.926}} & \textcolor{red}{\textbf{0.169}} & \textcolor{blue}{\uline{31.41}} & \textcolor{blue}{\uline{0.926}} & 0.171 & 31.36 & 0.926 & 0.171 \\
intrunk & \textcolor{red}{\textbf{32.39}} & 0.935 & 0.204 & 16.83 & 0.702 & 0.312 & 21.62 & 0.856 & 0.270 & 32.27 & 0.935 & 0.205 & 32.21 & \textcolor{red}{\textbf{0.936}} & \textcolor{blue}{\uline{0.203}} & 32.18 & 0.934 & 0.207 & \textcolor{blue}{\uline{32.27}} & \textcolor{blue}{\uline{0.935}} & \textcolor{red}{\textbf{0.203}} \\
decos & \textcolor{red}{\textbf{32.47}} & \textcolor{red}{\textbf{0.924}} & \textcolor{blue}{\uline{0.166}} & 15.05 & 0.629 & 0.310 & 21.33 & 0.776 & 0.263 & 32.29 & 0.922 & 0.168 & \textcolor{blue}{\uline{32.45}} & 0.923 & 0.167 & 32.28 & 0.922 & 0.167 & 32.44 & \textcolor{blue}{\uline{0.923}} & \textcolor{red}{\textbf{0.165}} \\
chair & \textcolor{red}{\textbf{28.97}} & \textcolor{red}{\textbf{0.887}} & 0.209 & 16.35 & 0.555 & 0.324 & 20.02 & 0.799 & 0.305 & 28.66 & 0.886 & 0.209 & \textcolor{blue}{\uline{28.71}} & 0.886 & \textcolor{red}{\textbf{0.209}} & 28.62 & \textcolor{blue}{\uline{0.887}} & \textcolor{blue}{\uline{0.209}} & 28.69 & 0.886 & 0.209 \\
workshop & 32.91 & 0.965 & 0.080 & 13.08 & 0.547 & 0.243 & 15.64 & 0.753 & 0.231 & \textcolor{blue}{\uline{33.17}} & \textcolor{red}{\textbf{0.966}} & 0.079 & 33.16 & 0.965 & 0.081 & \textcolor{red}{\textbf{33.20}} & 0.966 & \textcolor{blue}{\uline{0.079}} & 33.12 & \textcolor{blue}{\uline{0.966}} & \textcolor{red}{\textbf{0.078}} \\
\hline
\textbf{Avg.} & 31.78 & 0.921 & 0.183 & 18.01 & 0.697 & 0.285 & 20.97 & 0.813 & 0.262 & \textcolor{blue}{\uline{31.81}} & \textcolor{blue}{\uline{0.923}} & 0.182 & \textcolor{red}{\textbf{31.89}} & \textcolor{red}{\textbf{0.924}} & \textcolor{red}{\textbf{0.180}} & 31.76 & 0.922 & 0.183 & 31.79 & 0.922 & \textcolor{blue}{\uline{0.182}} \\
\end{tabular}}

\vspace{2mm}
\resizebox{\textwidth}{!}{
\setlength{\tabcolsep}{2pt}
\begin{tabular}{l|ccc|ccc|ccc|ccc|ccc|ccc|ccc|ccc}
\multicolumn{25}{c}{\large \textbf{Scene \& Flare}} \\  
& & & & \multicolumn{9}{c|}{ \textbf{Design alternatives}} & \multicolumn{9}{c|}{ \textbf{Component ablations}} & & & \\ \cmidrule(lr){5-13}\cmidrule(lr){14-22}
\multirow{2}{*}{Scene} & \multicolumn{3}{c|}{ Vanilla 3DGS} & \multicolumn{3}{c|}{Ours 2D} & \multicolumn{3}{c|}{Ours sup in-flare} & \multicolumn{3}{c|}{Ours unsup} & \multicolumn{3}{c|}{Ours wo $\delta_l$} & \multicolumn{3}{c|}{Ours wo G-SAM2} & \multicolumn{3}{c|}{Ours wo $\delta_f$} & \multicolumn{3}{c}{Ours} \\
 & PSNR$\uparrow$ & SSIM$\uparrow$ & LPIPS$\downarrow$ & PSNR$\uparrow$ & SSIM$\uparrow$ & LPIPS$\downarrow$ & PSNR$\uparrow$ & SSIM$\uparrow$ & LPIPS$\downarrow$ & PSNR$\uparrow$ & SSIM$\uparrow$ & LPIPS$\downarrow$ & PSNR$\uparrow$ & SSIM$\uparrow$ & LPIPS$\downarrow$ & PSNR$\uparrow$ & SSIM$\uparrow$ & LPIPS$\downarrow$ & PSNR$\uparrow$ & SSIM$\uparrow$ & LPIPS$\downarrow$ & PSNR$\uparrow$ & SSIM$\uparrow$ & LPIPS$\downarrow$ \\
\hline
hat & 35.60 & \textcolor{blue}{\uline{0.963}} & 0.122 & 30.08 & 0.942 & 0.141 & 34.75 & 0.956 & 0.138 & \textcolor{red}{\textbf{36.28}} & \textcolor{red}{\textbf{0.964}} & \textcolor{red}{\textbf{0.118}} & 35.62 & 0.962 & \textcolor{blue}{\uline{0.119}} & 34.17 & 0.956 & 0.138 & 35.32 & 0.961 & 0.121 & \textcolor{blue}{\uline{35.69}} & 0.962 & 0.119 \\
park & 31.01 & \textcolor{blue}{\uline{0.922}} & 0.298 & 24.87 & 0.870 & 0.342 & 30.29 & 0.911 & 0.318 & \textcolor{red}{\textbf{31.90}} & \textcolor{red}{\textbf{0.926}} & \textcolor{red}{\textbf{0.295}} & 31.17 & 0.919 & 0.304 & \textcolor{blue}{\uline{31.27}} & 0.919 & 0.305 & 30.99 & 0.921 & \textcolor{blue}{\uline{0.298}} & 30.84 & 0.918 & 0.299 \\
outtrunk & 29.16 & 0.883 & 0.210 & 26.89 & 0.871 & 0.229 & 29.65 & 0.869 & 0.242 & \textcolor{blue}{\uline{30.36}} & 0.880 & 0.214 & \textcolor{red}{\textbf{30.62}} & \textcolor{blue}{\uline{0.887}} & \textcolor{red}{\textbf{0.206}} & 29.43 & \textcolor{red}{\textbf{0.888}} & \textcolor{blue}{\uline{0.207}} & 30.16 & 0.880 & 0.216 & 30.36 & 0.882 & 0.212 \\
outtree & 32.71 & \textcolor{red}{\textbf{0.947}} & \textcolor{red}{\textbf{0.120}} & 29.29 & 0.933 & 0.138 & 32.93 & 0.928 & 0.151 & \textcolor{red}{\textbf{35.51}} & 0.943 & \textcolor{blue}{\uline{0.122}} & \textcolor{blue}{\uline{35.49}} & \textcolor{blue}{\uline{0.945}} & 0.123 & 35.28 & 0.944 & 0.123 & 34.51 & 0.941 & 0.126 & 35.42 & 0.944 & 0.123 \\
metro & 31.33 & 0.920 & 0.179 & 25.96 & 0.909 & 0.185 & 31.45 & 0.933 & 0.186 & 32.11 & 0.939 & 0.165 & \textcolor{red}{\textbf{32.49}} & \textcolor{red}{\textbf{0.940}} & \textcolor{blue}{\uline{0.161}} & 30.63 & 0.936 & 0.167 & \textcolor{blue}{\uline{32.44}} & \textcolor{blue}{\uline{0.940}} & 0.161 & 32.33 & 0.940 & \textcolor{red}{\textbf{0.160}} \\
intrunk & 32.55 & 0.942 & 0.192 & 26.51 & 0.927 & 0.214 & 31.34 & 0.931 & 0.245 & 34.08 & 0.949 & 0.192 & 33.56 & 0.949 & 0.196 & \textcolor{red}{\textbf{34.12}} & \textcolor{red}{\textbf{0.950}} & \textcolor{blue}{\uline{0.190}} & 34.04 & 0.949 & 0.191 & \textcolor{blue}{\uline{34.11}} & \textcolor{blue}{\uline{0.950}} & \textcolor{red}{\textbf{0.189}} \\
decos & 32.94 & \textcolor{blue}{\uline{0.949}} & 0.160 & 29.55 & 0.936 & 0.173 & 32.18 & 0.934 & 0.201 & \textcolor{red}{\textbf{33.79}} & \textcolor{red}{\textbf{0.949}} & \textcolor{red}{\textbf{0.149}} & \textcolor{blue}{\uline{33.13}} & 0.945 & 0.162 & 33.10 & 0.946 & 0.161 & 32.96 & 0.946 & 0.159 & 33.09 & 0.946 & \textcolor{blue}{\uline{0.157}} \\
chair & 30.47 & 0.914 & 0.193 & 25.33 & 0.898 & 0.214 & 31.12 & 0.902 & 0.241 & \textcolor{red}{\textbf{32.64}} & \textcolor{red}{\textbf{0.920}} & \textcolor{red}{\textbf{0.185}} & 31.82 & 0.918 & 0.188 & 30.35 & 0.915 & 0.198 & 31.99 & 0.919 & 0.188 & \textcolor{blue}{\uline{32.15}} & \textcolor{blue}{\uline{0.919}} & \textcolor{blue}{\uline{0.187}} \\
workshop & \textcolor{blue}{\uline{35.57}} & \textcolor{red}{\textbf{0.983}} & \textcolor{red}{\textbf{0.046}} & 31.32 & 0.968 & 0.077 & 26.70 & 0.941 & 0.129 & \textcolor{red}{\textbf{36.37}} & \textcolor{blue}{\uline{0.981}} & \textcolor{blue}{\uline{0.057}} & 33.59 & 0.973 & 0.068 & 31.24 & 0.968 & 0.077 & 33.68 & 0.972 & 0.070 & 33.53 & 0.973 & 0.068 \\
\hline
\textbf{Avg.} & 32.37 & 0.936 & 0.169 & 27.76 & 0.917 & 0.190 & 31.16 & 0.923 & 0.206 & \textcolor{red}{\textbf{33.67}} & \textcolor{red}{\textbf{0.939}} & \textcolor{red}{\textbf{0.166}} & \textcolor{blue}{\uline{33.06}} & \textcolor{blue}{\uline{0.938}} & 0.170 & 32.18 & 0.936 & 0.174 & 32.90 & 0.937 & 0.170 & \textcolor{blue}{\uline{33.06}} & 0.937 & \textcolor{blue}{\uline{0.168}} \\
\end{tabular}
}

\end{table*}

\subsection{3D Reconstruction with Flare Corruption}
\label{sec:lensflaresremoval3d}
In this experiment, we demonstrate the performance of our model in removing flares consistently across multiple views. To achieve this, we train a 3DGS model on sequences captured with a light source and flare corruption in training views. We select approximately $10\%$ of the images for testing, ensuring they do not contain flares or light sources for a fair evaluation. We evaluate against ground-truth images (\textit{not} preprocessed by a flare-removal model).

We train a 3DGS model on different sets of flare-removed images generated by our baseline flare-removal models. Quantitatively, 3DGS trained on our models' outputs outperforms existing methods, as shown in~\tabref{tab:3dgs_removed}. These results are well substantiated qualitatively in~\figref{fig:3dgs_removed}. It can be seen that inconsistent lens-flare removal across views by models leads to artifacts in reconstructions. It can also be seen that BilateralGrid~\cite{wang2024bilateral} fails to remove flares, as they are inherently different from image exposure changes across views. 3DGS trained on our flare-removed models yields far fewer artifacts, resulting in high-quality reconstructions.

\subsection{Flare Reconstruction}
\label{sec:flarereconstruction}
We evaluate the performance of our flare representation model for reconstruction and decomposition. We compare these reconstructions with our baseline and ablation models. %

\paragraph*{What counts as a correct decomposition.} The target is not self-evident. A light source is a genuine scene emitter, yet the glare surrounding it is produced by the imaging system, so the boundary between `scene' and `flare' has to be stated rather than assumed. We settle it by choice of reference. Taking $I_{\text{removed}}$, the output of our removal model \ourswo, as the flare-free reference, we require (i) the scene render to match $I_{\text{removed}}$, and (ii) the flare branch's contribution to match the residual $I_{\text{input}} - I_{\text{removed}}$. Because \ourswo\ retains the light source while removing the flare and the glare around it (\secref{sec:flareremoval}), the emitter falls on the scene side and the artifacts it induces on the flare side. This reference measures the contribution of flares to the composite scene rather than the flare itself. We need this difference-based reference because rendering the flare Gaussians alone would composite them against an empty background, and we have no direct reference against which to evaluate such a render. We therefore compute the flare contribution as $|I_{\text{scene+flare}} - I_{\text{scene}}|$. Both sides of (ii) are thus formed alike, which makes the criterion agnostic to the compositing strategy.

\paragraph*{What the residual measures.} The residual in (ii) is not a pure flare layer. A flare saturates and occludes the scene behind it, so no exact flare is recoverable from a single image, and $I_{\text{input}} - I_{\text{removed}}$ therefore also absorbs the discrepancy between the true occluded content and \ourswo's reconstruction of it, including any light-source energy the removal model does not reproduce exactly. The flare branch is trained to account for whatever the removal model does not, so the two components together reconstruct the input regardless of where the removal model draws its boundary. Because flare Gaussians are composited in front of the scene, a transferred flare also carries its own light source into a new target (\figref{fig:teaser}). It remains a close proxy for flare reconstruction quality, because reproducing the flare's contribution to the composite is exactly what our flare branch is responsible for. As $I_{\text{removed}}$ is model-derived, the scene and flare metrics measure agreement with this reference rather than with an unobtainable ground truth: they are meaningful for comparing variants scored against a common reference, but not as absolute quality figures. Two facts limit this circularity: all metrics are computed at held-out frames that no model sees during training, and \ourswo\ was never trained on these scenes; it is validated separately on real paired captures (Flare7K++, \tabref{tab:flareremoval}) and, across views, by reconstructions that best match real flare-free captures (\secref{sec:lensflaresremoval3d}). What remains is shared bias: variants supervised with $I_{\text{removed}}$ inherit its systematic errors, which the reference also carries, so their scores are modestly favoured over variants that are not. Only the composite render is compared to the captured image directly, making it the sole removal-independent measure.

\begin{figure*}
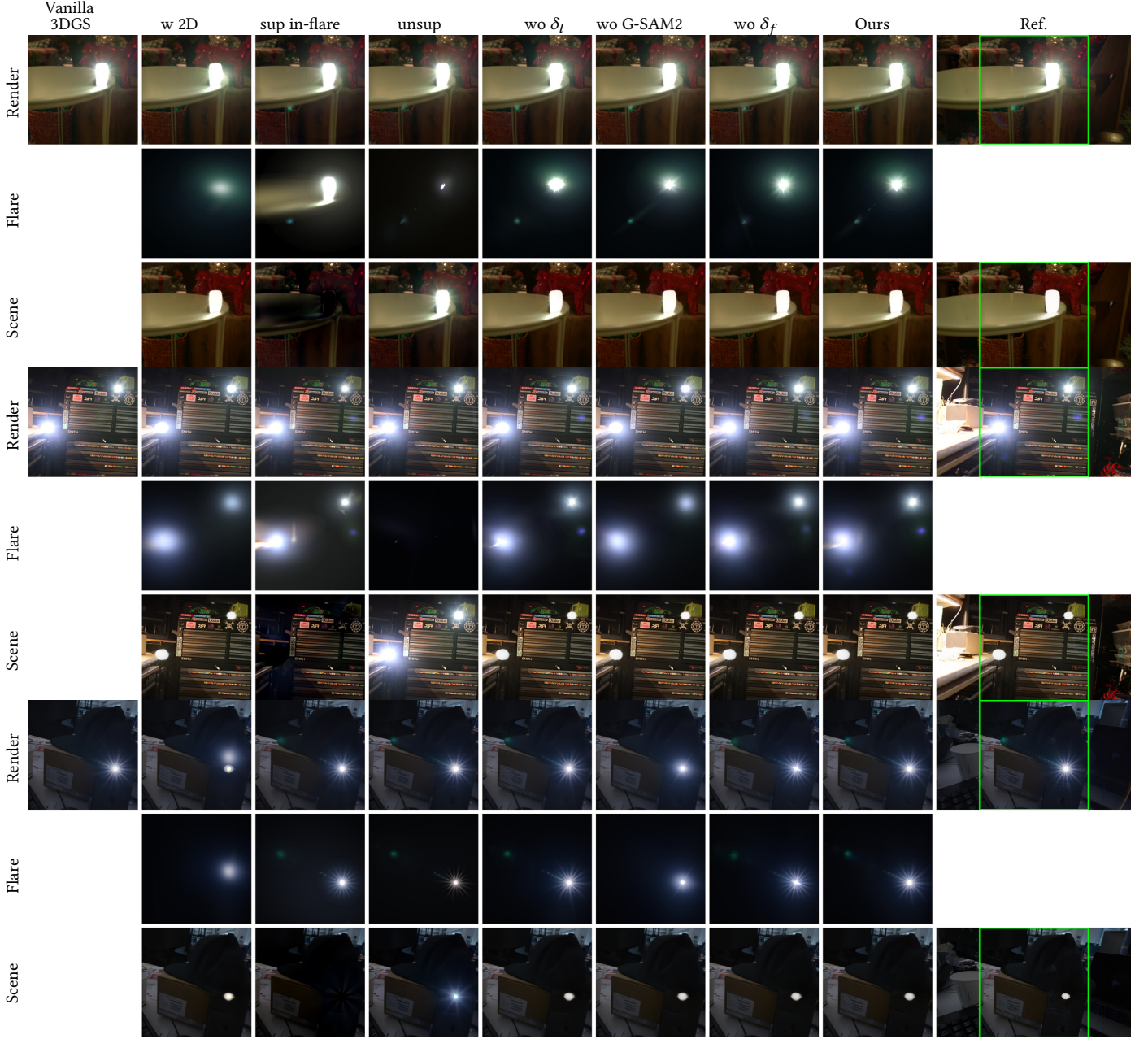

    \centering
    \vspace{5mm}
    \input{figures/fmod/images/decos_00020_overpic}
    \input{figures/fmod/images/workshop_00027_overpic}
    \input{figures/fmod/images/redhat_00020_overpic}
    \caption{Qualitative comparison (\emph{decos}: top, \emph{workshop}: middle, \emph{hat}: bottom) of flare representation and decomposition with baselines (3DGS), and ablations. We show the results of 3DGS on the left, followed by our ablation models: `Ours 2D', `in-flare', `unsup', `wo $\delta_l$', `wo G-SAM2', `wo $\delta_f$', and `Ours'. We are interested in a model that provides the best decomposition of lens flares and 3D scenes. 3DGS cannot reconstruct flares because they are not 3D-consistent. `Ours 2D', being unconstrained on the flare locations, struggles to reconstruct them. `in-flare' supervises the scene to reconstruct everything except what the flare model models. `unsup' leads to inconsistent decompositions as the scene is unsupervised. `wo $\delta_l$' performs on par with our model, since the finer deformations absorb small light-location errors. `wo G-SAM2' still recovers the line equation, as $\delta_l$ and $\delta_f$ absorb the error of the centroid-based light estimate. In `Ours wo $\delta_f$', the lens flares lack asymmetric details. Finally, our model leads to better decompositions and flare modeling. `Ref.' denotes the reference images the metrics use, derived from the captured input and our removal output (\secref{sec:flarereconstruction}).}
    \label{fig:flare_modeling_comparison}
    \Description{A grid of rendered scene comparisons across three scenes (rows: decos, workshop, hat), each showing 3DGS alongside our ablation variants, left to right: 3DGS, `Ours 2D', `in-flare', `unsup', `wo delta-l', `wo G-SAM2', `wo delta-f', and `Ours', plus a reference column. The 3DGS column shows scenes with reflective flares either missing or baked in as artifacts, while later columns progressively better isolate and reconstruct distinct flare shapes near each scene's light source(s), with the final `Ours' column most closely matching the reference decomposition into scene and flare components.}
\end{figure*}

\paragraph*{Comparison to 3DGS.} The results in~\figref{fig:flare_modeling_comparison} demonstrate that 3DGS initialized with Colmap~\cite{schoenberger2016mvs,schoenberger2016sfm} can reconstruct some scattering flares close to the light source that are approximately 3D consistent, as Colmap provides some initialization for these flares. However, reflective flares in the given images are completely ignored, as they are highly variable in shape, color, and size across views, and Colmap cannot provide initializations for such 3D inconsistent objects. Our model, `Ours', on the other hand, helps reconstruct the highly variable flare objects and represents them consistently across different views (see supplementary video for flares rendered from novel views). 
On the composite Scene\,\&\,Flare metric our model exceeds vanilla 3DGS by only $0.69$\,dB; this is expected, since both are fit to the same captured images and flare-affected pixels are a limited fraction of each frame, so even a model that smears or ignores the flare scores well on a whole-image metric. The two differ on decomposition (\tabref{tab:flare_modeling_comparison}, \figref{fig:flare_modeling_comparison}), where vanilla 3DGS produces no flare component at all and cannot be scored -- whole-image metrics measure reconstruction quality rather than decomposition, so composite fidelity is a sanity check here, not the axis of comparison.

\paragraph*{Rendering speed.} Our representation renders in real time at capture resolution. Flare Gaussians are emitted as standard 3DGS primitives and concatenated with the scene Gaussians for a single rasterization pass, so no additional render is required. Measured on an RTX 4090 at $1920\!\times\!1080$ on a single-light scene (\emph{hat}; $10{,}965$ flare Gaussians alongside $162{,}283$ scene Gaussians), the deformation pass takes $3.92$\,ms and the joint rasterization of all $173{,}248$ Gaussians takes $6.86$\,ms, for a frame time of $10.79$\,ms ($92.7$\,FPS). Rasterizing the scene alone -- vanilla 3DGS on the same scene -- takes $6.74$\,ms ($148$\, FPS), so batching the flare Gaussians into the same pass adds only $1.8\%$ to rasterization cost; the deformation pass is the dominant expense of the flare model. It evaluates per Gaussian rather than per pixel, and we measure it to be independent of output resolution. We instantiate one flare model per light source, each with its own deformation network and canonical Gaussians, so this cost grows with the number of lights: $3.9$, $6.8$, and $11.1$\,ms for one, two, and three lights.

\subsubsection{Ablations} \label{sec:ablations} Our goals for the flare removal model are to decompose the input images to have a good 3D scene reconstruction alongside lens flares that can generalize to novel views. We perform ablations on our flare representation model to evaluate our design choices towards these goals. We show qualitative results in~\figref{fig:flare_modeling_comparison}, and quantitative results in~\tabref{tab:flare_modeling_comparison}. In `Ours 2D', we allow the flare Gaussians to move freely in the 2D plane, without our 1D representation. This model struggles to reconstruct flares since the Gaussians are unconstrained (flare PSNR $27.77$\,dB vs.\ $33.27$\,dB for Ours), and the composite reconstruction degrades correspondingly (Scene+Flare PSNR $27.76$\,dB, the lowest of all ablations). Next, we evaluate the effect of scene supervision on decomposition. In `Ours sup in-Flare' and `Ours unsup', we train the scene Gaussians with the reconstructed flare image subtracted from the input image and with no scene-only supervision, respectively. These models lead to inconsistent decompositions, missing different aspects of flares inconsistently across different scenes. However, `Ours unsup', whose flare model augments 3DGS in reconstructing aspects of the captures that 3DGS misses, namely reflective lens flares, attains the highest composite PSNR of all variants. In `Ours wo $\delta_l$', we train our model without the light position correction in \eqref{eq:light_pos_correction}. It performs on par with `Ours', as the finer deformations $\delta_f$ absorb the light-position error (\secref{sec:light_robustness}). Further, in `Ours wo G-SAM2', we initialize the light source locations using the centroid of the scene's Colmap point cloud with added noise, demonstrating that our model can fairly accurately reconstruct flares even without explicit light source detection, thanks to the light-position correction and finer deformations (\eqref{eq:light_pos_correction}, \eqref{eq:finer_deformations}). In `Ours wo $\delta_f$', where we train the model without the finer deformations $\delta_f$ (see~\eqref{eq:finer_deformations}), the model struggles to reconstruct asymmetric flares, leading to a slightly lower performance on flare reconstruction (PSNR of $32.99$\,dB). While the `Ours unsup' achieves higher composite Scene+Flare PSNR ($33.67$ vs.\ $33.06$\,dB for Ours)---since the flare model's additional capacity improves overall reconstruction---it performs poorly on scene decomposition ($20.97$ vs.\ $31.79$\,dB for Ours). Our model achieves the best flare decomposition (PSNR $33.27$\,dB) while remaining within $0.1$\,dB of the highest scene-decomposition PSNR ($31.89$ for `Ours wo G-SAM2' vs.\ $31.79$\,dB for `Ours'). However, Ours outperforms `Ours wo G-SAM2' in Flare decomposition reconstruction by $\sim1.1$\,dB PSNR. Our model therefore offers the best scene/flare separation, which is the desired trade-off for our decomposable-reconstruction goal.

\paragraph*{Operating modes.} The ablations above are not only design controls; each also corresponds to a useful operating point of our framework. The default \emph{Ours} setting bundles the light source and its flare into a single, self-contained 3D flare effect, suitable for transferring the captured flare to other scenes (e.g., VFX). \emph{Ours~unsup} instead prioritizes faithful composite reconstruction with flares, giving the highest Scene+Flare PSNR ($33.67$, vs.\ $33.06$\,dB for the default), and serves applications that need flare-preserving re-rendering. The user can thus pick the training configuration matching the downstream task.

\begin{figure}
  \begin{center}
    \centering
    \begin{overpic}[width=\linewidth, ]{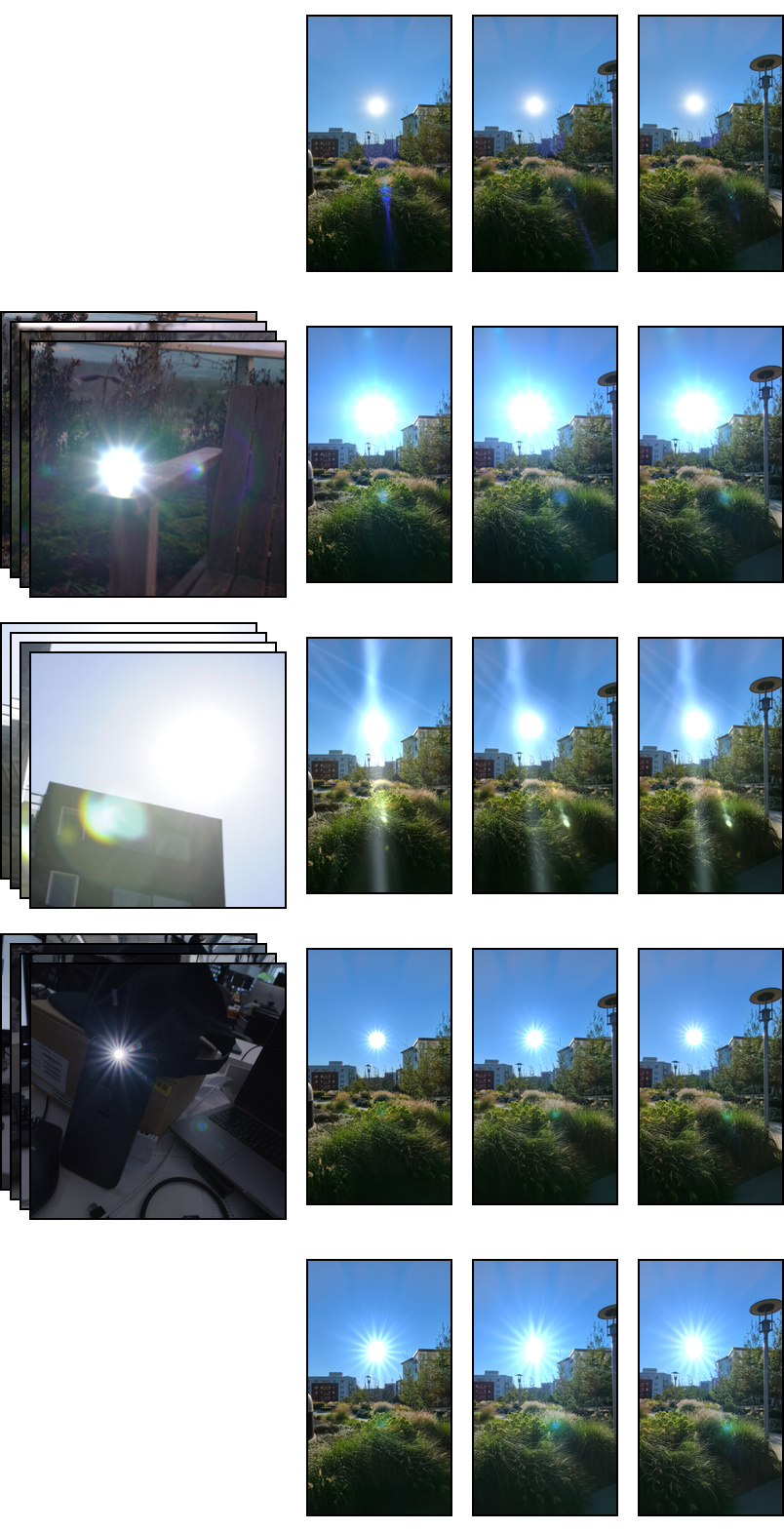}
        \put(31,100){\small\bfseries Target Scene}
        \put(5,-1){\small\bfseries Source Flares}
        \put(24,-1){\small\bfseries Lens Flares Transferred to Scene}
        \put(18,8){\small\bfseries \rotatebox[]{90}{Scaled  $2 \times$} }
    \end{overpic}
    \captionof{figure}{%
      \textbf{Lens-flare transfer to a novel scene.}
        We transfer flares reconstructed from three \emph{source} scenes (left column) -- two from our captured reconstruction dataset (\secref{sec:flarereconstruction}) and one additional captured scene, also used for source flares in Fig.~\ref{fig:teaser} -- onto a single \emph{target} scene that does not appear in either of our captured datasets (\secref{sec:datasets}), rendered from three novel views (top row). Each following row composites the corresponding source flare onto every view of the target scene shown above it. The bottom row additionally shows one of the transferred flares with its Gaussians uniformly scaled $2\times$, demonstrating simple post-hoc artistic control over the flare's apparent intensity and extent.}
    \label{fig:scenetransfer}
  \end{center}
  \Description{A grid transferring three source flares onto a target scene of an apartment courtyard with grass and a lamp post, shown from three novel viewpoints (top row, no flare). The three source flares (left column) are a bright deck-railing light with colored ghost artifacts, a sun behind a building corner with a hazy halo, and a headlamp on a desk with a bright starburst. Each subsequent row composites one source flare onto the three target views, adding a bright sun-like glare with radial streaks near the sky in each. A final row shows one transferred flare with its Gaussians scaled up, producing a visibly larger and brighter starburst than the corresponding un-scaled version.}
\end{figure}
\subsection{Application: Flare transfer}\label{sec:transfer}
\paragraph*{Transfer to novel images.} We first demonstrate transfer onto a
target that has no associated 3D reconstruction at all: an arbitrary 2D target
image (Fig.~\ref{fig:teaser}). Because such a target carries no camera pose,
scene geometry, or light-source depth, the camera- and light-conditioned
anchoring used for scene transfer (below) does not apply. Instead, we specify a
single 2D point in the target image as the desired light-source location --
either manually or via the same Grounded-SAM2 detection used elsewhere in our
pipeline -- and query the flare MLP with the source camera and source light as
before, so the transferred flare keeps its learned in-distribution shape and
color, while its placement is now driven by this target-image point rather than
a projected 3D light position. The resulting flare Gaussians are rendered and
alpha-composited directly onto the target image. As the target image carries no
depth information, we do not perform a visibility test in this setting: the
flare is composited unconditionally on top of the target image.

\paragraph*{Transfer to novel scenes.} To demonstrate that our learned flares are not tied to the scene they were
optimized on, we transfer flares trained on three \emph{source} scenes -- two
from our captured reconstruction dataset (\secref{sec:flarereconstruction}) and
one additional captured scene not part of either dataset (also used for source
flares in Fig.~\ref{fig:teaser}) -- onto a single \emph{target} scene that
likewise does not appear in either of our captured datasets
(\secref{sec:datasets}) and is used only as a transfer target
(Fig.~\ref{fig:scenetransfer}). Both of these additional captures use the sun
as the light source, diversifying beyond the light types in
\secref{sec:flarereconstruction}. Each flare MLP
ensemble is optimized jointly with its source scene's Gaussians and is
conditioned on the source cameras and light positions. At transfer time we
decouple \emph{appearance} from \emph{placement}. The MLP is conditioned on camera
and light positions and can be queried at any of them, as the supplementary video shows
by sweeping the light along the principal-point line and then moving the camera as well.
For transfer, however, we deliberately query it with the source camera and the source light it was trained with, since a target
camera may lie well outside the source capture's distribution and would elicit an
extrapolated flare. This keeps the learned flare shape and color in-distribution, while
the resulting flare Gaussians are anchored to the target scene by placing them relative
to the target camera and aiming them at the target scene's light position(s). For each frame we first
rasterize the target Gaussians from the target camera, which also yields an
inverse-depth map used for a per-light visibility test: a flare is instantiated
only for target lights that are not occluded by target geometry. One flare is
built per visible target light and the instances are merged, then composited
with the target render so that the transferred flare correctly blends with (and
is occluded by) the target scene. Because the flare follows the target cameras,
we render along the target scene's smooth video trajectory to produce the
transfer sequences. Since the flare is represented explicitly as Gaussians, it
also remains directly editable after transfer: the bottom row of
Fig.~\ref{fig:scenetransfer} shows a transferred flare with its Gaussians
uniformly scaled by $2\times$, giving simple artistic control over the flare's
apparent intensity and extent without any retraining.

\subsection{Light-Location Robustness Analysis}
\label{sec:light_robustness}
A key practical concern is how sensitive our flare representation is to errors in the input light-source locations. In the wild, Grounded-SAM2 detections can be inconsistent across views or off by tens-to-hundreds of pixels under saturation, motion blur, or occlusion. Any flare model that hard-codes $l$ as a fixed prior is therefore vulnerable to a noisy front-end. To stress-test our pipeline, we displace each light's estimated 3D position by a single Gaussian offset of standard deviation $\sigma\in\{0.0,0.1,0.5,1.0\}$ (in scene units), held fixed for the whole sequence; projected into the views, this corresponds to an average error of up to $\sim\!262$\,px, on five of our scenes (\emph{chair}, \emph{hat}, \emph{metro}, \emph{outtree}, \emph{workshop}, spanning one to three light sources), and evaluate every combination of our light-position correction $\delta_l$ (\eqref{eq:light_pos_correction}) and finer deformations $\delta_f$ (\eqref{eq:finer_deformations}).

\begin{figure}[t]
    \centering
    \includegraphics[width=\linewidth]{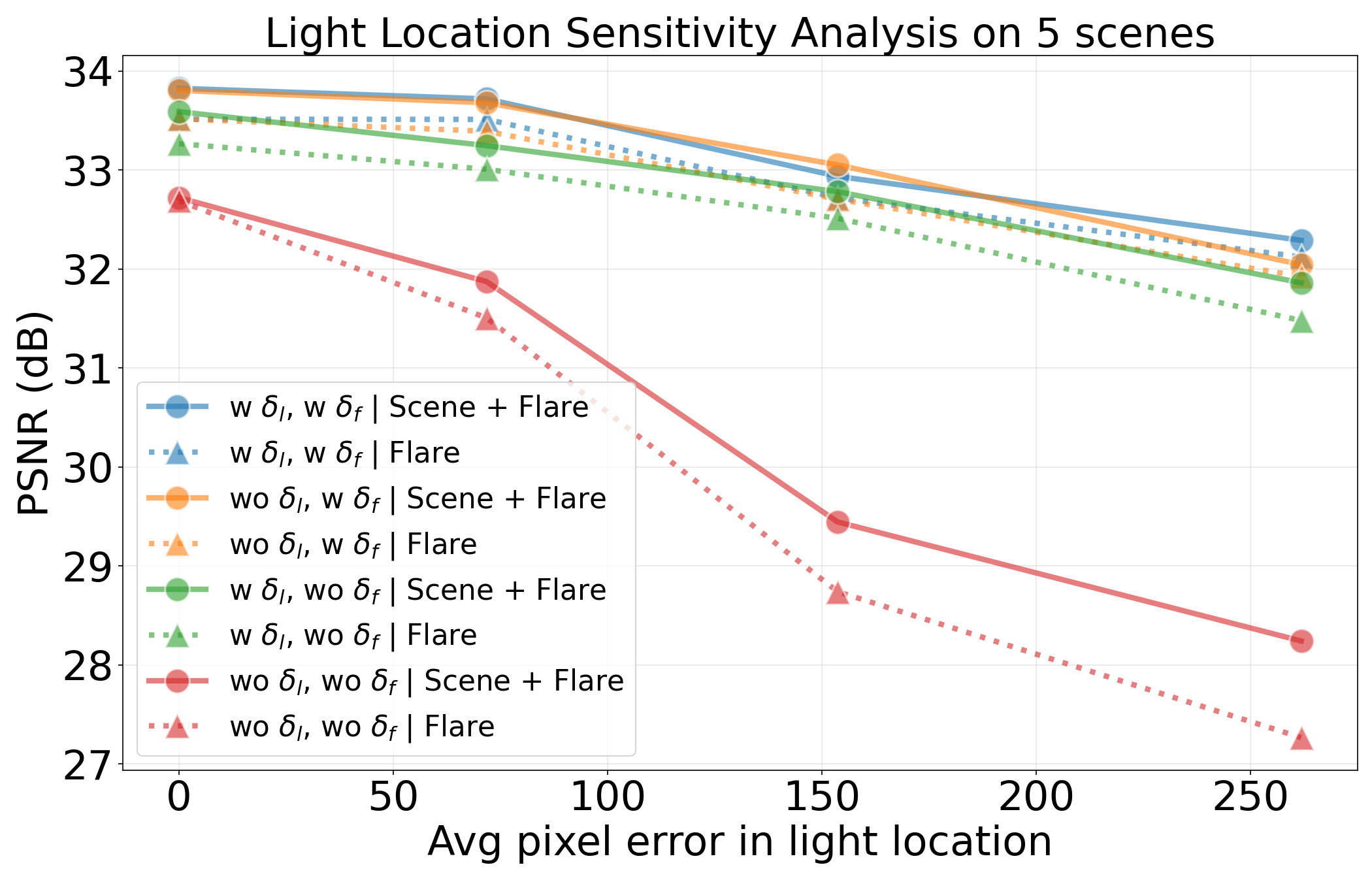}
    \caption{Robustness of our flare representation to error in the input light position $l$, on five scenes. PSNR against the average pixel error injected into $l$, for the composite render (solid) and the flare (dotted); colour indicates which of the light-position correction $\delta_l$ and finer deformations $\delta_f$ are enabled. With either correction the model degrades gracefully; with both removed, composite and flare PSNR fall by over $4$\,dB.}
    \label{fig:light_robustness}
    \Description{A line chart titled ``Light Location Sensitivity Analysis on 5 scenes'', plotting PSNR in decibels (y-axis, 27 to 34) against average pixel error in light location (x-axis, 0 to about 260). Eight lines show four correction settings (with/without delta-l and with/without delta-f) each for Scene+Flare PSNR (solid) and Flare-only PSNR (dotted). The three settings with at least one correction enabled cluster together, declining gently from about 33.5-34 dB to about 31.5-32 dB. The setting with both corrections removed (red) declines much more steeply, from about 32.7 dB to about 27.2-28.2 dB.}
\end{figure}

\begin{figure}[t]
    \centering
    \setlength{\tabcolsep}{1pt}
    \renewcommand{\arraystretch}{0.6}
    \begin{tabular}{@{}c@{\hspace{2pt}}c@{}}
        \multicolumn{2}{c}{\scriptsize Scene + Flare} \\
        \rotatebox{90}{\scriptsize \shortstack{w $\delta_l$ \\ \& w $\delta_f$}} &
        \includegraphics[width=0.88\linewidth]{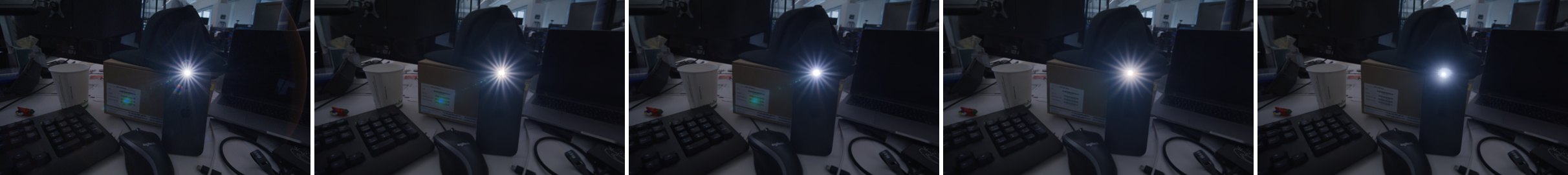} \\
        \rotatebox{90}{\scriptsize \shortstack{wo $\delta_l$ \\ \& w $\delta_f$}} &
        \includegraphics[width=0.88\linewidth]{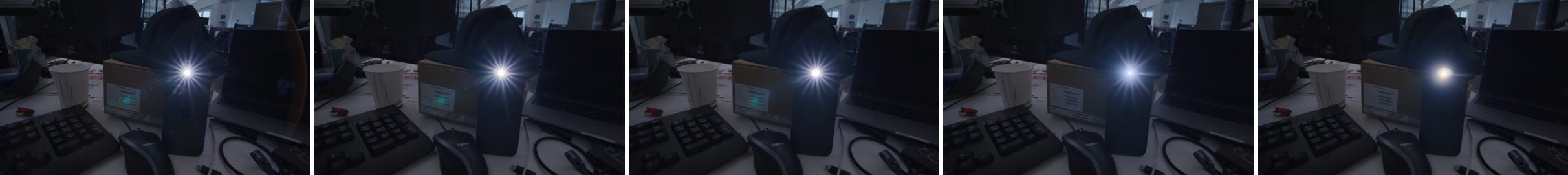} \\
        \rotatebox{90}{\scriptsize \shortstack{w $\delta_l$ \\ \& wo $\delta_f$}} &
        \includegraphics[width=0.88\linewidth]{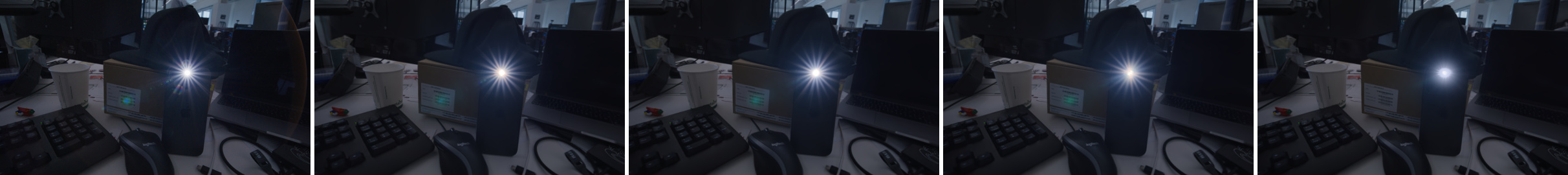} \\
        \rotatebox{90}{\scriptsize \shortstack{wo $\delta_l$ \\ \& wo $\delta_f$}} &
        \includegraphics[width=0.88\linewidth]{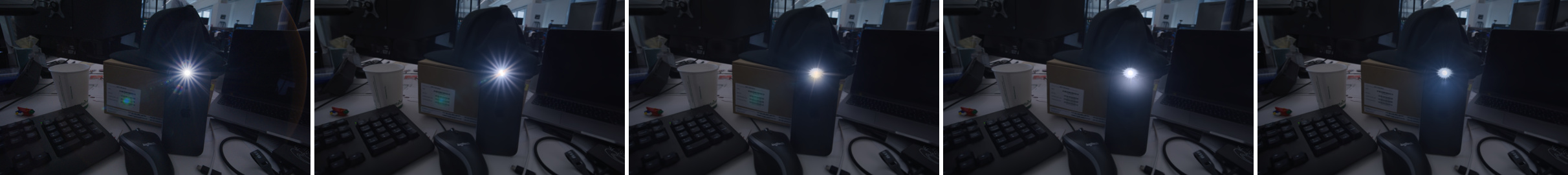} \\
        \multicolumn{2}{c}{\scriptsize Flare} \\
        \rotatebox{90}{\scriptsize \shortstack{w $\delta_l$ \\ \& w $\delta_f$}} &
        \includegraphics[width=0.88\linewidth]{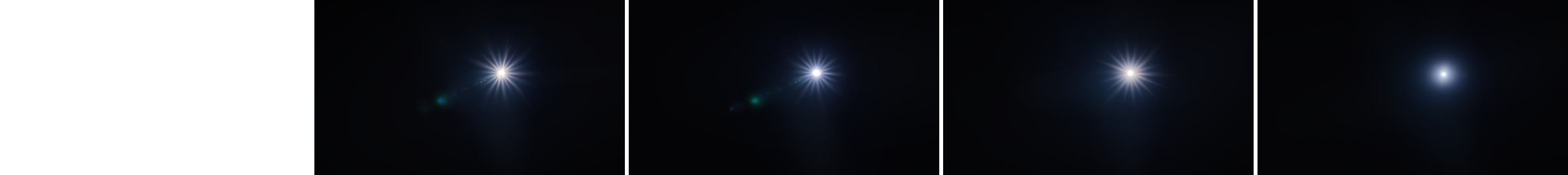} \\
        \rotatebox{90}{\scriptsize \shortstack{wo $\delta_l$ \\ \& w $\delta_f$}} &
        \includegraphics[width=0.88\linewidth]{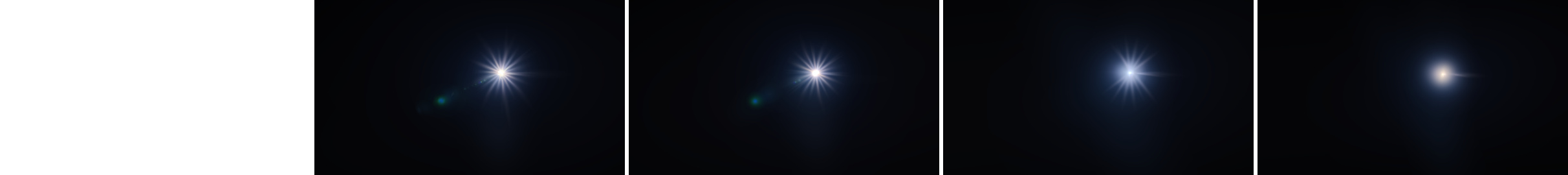} \\
        \rotatebox{90}{\scriptsize \shortstack{w $\delta_l$ \\ \& wo $\delta_f$}} &
        \includegraphics[width=0.88\linewidth]{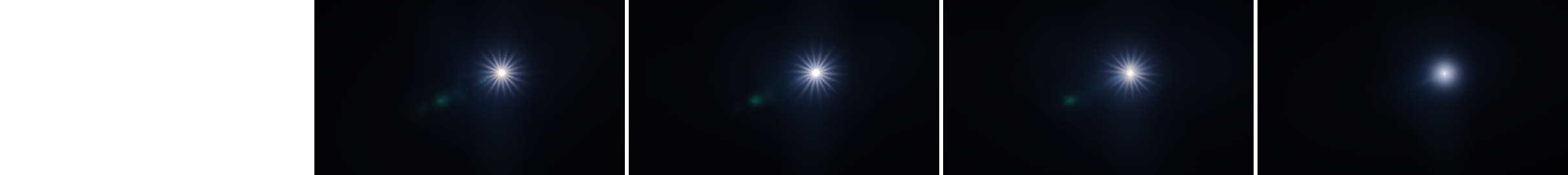} \\
        \rotatebox{90}{\scriptsize \shortstack{wo $\delta_l$ \\ \& wo $\delta_f$}} &
        \includegraphics[width=0.88\linewidth]{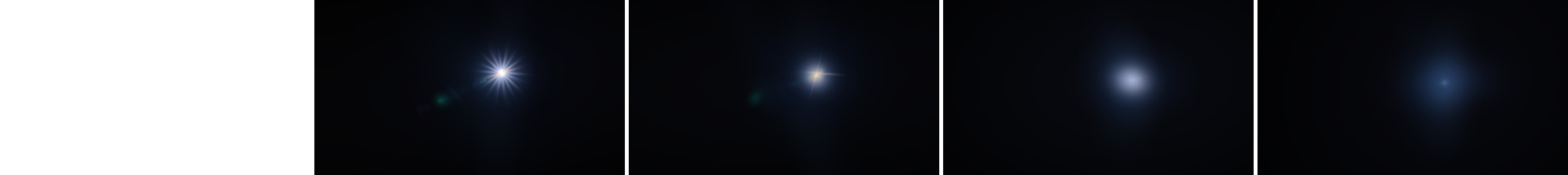} \\
        & \begin{minipage}[t]{0.95\linewidth}
          \begin{tabular}{@{}p{0.2\linewidth}@{}p{0.2\linewidth}@{}p{0.2\linewidth}@{}p{0.2\linewidth}@{}p{0.2\linewidth}@{}}
            \centering\scriptsize GT & \centering\scriptsize $\sigma\!=\!0.0$ & \centering\scriptsize $\sigma\!=\!0.1$ & \centering\scriptsize $\sigma\!=\!0.5$ & \centering\scriptsize $\sigma\!=\!1.0$
          \end{tabular}
        \end{minipage} \\
    \end{tabular}
    \caption{Qualitative robustness on the \emph{hat} sequence to noise in the input light-source location $l$. Top four rows: composite render (scene$+$flare); bottom four rows: isolated flare component. Rows within each group enable, in order, both corrections, only $\delta_f$, only $\delta_l$, and neither. With at least one correction the flare remains centred on the source and retains its starburst through $\sigma=0.5$, softening at $\sigma=1.0$; with neither, structure is lost from $\sigma=0.1$ onward.}
    \label{fig:light_robustness_quals}
    \Description{A table of image strips on the \emph{hat} sequence, with eight rows grouped into two sets of four: the top four rows show the composite scene-plus-flare render, and the bottom four show the isolated flare component. Within each set of four, rows correspond to: both corrections enabled, only the finer deformation, only the light-position correction, and neither. Each row has four columns for increasing injected noise levels (sigma = 0, 0.1, 0.5, 1.0) plus a ground-truth column. Rows with at least one correction retain a sharp, centered starburst flare through moderate noise; the row with neither correction loses its starburst structure starting at the lowest noise level and degrades to a faint blob at the highest noise level.}
\end{figure}

As shown in~\figref{fig:light_robustness}, the composite PSNR of our full model degrades by $1.5$\,dB across the full noise range (from $33.82$ to $32.29$\,dB at an average light-position error of $262$\,px) and flare PSNR by $1.4$\,dB ($33.48$ to $32.12$\,dB). Scene PSNR, omitted from the plot, stays within $0.1$\,dB for every variant except the one with both corrections removed, where it drops from $32.19$ to $31.57$\,dB: light-position error leaks into the scene only when the flare model cannot absorb it. Removing either correction alone costs little: at the highest noise level the composite reaches $32.04$\,dB without $\delta_l$ and $31.86$\,dB without $\delta_f$. Removing both is what breaks robustness -- the composite falls from $32.72$ to $28.24$\,dB and flare PSNR from $32.69$ to $27.27$\,dB. $\delta_l$ and $\delta_f$ are therefore largely redundant defences against detection noise: each gives the flare model enough freedom to compensate for a misplaced principal-point line. \figref{fig:light_robustness_quals} shows the same pattern on the \emph{hat} sequence: with at least one correction, the flare stays centred on the source and keeps its structure through moderate noise, softening only at $\sigma=1.0$; with both removed, its starburst is already lost at $\sigma=0.1$ and it collapses to a faint blob by $\sigma=1.0$. Because each light is displaced once for the whole sequence, the resulting error is consistent across views rather than independent per view, and so cannot be averaged out. Together with the \emph{wo G-SAM2} ablation, which discards detection entirely and places every light near the scene centroid, this indicates that our reconstruction tolerates both inaccurate and missing light detections.

\begin{figure}[t]
    \centering
    \input{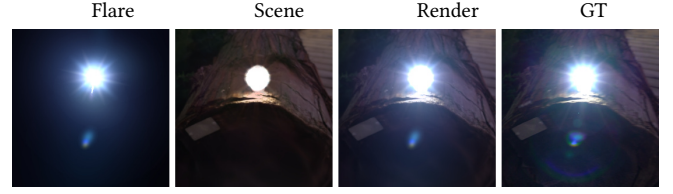}
    \caption{A limitation of our model. While our model can accurately reconstruct the locations and colors of most flares, it struggles with ring-shaped flares, as Gaussians are inherently more opaque towards the center. }
    \Description{Four images of a tree-trunk night scene with a bright light and a small blue secondary flare. The isolated ``Flare'' render shows a glowing ring with a dark hollow center instead of the expected ring shape. The ``Scene'', ``Render'', and ``GT'' (ground truth) images look nearly identical, showing the tree trunk with a bright light and, in the ground truth, a faint colorful ring artifact near the blue secondary flare that is missing from the reconstructed render.}
    \label{fig:limitations}
\end{figure}
\section{Future Work}
Our flare representation model improves 3DGS's performance in scenes with lens flares by decomposing the scene into a lens-flare component and a flare-free component, enabling many downstream applications such as flare transfer and editing. While our 2D removal model already handles ring-shaped flares (we explicitly include procedurally generated rings in its training set, see \secref{sec:datasets}), their \emph{3D representation} remains an open problem: because our flare Gaussians have peak opacity at the mean and decrease radially away from it (see~\figref{fig:limitations}), a single Gaussian cannot form the annular profile required for rings. We therefore leave the 3D representation of ring-shaped flares for future work. Further, light sources out of frame are handled based on projected depth, however, those occluded by scene geometry within a captured view are currently handled implicitly through our flare representation model, rather than through an explicit occlusion-aware mechanism; we leave this to future work.

Moreover, our flare removal model is bounded by the flare types represented in our training datasets (\secref{sec:datasets}); flares that fall far outside this distribution are harder for it to remove convincingly. Additionally, since a real scene cannot be captured simultaneously with and without the same flare, the removal model is trained on flares composited onto otherwise clean images rather than on paired real captures; closing this synthetic-to-real gap would likely require a capture rig that optically isolates the flare, which we leave to future work.

\section{Conclusion}
We introduced two models: a lens-flare removal model and, to our knowledge, the first lens-flare representation reconstructed from captured multi-view images. We compiled a dataset of full-image lens flares from publicly available real-world data. Further, we evaluated our flare removal models on the established Flare7k++ benchmark and our proposed VFX benchmark, where on each one of our two models outperforms all compared baselines. Moreover, our 2D flare removal model outperforms baselines, even in 3D scenes, demonstrating state-of-the-art performance on our proposed benchmark for consistent flare removal across multiple views. We used flare removal to develop the lens-flare representation model and demonstrated a decomposed reconstruction of a given scene with flares into scene Gaussians and our flare Gaussians. We believe these models would be useful for different downstream tasks and end-user applications.

\bibliographystyle{ACM-Reference-Format}
\bibliography{references}
\clearpage
\appendix
\section{Appendix}
In this supplementary material, we first provide implementation details for our models, followed by a visualization of our multi-view flare removal benchmark. Finally, we show additional qualitative results of our experiments in the main paper. Moreover, we show novel views of our reconstructed flare models in the supplementary video.
\subsection{Implementation}
\label{sec:impl}
\subsubsection{Flare removal model} We base our model and implementation on Difix3D+~\cite{wu2025difix3d}. However, as described in Sec. 3.1 (main), unlike in Difix3D+, we use LoRA~\cite{hu2022lora} to fine-tune the latent UNet diffusion model and the VAE. In total, we update ${\approx}19.6$M parameters -- rank-$16$ LoRA adapters on the UNet ($11.8$M) and VAE ($3.6$M), plus the VAE's skip-connection and output convolutions ($4.2$M) -- against $950$M frozen UNet and VAE weights, about $2\%$ of the model.

We train the model with $4$ different kinds of flare datasets: (i) Flare7K++ real scattering flares, (ii) Flare7K++ reflective flares, (iii) our proposed VFX dataset, and (iv) procedurally generated reflective flares. These flares are added to images from the Flickr24K dataset~\cite{zhang2018single}. We condition the models on the text prompt: ``remove flares and keep the original image. High quality. Retain original colors''. We train the model with a batch size of $1$ for $200k$ iterations with a learning rate of $1e^{-4}$ on an Nvidia A100 GPU, which takes approximately $20$ hours; training only  ${\approx}2\%$ of the model is the main reason it converges so much faster than the baselines, which are trained from scratch and require more than four days. We train the three baselines (ACL-FR~\cite{zhou2025improving}, FR~\cite{wu2021train}, and Flare7K++~\cite{dai2023flare7kpp}) on these datasets, strictly adhering to each method's training strategy and hyperparameters, and selecting the best-performing model (Uformer) for each method.

\subsubsection{Flare representation} For each light source, the deformation network is an instant-NGP-style model: the light input (the absolute normalized image-plane offset of the projected light from the principal point, together with its norm) and the camera position are encoded with multiresolution hash grids of $16$ levels and $2$ features per level, a base resolution of $16$, hash-table sizes of $2^{17}$ and $2^{15}$, and per-level scales of $1.8$ and $1.6$, respectively; the 1D canonical position uses a frequency encoding with $3$ bands. The result is decoded by an MLP backbone with $3$ hidden layers of width $128$ and ReLU activations. Flare Gaussians are placed at a depth of $0.21$ in the camera frame, just beyond the near plane. We optimize the network and encodings with Adam at a learning rate of $10^{-4}$. The canonical flare Gaussians use a position learning rate of $1.6\times10^{-4}$, decayed exponentially to $1.6\times10^{-6}$, and learning rates of $2.5\times10^{-3}$, $0.05$, $0.1$ and $10^{-3}$ for their features, opacities, scales and rotations, respectively. We train for $30{,}000$ iterations. We initialize $200$ flare Gaussians at different positions along a line with reasonable sizes, ensuring that they do not occupy the entire image. We clone and prune the flare Gaussians every $500$ iterations, from $1k$ to $12k$ iterations. We allow cloning to be aggressive for all opacities above $0.005$, with a clone ratio of $0.2$, and perturb cloned Gaussians with Gaussian noise of standard deviation $0.02$. We only prune Gaussians when the opacity is below $0.002$. In the \emph{wo G-SAM2} ablation, each light is initialized at the centroid of the scene's Colmap point cloud, offset by uniform noise in $[0,1)$ scene units along each axis.

\subsection{3D Reconstruction with Flares}

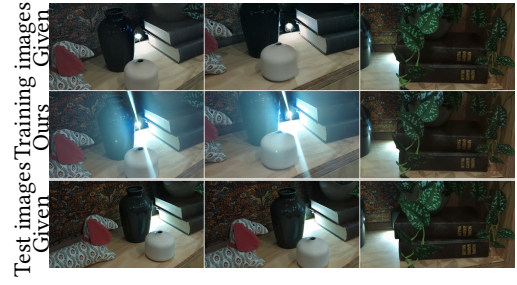
\begin{figure}
    \centering
    \begin{overpic}[width=0.72\linewidth, tics=5]
    {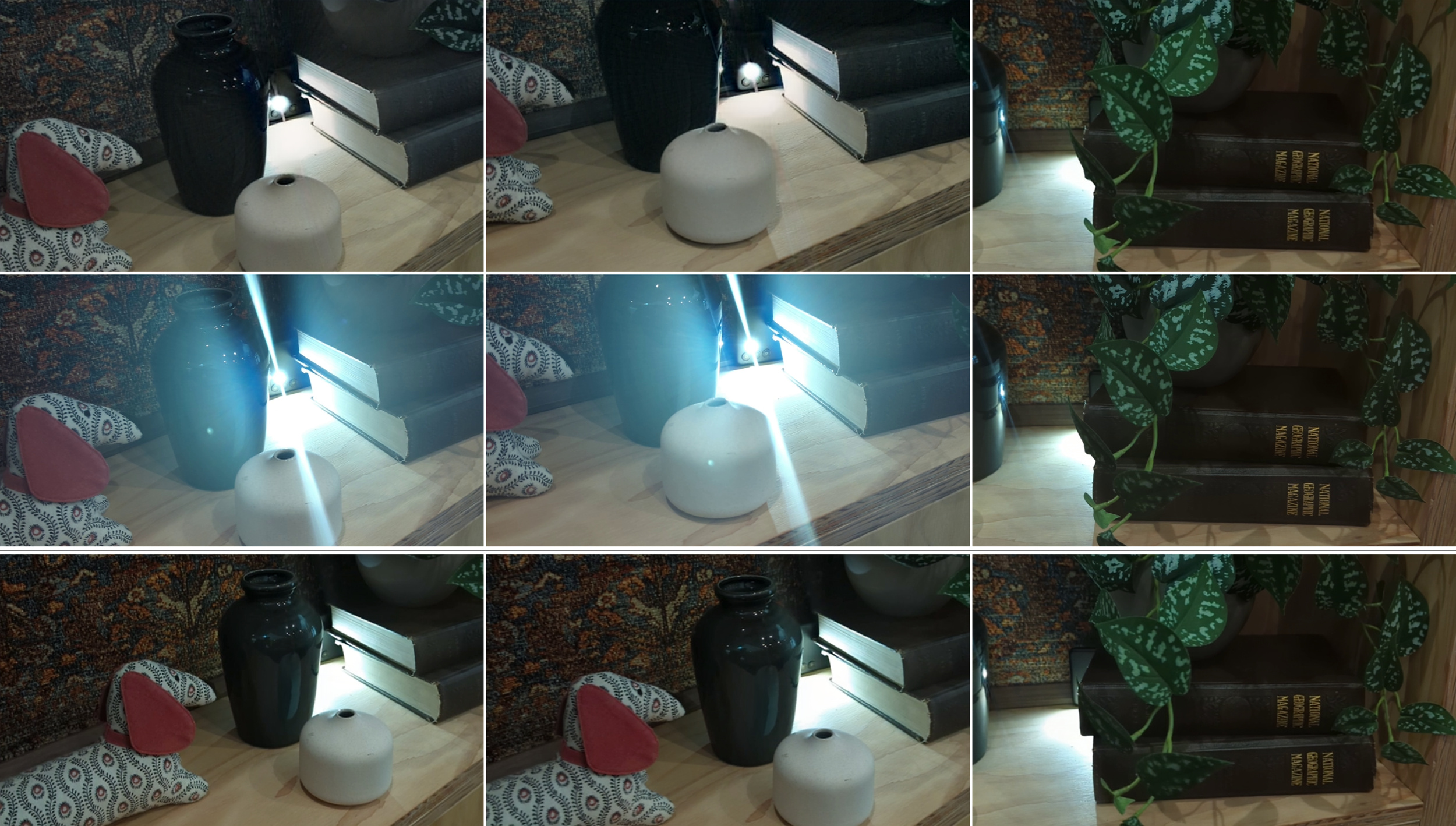}
    \put(-7.5,23){\rotatebox{90}{Training images}}
    \put(-3.5,43){\rotatebox{90}{Given}}
    \put(-3.5,25){\rotatebox{90}{\ourswo}}  
    \put(-3.5,3){\rotatebox{90}{Given}}
    \put(-7.5,-2){\rotatebox{90}{Test images}}
    \end{overpic}
    \caption{Example train and test images from `dog' sequence for our `3D Reconstruction with Flare Corruption' experiment in Sec. 5.2 (main). We show the given training images in the top row and the flare-removed training images in the second row using our flare-removal model. In the bottom row, we show example test images used for evaluation.}
    \Description{A grid of photographs from a `dog' sequence, arranged in three rows. The top row (``Training images'' / ``Given'') shows several photos of a scene with visible lens flares. The middle row (``Training images'' / \ourswo) shows the same photos with flares removed by our model. The bottom row (``Test images'' / ``Given'') shows held-out photos of the scene without flares, used as evaluation ground truth.}
    \label{fig:datasetmvflare}
\end{figure}

We show example images in~\figref{fig:datasetmvflare} from a scene in the dataset of our 3D reconstruction with flares in Sec. 5.2 (main).

\subsection{Reliability of Light-Source Detection}
\label{sec:exp_light_reliability}

\begin{figure}[t]
  \centering
  \includegraphics[width=\linewidth]{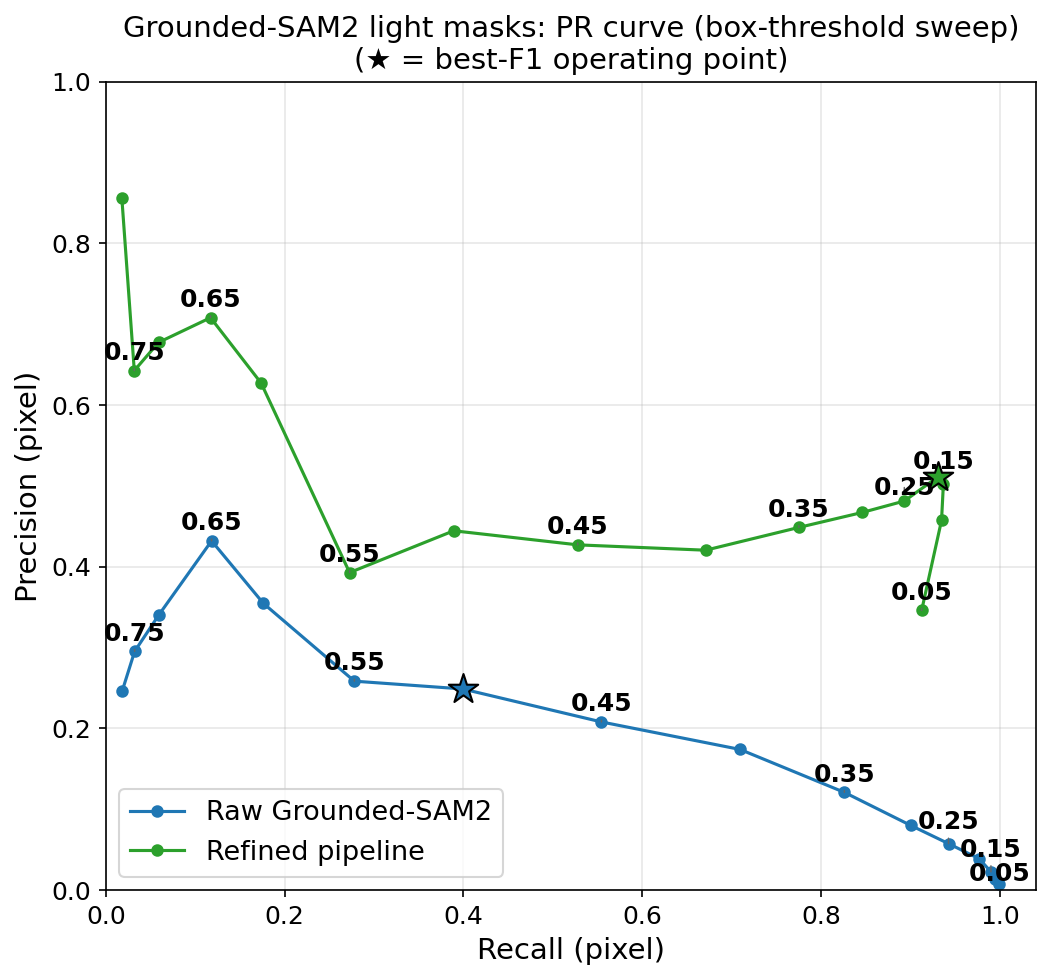}
  \caption{Pixel-level precision--recall of light-source detection on the synthetic
  benchmark ($1{,}128$ images, $1$--$5$ sources each), sweeping the GroundingDINO
  box-confidence threshold (labels). Our refined pipeline (green) improves precision over the raw
  Grounded-SAM2 output (blue), at a small cost in recall at the lowest thresholds; $\star$ marks the best-F1
  operating point.}
  \Description{A precision-recall scatter plot titled ``Grounded-SAM2 light masks: PR curve (box-threshold sweep)''. Two lines, with every other point labeled by its GroundingDINO confidence threshold (the sweep runs from 0.05 to 0.80), plot precision (pixel) on the y-axis against recall (pixel) on the x-axis. The blue ``Raw Grounded-SAM2'' line peaks near precision 0.43 at low recall and falls steadily to near 0 precision at recall 1.0, with a star marking its best-F1 point around precision 0.25, recall 0.40. The green ``Refined pipeline'' line stays well above the blue line across the range, peaking near precision 0.85 at low recall and remaining around 0.35--0.5 precision at high recall, with a star marking its best-F1 point around precision 0.51, recall 0.93.}
  \label{fig:light_pr}
\end{figure}

\newlength{\tilew}\setlength{\tilew}{0.31\linewidth}
\newcommand{\dtile}[2]{%
  \begin{overpic}[width=\tilew,unit=1mm]{figures/tiles/#1}%
    \put(1.5,1.5){\scriptsize\colorbox{black}{\textcolor{white}{\textbf{#2}}}}%
  \end{overpic}}
\begin{figure}[t]
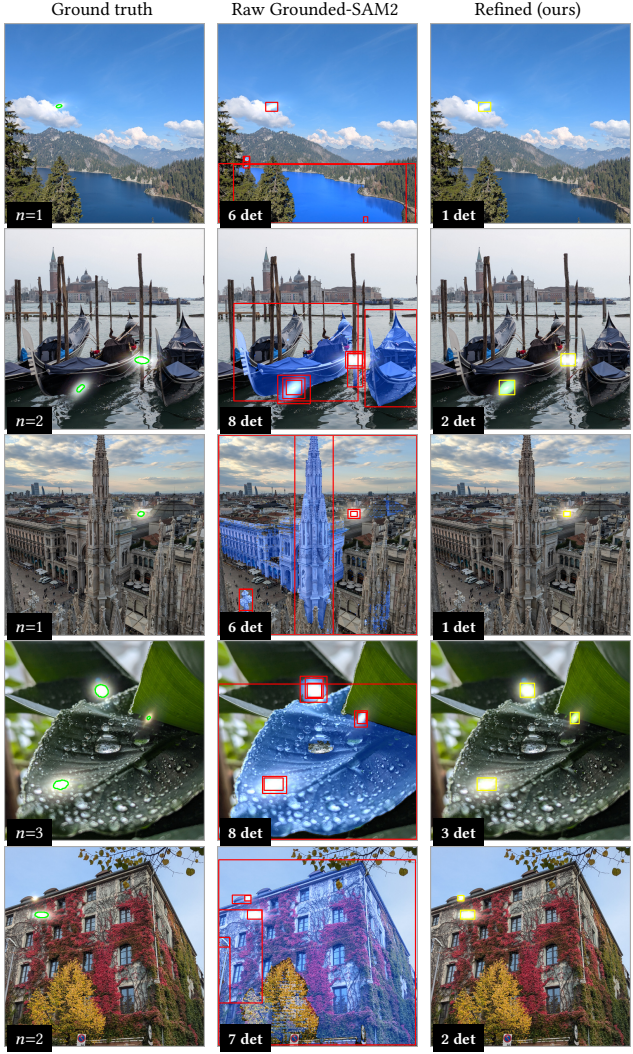

  \centering
  \footnotesize
  \setlength{\tabcolsep}{1.5pt}
  \renewcommand{\arraystretch}{1.2}
  \begin{tabular}{@{}ccc@{}}
    Ground truth & Raw Grounded-SAM2 & Refined (ours) \\[1pt]
    \dtile{tile_r0_gt.jpg}{$n{=}1$} & \dtile{tile_r0_raw.jpg}{6 det} & \dtile{tile_r0_ref.jpg}{1 det} \\
    \dtile{tile_r1_gt.jpg}{$n{=}2$} & \dtile{tile_r1_raw.jpg}{8 det} & \dtile{tile_r1_ref.jpg}{2 det} \\
    \dtile{tile_r2_gt.jpg}{$n{=}1$} & \dtile{tile_r2_raw.jpg}{6 det} & \dtile{tile_r2_ref.jpg}{1 det} \\
    \dtile{tile_r3_gt.jpg}{$n{=}3$} & \dtile{tile_r3_raw.jpg}{8 det} & \dtile{tile_r3_ref.jpg}{3 det} \\
    \dtile{tile_r4_gt.jpg}{$n{=}2$} & \dtile{tile_r4_raw.jpg}{7 det} & \dtile{tile_r4_ref.jpg}{2 det} \\
  \end{tabular}
  \caption{Qualitative light detection at box threshold $0.20$ on our own
  captured images. \emph{Left:} ground-truth outline (green) with source
  count $n$. \emph{Middle:} raw Grounded-SAM2 (boxes red, SAM2 masks blue),
  which fires on specular and high-contrast non-emissive regions --
  reflective water and gondola hulls, building facades and spires, wet
  foliage -- inflating the detection count well beyond $n$. \emph{Right:}
  our refined pipeline (boxes yellow, masks green), which removes these
  false positives via the exposure gate and merges duplicate/overlapping
  boxes onto a single source per light.}
  \Description{A table of five rows, each showing a captured scene as three tiles: ground truth (green outline around the true light sources, with source count n labeled 1 to 3), raw Grounded-SAM2 detections (red boxes and blue masks, labeled with a detection count of 6 to 8, generally over-detecting compared to n by firing on reflective or high-contrast non-emissive regions such as water, building facades, and foliage), and the refined pipeline (yellow boxes and green masks, labeled with a detection count close to n, showing far fewer false positives).}
  \label{fig:light_examples}
\end{figure}

Our flare representation relies on Grounded-SAM2 to obtain the initial 2D
light-source locations that are aggregated across views and refined by our
model (Sec.~\ref{sec:flarerep}, main). As this detection front end is an off-the-shelf
component rather than a contribution of our method, we characterize its
reliability here. Real captures lack ground-truth light-source annotations, so
we build a synthetic benchmark to quantify the reliability of the 2D detection
stage, and separately analyze its behavior on our captured scenes.

We composite real light sources from
Flare7K++~\cite{dai2023flare7kpp} onto light-free backgrounds from
Flickr24K~\cite{zhang2018single} and our captured images ($1$--$5$ sources per image, random affine
transforms), yielding $1{,}128$ images with exact ground-truth masks.

We compare the union of predicted SAM2 masks against the ground-truth masks at
the pixel level (precision, recall, F1, IoU), sweeping GroundingDINO's
box-confidence threshold from $0.05$ to $0.80$ (~\figref{fig:light_pr}), since
this threshold alone governs the pipeline's precision--recall trade-off (SAM2 only segments the boxes it is given). The raw pipeline peaks at F1 $0.31$
(threshold $0.50$): low thresholds yield many spurious boxes, and SAM2 tends to
segment the illuminated halo rather than the compact source. Our refinements --
resolving overlapping boxes and discarding non-over-exposed detections --
substantially improve precision over the raw pipeline, at a small cost in recall at the lowest thresholds, reaching F1 $0.66$ (precision
$0.51$, recall $0.93$, IoU $0.49$) at threshold $0.20$ (Fig.~\ref{fig:light_examples}).

On our nine captured scenes for flare reconstruction in \secref{sec:flare_rec_dataset} ($\sim$10.8K frames: every frame of the videos, rather than only those sampled for training and testing; no ground truth), at the box threshold of $0.35$ used in our pipeline, GroundingDINO
proposes $10{,}526$ boxes; the over-exposure filter discards $993$ ($9.4\%$) and
overlap-resolution a further $499$ ($5.2\%$ of the remainder), leaving $9{,}034$ ($85.8\%$). Notably, $93.6\%$ of the
overlap-removed boxes fall below the IoU threshold and would survive standard
NMS, motivating our containment-based criterion instead. Lowering the threshold
to $0.10$ inflates proposals roughly tenfold, of which the refinements prune only
$\sim\!82\%$. Detection is also intermittent across frames ($2{,}216$ frames with
no detection, $319$ fully filtered), which is precisely what our multi-view 3D
aggregation with TRASE~\cite{li2026trase} resolves, consolidating noisy per-frame
2D detections into a consistent set of 3D light sources per scene.

\end{document}